\documentclass[twoside]{article}

\usepackage[accepted]{aistats2024}
\usepackage{natbib}

\usepackage{hyperref}
\usepackage{amsmath,amssymb,amsthm}
\usepackage{bm}
\usepackage{pgfplots,tikz}
\pgfplotsset{compat=newest}
\usepackage{enumitem}
\usepackage{caption,subcaption}
\usepackage{booktabs,multirow}
\usepackage[ruled,linesnumbered]{algorithm2e}

\newtheorem{theorem}{Theorem}
\newtheorem{lemma}[theorem]{Lemma}
\newtheorem{definition}{Definition}
\newtheorem{remark}{Remark}

\newcommand{\f}{\mathbf{f}}
\newcommand{\bv}{\mathbf{v}}
\newcommand{\w}{\mathbf{w}}
\newcommand{\x}{\mathbf{x}}
\newcommand{\y}{\mathbf{y}}
\newcommand{\R}{\mathbb{R}}

\newcommand{\E}[2]{\mathbb{E}_{#2}\left[#1\right]}
\newcommand{\KL}{\mathrm{KL}}
\newcommand{\D}{\mathcal{D}}
\newcommand{\p}{\partial}
\renewcommand{\P}{\mathcal{P}}
\newcommand{\F}{\mathcal{F}}
\newcommand{\Ec}{\mathcal{E}}
\newcommand{\G}{\mathcal{G}}
\newcommand{\g}{\mathbf{g}}
\renewcommand{\H}{\mathcal{H}}
\renewcommand{\d}{\mathrm{d}}
\newcommand{\ndcg}{\mathrm{NDCG}}
\newcommand{\dcg}{\mathrm{DCG}}
\renewcommand{\L}{\mathcal{L}}

\DeclareMathOperator*{\argmax}{argmax}

\begin{document}

%
\runningtitle{Multi-Objective Optimization via Wasserstein-Fisher-Rao Gradient Flow}

%
\runningauthor{Y. Ren, T. Xiao, T. Gangwani, A. Rangi, H. Rahmanian, L. Ying, S. Sanyal}

\twocolumn[

\aistatstitle{Multi-Objective Optimization\\ via Wasserstein-Fisher-Rao Gradient Flow}

\aistatsauthor{ 
    Yinuo Ren \\ Stanford University \\
    \And Tesi Xiao \\ Amazon \\
    \And Tanmay Gangwani \\ Amazon \\
    \And Anshuka Rangi \\ Amazon \\
    \AND Holakou Rahmanian \\ Amazon \\
    \And Lexing Ying \\ Stanford University \\
    \And Subhajit Sanyal\\ Amazon}
\aistatsaddress{} 
]

\begin{abstract}
    Multi-objective optimization (MOO) aims to optimize multiple, possibly conflicting
    objectives with widespread applications. We introduce a novel interacting particle method for MOO inspired by molecular dynamics simulations.
    Our approach combines overdamped Langevin and birth-death
    dynamics, incorporating a ``dominance potential'' to steer particles toward global Pareto optimality. In contrast to previous methods, our method is able to relocate dominated particles, making it particularly adept at managing Pareto fronts of complicated geometries. Our method is also theoretically grounded as a Wasserstein-Fisher-Rao gradient flow with convergence guarantees. Extensive
    experiments confirm that our approach outperforms state-of-the-art methods on challenging synthetic and real-world datasets.
\end{abstract}

\section{INTRODUCTION}

Multi-objective optimization (MOO) addresses optimization scenarios where multiple objectives are simultaneously and systematically optimized. Given that these objectives may inherently conflict, MOO seeks to identify a diversified set of solutions on the \emph{Pareto front}, where no solution can enhance one objective without deteriorating at least one other. Determining the Pareto front is intricate due to its typically non-closed-form expression and potentially complicated geometries. Real-world applications of MOO span various domains, including control systems~\citep{gambier2007multi}, energy saving~\citep{cui2017multi}, and economics and finance~\citep{tapia2007applications}.

Over the past few decades, MOO has been extensively explored in literature. Notable traditional methods include the evolutionary algorithms~\citep{tamaki1996multi,deb2002fast,konak2006multi,reyes2006multi,zhang2007moea,zhou2011multiobjective}, and
Bayesian optimization~\citep{laumanns2002bayesian,belakaria2020uncertainty,konakovic2020diversity,tu2022joint}. These methods, while effective, can be computationally expensive and less adaptable to high-dimensional problems.
Recently, gradient-based MOO methods have been developed for various machine learning tasks~\citep{sener2018multi}. Many such methods rely on \emph{preference vectors}~\citep{lin2019pareto,mahapatra2020multi,liu2021stochastic}. However, their efficacy often hinges on the vector selection, making them potentially heuristic for unknown Pareto fronts. Another line
of research has explored \emph{hypernetwork} methods~\citep{navon2020learning,lin2020controllable,chen2022multi,hoang2023improving}, which may fail to capture the discontinuity of the Pareto front and thus only work for simple Pareto front geometries. We refer the readers to Appendix~\ref{app:related_works} for a more comprehensive literature review.
\vspace{-.5em}
\paragraph{Contributions.}
In this paper, we propose a novel interacting particle method for MOO inspired by molecular dynamics simulations in computational physics\footnote{Code is accessible at \url{https://github.com/yinuoren/particlewfr}}. Particle diversity is maintained by integrating a repulsive two-body interatomic potential. A standout feature of our method is the introduction of the \emph{dominance potential}, which assigns higher potential to particles being dominated. Together with the incorporation of the stochastic birth-death process, we facilitate the direct relocation or, more intuitively, the \emph{teleportation} of dominated particles to the Pareto front. This not only ensures the global Pareto optimality of the solutions but also significantly bolsters performance. 
\begin{itemize}[leftmargin=*]
    \setlength\itemsep{0.01em}
    \item We propose the \emph{Particle-WFR} method, evolving a population of randomly-initialized \emph{particles} by a combination of overdamped Langevin and stochastic birth-death dynamics towards the Pareto front.
    \item Theoretical grounding of our method is achieved by formulating it as a Wasserstein-Fisher-Rao gradient flow in the space of probability measures, providing provable convergence guarantees.
    \item Extensive experiments are conducted on both synthetic and real-world datasets, and the results demonstrate the superiority of our method over the state-of-the-art methods in addressing challenging tasks with complicated Pareto fronts.
\end{itemize}

\section{PRELIMINARIES}

This section defines the multi-objective optimization problem and introduces fundamental concepts and notations. A brief overview of gradient flows in the space of probability measures is also provided.

\subsection{Pareto Optimality}

In MOO, we consider the following problem of minimizing $m$ objective functions simultaneously:
\begin{equation}
    \min_{\x \in \D} \f(\x) = \left(f_1(\x), \cdots, f_m(\x)\right),
\end{equation}
where $\D\subseteq\R^d$ is the feasible region, and $f_i:\R^d\to\R$, $i\in[m]=\{1,\dots,m\}$ are objective functions. Unlike single-objective optimization problems, we need the following notion of \emph{Pareto optimality} to determine the superiority of a solution:
\begin{definition}[Pareto optimality]
    A solution $\x^*$ is said to be \emph{Pareto optimal} if there does not exist another solution $\x'$ such that $f_i(\x') \leq f_i(\x^*)$ for all $i\in[m]$ and $f_j(\x') < f_j(\x^*)$ for at least one $j\in[m]$. A solution $\x^*$ is said to be \emph{locally Pareto optimal} if there exists a neighborhood $\mathcal{N}(\x^*)$ of $\x^*$ such that $\x^*$ is Pareto optimal in $\mathcal{N}(\x^*)$.
\label{def:pareto}
\end{definition}

In real applications, we are often interested in the set of Pareto optimal
solutions, called the \emph{Pareto front}, denoted by $\P$. Our goal is to find a set of
solutions on the Pareto front, which should favorably be
diversified, explicitly showcasing the characteristics of the problem for the
decision maker to make the final choice~\citep{tamaki1996multi}. However, the
Pareto front may exhibit complicated geometries that are disconnected or highly
non-convex~\citep{kulkarni2022regularities}, making it very challenging to
compute. Singularities may arise even with only two quadratic objective
functions~\citep{sheftel2013geometry}. 

\subsection{Wasserstein-Fisher-Rao Gradient Flow}

In many sampling problems, one would like to design an evolution of probability measures $\rho_t$ that converges to a target distribution $\rho^*$
as $t\to\infty$. One of the most intuitive and powerful tools for this
purpose is the gradient flow. Specifically, a gradient flow represents a continuous-time dynamical system guiding $\rho_t$ towards the target distribution $\rho^*$, recognized as the minimizer of a certain energy functional $\Ec[\rho]$. 
Generally, the introduction of different Riemannian metrics yields various gradient flows, each defining unique geodesics within the space of probability measures.

Consider two probability measures $\rho_0$ and $\rho_1$, the \emph{Wasserstein} distance is defined as
\begin{equation}
    d_{\sf W}(\rho_0, \rho_1) = \inf_{\pi \in \Pi(\rho_0, \rho_1)} \int_{\R^d \times \R^d} \|\x - \y\|^2  \pi(\d \x,\d \y),
    \label{eq:wasserstein}
\end{equation}
where $\Pi(\rho_0, \rho_1)$ denotes the set of all joint probability measures on $\R^d \times \R^d$ with marginals $\rho_0$ and $\rho_1$. The Benamou-Brenier theorem~\citep{benamou2000computational} provides an insightful geodesic interpretation for the Wasserstein metric:
\begin{equation*}
\begin{aligned}
    &d_{\sf W}(\rho_0, \rho_1) \\
    =& \inf \left\{ \int_0^1 \int \|\bv_t\|^2 \d \rho_t \d t\bigg| \p_t \rho_t = - \nabla \cdot (\rho_t \bv_t)\right\},
\end{aligned}
\end{equation*}
and the corresponding \emph{Wasserstein gradient flow} of the energy $\Ec[\rho]$ is
\begin{equation}
    \p_t \rho_t = \nabla \cdot (\rho_t \nabla \delta_\rho\Ec[\rho_t]),
    \label{eq:wgf}
\end{equation}
where $\delta_\rho \Ec[\rho]$ denotes the Fr\'echet derivative of $\Ec[\rho]$ w.r.t. $\rho$. One notable relevant result is that when $\Ec[\rho]$ is selected as the Kullback-Leibler divergence between $\rho$ and $\rho^*$, the resulting Wasserstein gradient flow is the overdamped Langevin dynamics~\citep{jordan1998variational}.

In parallel, the \emph{Fisher-Rao} metric is another important metric in the space
of probability measures, resonating with concepts including the Fisher information and the Hellinger distance familiar to statisticians. It also has a geodesic interpretation:
\begin{equation*}
    d_{\sf FR}(\rho_0, \rho_1) = \inf\left\{\int_0^1 \int \widetilde\beta_t^2\d \rho_t \d t \bigg|  \p_t \rho_t = \rho_t \widetilde \beta_t  \right\},
\end{equation*}
where $\widetilde \cdot$ is a shorthand notation for $\cdot - \E{\cdot}{\rho_t}$.
The corresponding \emph{Fisher-Rao gradient flow} is given by
\begin{equation}
    \p_t \rho_t = - \rho_t \widetilde {\delta_\rho \Ec[\rho_t]}.
    \label{eq:frgf}
\end{equation}

Intuitively, as the Wasserstein gradient flow derives from the optimal transport problem~\eqref{eq:wasserstein}, it redistributes probability
densities by transporting them along paths directed by the Kantorovich potential $\delta_\rho \Ec[\rho]$. On the other hand, the Fisher-Rao gradient flow teleports mass, \emph{i.e.} locally reshape probability densities, according to the deviation
of $\delta_\rho \Ec[\rho]$ from its expectation.

Recent advances~\citep{liero2016optimal,liero2018optimal,chizat2018interpolating} suggest the following \emph{Wasserstein-Fisher-Rao} (WFR) distance, also known as the \emph{spherical
Hellinger-Kantorovich} distance:
\begin{equation*}
    \begin{aligned}
        d_{\sf WFR}(\rho_0, \rho_1) = \inf\bigg\{ &\int_0^1 \int \|\bv_t\|^2 + \widetilde\beta_t^2\d \rho_t \d t \bigg| \\
        &\quad\quad \p_t \rho_t = - \nabla \cdot (\rho_t \bv_t) + \rho_t \widetilde \beta_t  \bigg\},
    \end{aligned}
\end{equation*}
as a natural combination of the Wasserstein and
Fisher-Rao distances for the study of unbalanced optimal
transport, \emph{e.g.} accelerating Langevin sampling~\citep{lu2019accelerating}
and learning Gaussian mixture~\citep{yan2023learning}. It leads to the following
WFR gradient flow of $\Ec[\rho]$:
\begin{equation}
    \p_t \rho_t = \nabla \cdot (\rho_t \nabla \delta_\rho\Ec[\rho_t]) - \rho_t \widetilde {\delta_\rho \Ec[\rho_t]},
    \label{eq:wfrgf}
\end{equation}
which offers promising implications, which we delve deeper into in the next section.

\section{ALGORITHM}

In this section, we present our proposed method for MOO. We first design the functional $\Ec[\rho]$ on which we employ the \emph{Wasserstein-Fisher-Rao} (WFR) gradient flow and thereby encourage the diversity and
global Pareto optimality. We then offer the theoretical formulation of our method as the
WFR gradient flow. Finally, we discuss the interacting
particle implementation of the WFR gradient flow, which is realized through the alternate application of the overdamped
Langevin and birth-death dynamics.

\subsection{Designing the Functional $\Ec[\rho]$}

Consider the initial setup where the probability measure $\rho_0$ is arbitrarily initialized within the feasible region $\D$. As we implement the WFR gradient flow of $\Ec[\rho]$
to evolve $\rho_t$ towards the target distribution $\rho^*$, we would like to
design the functional $\Ec[\rho]$ such that its minimizer $\rho^*$ has the
following properties:
\begin{enumerate}[leftmargin=*]
    \item {\bf Global Pareto optimality:} $\rho^*$ should in close proximity to the Pareto front. Particularly, $\rho^*$ should only cover
    \emph{global} Pareto optimal solutions and exclude those only \emph{local} Pareto optimal.
    \item {\bf Diversity:} To ensure a comprehensive representation, $\rho^*$ should be diversified, spanning the entirety of the Pareto front. It should not be concentrated only on a subset of the Pareto
    front.
\end{enumerate}
For each property above, we propose a corresponding term in the
functional $\Ec[\rho]$, as explained below:
\paragraph{Objective Function Term.} 


A straightforward strategy to force the minimizer $\rho^*$ closer to
the Pareto front is the weighted sum method, \emph{i.e.} minimizing a linear
combination of the objective functions $F(\x) = \sum_{i=1}^m \alpha_i f_i(\x)$
where $\alpha_i$, $i\in[m]$ are predetermined parameters. However, 
this method is susceptible to the varied scales and ranges of the 
objective functions, and the concavity of the Pareto front.

Addressing this challenge, the Multi-Gradient Descent Algorithm
(MGDA)~\citep{desideri2012multiple} relaxes the parameters $\alpha_i$ to be
space-dependent determined by solving the following
minimal norm optimization problem for each $\x$:
\begin{equation}
    \min_{\bm{\alpha}\in\Delta_m} \quad \left\|\sum_{i=1}^m \alpha_i(\x) \nabla f_i(\x) \right\|,
    \label{eq:mgda}
\end{equation}
where $\Delta_m$ denotes the $m$-dimensional probability simplex. We will denote the optimal linear combination in~\eqref{eq:mgda} at each $\x$ as $\g^\dagger(\x)$.

An argument by Lagrangian duality allows us to derive that $\g^\dagger(\x)$ conforms to:
\begin{equation*}
    - \|\g^\dagger(\x)\|^2 = \min_{\|\g\|\leq 1}\min_{i\in[m]} -\g^\top \nabla f_i(\x).
\end{equation*}
Intuitively, $-\g^\dagger$ is the direction where objective functions see
the most common decrease, quantified by $\|\g^\dagger\|^2$. As shown in~\citet[Theorem 2.2]{desideri2012multiple}, a small magnitude of $\|\g^\dagger\|$ indicates
misalignment among the objective function, \emph{i.e.} $\x$ is close to
local Pareto optimality.

Thus, we design the objective function term $\F_1[\rho]$ as
\begin{equation*}
    \F_1[\rho] = \int_{\D} \|\g^\dagger(\x)\|^2 \rho(\d \x),\text{ with }\delta_\rho \F_1(\cdot) = \|\g^\dagger(\cdot)\|^2,
\end{equation*}
pushing $\rho_t$ towards the Pareto front.

\paragraph{Dominance Potential Term.} To ensure the global Pareto optimality of
the minimizer $\rho^*$, we also design a \emph{dominance potential}
term $\F_2[\rho]$ of the following form:
\begin{equation}
    \F_2[\rho] = \int_{\D} \left(\int_{\P} D(\f(\x), \f(\y)) \mu_{\P}(\d \y)\right) \rho(\d \x) ,
    \label{eq:dominanceg}
\end{equation}
where $\mu_\P$ is a predetermined measure on the Pareto front, and the kernel $D(\cdot, \cdot)$ satisfies
\begin{equation}
    D(\f(\x), \f(\y)) = \prod_{i=1}^m  \max\left\{0, f_i(\x) - f_i(\y) \right\}.
    \label{eq:dominance_kernel}
\end{equation}
One should notice this potential term is still linear in $\rho$ with the Fr\'echet derivative given by
\begin{equation}
    \delta_\rho \F_2(\cdot) = \int_{\P} D(\f(\cdot), \f(\y)) \mu_{\P}(\d \y).
    \label{eq:dominance}
\end{equation}

\begin{figure}[t]
    \centering
    \begin{subfigure}[b]{0.48\linewidth}
        \includegraphics[width=\textwidth]{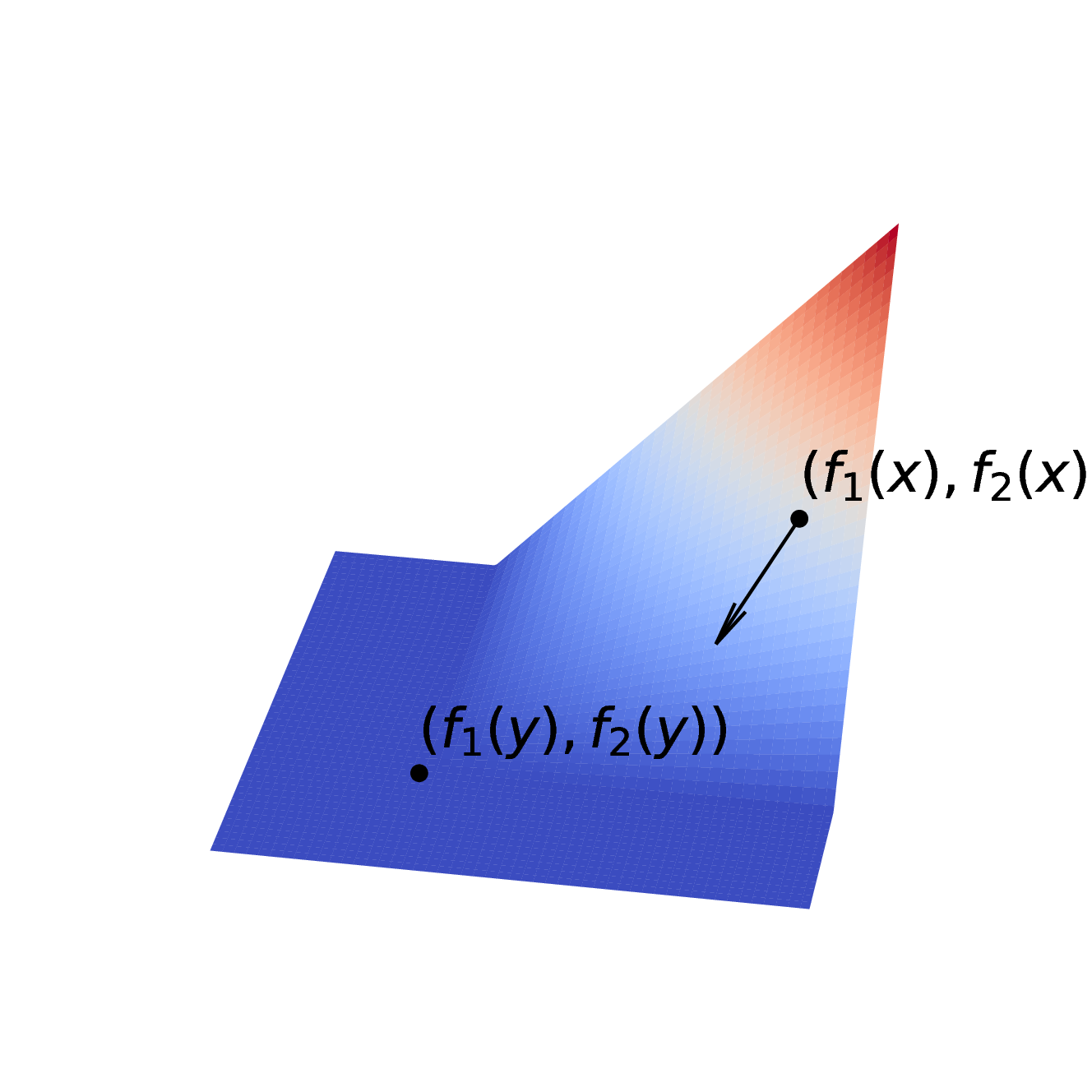}
        \caption{$D(\f(\x),\f(\y))$}
        \label{fig:dominance}
    \end{subfigure}
    \begin{subfigure}[b]{0.48\linewidth}
        \includegraphics[width=\textwidth]{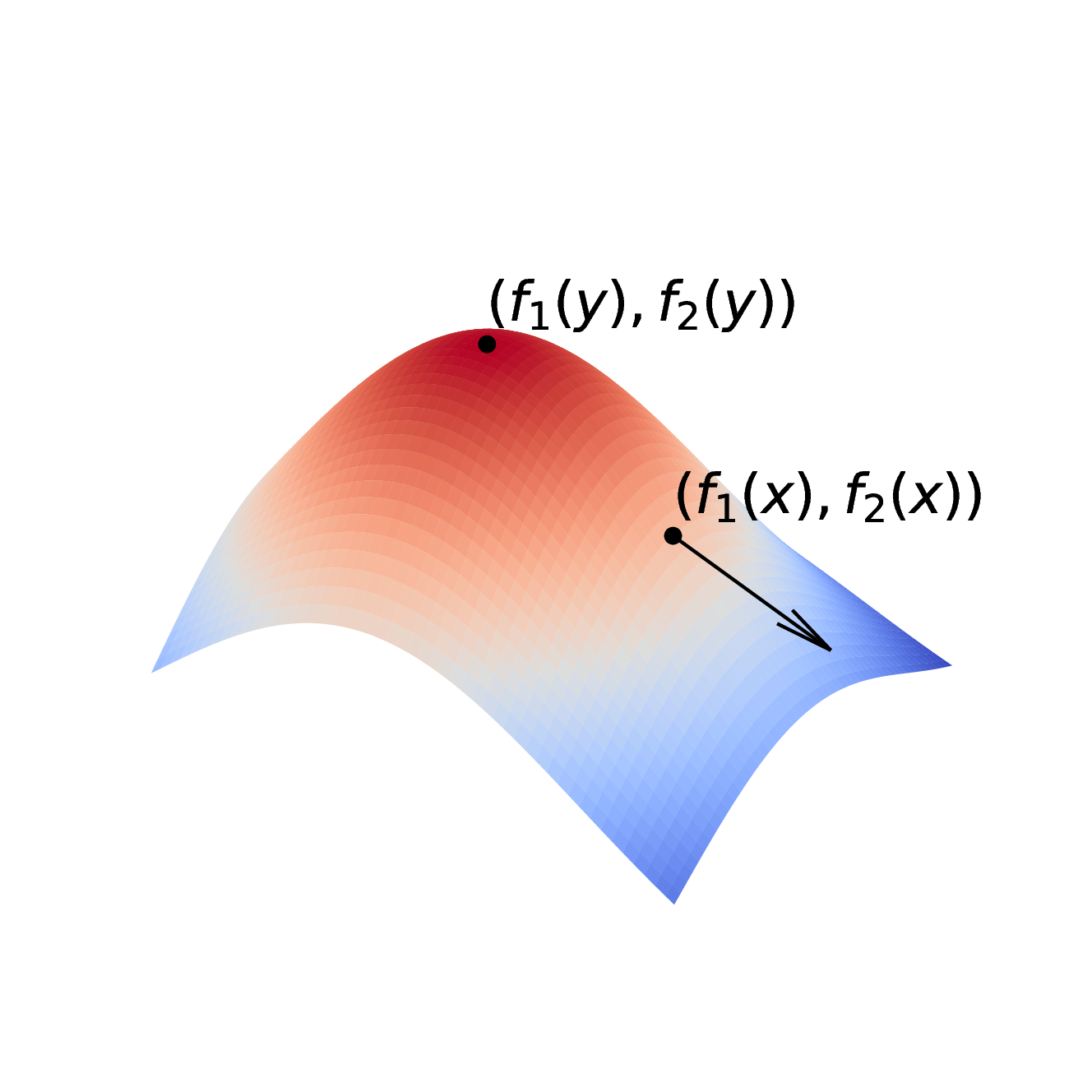}
        \caption{$R(\f(\x),\f(\y))$}
        \label{fig:repulsive}
    \end{subfigure}
    \caption{Illustration of structural potential terms. This visualization
    explains the role of the dominance
    potential $\F_2$ and the repulsive potential $\G$ in a setting with two objective functions ($m=2$). (a) Suppose $\y$ is from $\mu_{\P}(\cdot)$ with corresponding objective function values $\f(\y) = (f_1(\y), f_2(\y))$. When a point $\x$ is introduced, the dominance kernel $D(\cdot, \f(\y))$ acts to shift
    $\x$ out of the region dominated by $\y$. (b) For two samples $\x$ and $\y$ the repulsive kernel $R(\cdot, \f(\y))$ repels the objective function values of $\x$ away from those of $\y$.}
\end{figure}

As shown in Figure~\ref{fig:dominance}, $D(\f(\x), \f(\y))$ is asymmetric and is non-zero if and only if $\x$ is dominated by $\y$, and $\int_{\P} D(\f(\x), \f(\y)) \mu_\P(\d \y)$ indicates how much $\x$ is dominated by the Pareto front $\P$. Ideally, the target distribution $\rho^*$ is expected to have a small value of $\F_2[\rho^*]$. This term is crucial for implementing the birth-death process, which eliminates dominated particles, ensures global Pareto optimality, and accelerates the convergence of our method.

\paragraph{Entropy Term.} We add a negative entropy term $-\H[\rho]$ to the functional
$\Ec[\rho]$ to encourage the diversity of the minimizer $\rho^*$:
\begin{equation*}
    -\H[\rho]= \int_{\D} \rho(\x) \log \rho(\x) \d \x,
\end{equation*}
and its Fr\'echet derivative is given by $\delta_\rho (-\H[\rho]) =\log \rho(\x)+1$. As we will show later, this term effectively injects stochasticity into the evolution of $\rho_t$, thus also encouraging the exploration of the entire Pareto front.


\paragraph{Repulsive Potential Term.}
To further encourage the diversity of the minimizer $\rho^*$ explicitly, we also design two-body \emph{repulsive potential} term $\G[\rho]$ of the following form:
\begin{equation*}
    \G[\rho] = \dfrac{1}{2}\int_{\D \times \D}  \rho(\d \x) R(\f(\x), \f(\y)) \rho(\d \y),
\end{equation*}
with the Fr\'echet derivative being
\begin{equation*}
    \delta_\rho \G[\rho] = \int_{\D} R(\f(\cdot), \f(\y)) \rho(\d \y).
\end{equation*}
Specifically, the repulsive kernel $\R(\cdot,\cdot)$ can adopt various forms depending on the requirements, including the Gaussian potential $R(\x, \y) = \exp(-\|\x -
\y\|^2/\sigma^2)$ or the Coulomb potential $R(\x, \y) = 1/\|\x - \y\|$. As illustrated in Figure~\ref{fig:repulsive}, this term pushes particles in the mass $\rho$ away
from each other, contributing to enhancing the dispersion of the minimizer $\rho^*$.

Combining the above terms, we define the functional $\Ec[\rho]$ as
\begin{equation}
    \Ec[\rho] = \F[\rho] + \beta \G[\rho] - \gamma \H[\rho],
\label{eq:energy}
\end{equation}
where $\F[\rho] = \alpha_1\F_1[\rho] + \alpha_2 \F_2[\rho]$, with $\alpha_1$, $\alpha_2$, $\beta$, and $\gamma$ being hyperparameters.

\subsection{Theoretical Analysis}
\label{sec:theory}

As we are implementing the WFR gradient flow of $\Ec[\rho]$, we have the following overall convergence guarantee:
\begin{theorem}
    Let $\rho_t$ follow the WFR gradient flow of
    $\Ec[\rho]$~\eqref{eq:wfrgf}, then the following decay of the functional value
   $\Ec[\rho_t]$ holds:
    \begin{equation}
        \p_t \Ec[\rho_t] = - \int_{\D} \rho_t \left\|\nabla \delta_\rho \Ec[\rho_t]\right\|^2 + \rho_t \widetilde{\delta_\rho \Ec[\rho_t]}^2 \d \x \leq 0.
    \label{eq:decay}
    \end{equation}
    Furthermore, if $\beta>0$ or $\gamma>0$, the density $\rho_t$ converges to the
    \emph{unique} minimizer $\rho^*$ of $\Ec[\rho]$, as $t\to\infty$.
    \label{thm:convergence}
\end{theorem}
For the special case where the diversity of the mass is only fostered by the entropy $\H[\rho]$, we have the following
exponential convergence guarantee:
\begin{theorem}
    Let $\rho_t$ follow the WFR gradient flow of
    $\Ec[\rho]$ in ~\eqref{eq:wfrgf} with $\beta=0$. The unique minimizer
    $\rho^*$ has the following explicit Gibbs-type expression:
    \begin{equation}
        \rho^* \propto \exp\left(-\dfrac{\alpha_1\|\g^\dagger\|^2 + \alpha_2 \int_{\P} D(\f(\cdot), \f(\y)) \mu_{\P}(\d \y)}{\gamma}\right).
        \label{eq:gibbs}
    \end{equation}
    Assume the initialization satisfies $\inf_{\x\in\D}\rho_0(\x)/\rho^*(\x)\geq e^{-M}$,
    then the following exponential convergence holds:
    \begin{equation}
        \KL(\rho_t|\rho^*)\leq M e^{-\gamma t} + e^{-\gamma t+Me^{-\gamma t}}\KL(\rho_0|\rho^*).
        \label{eq:exponential}
    \end{equation}
    \label{thm:exponential}
\end{theorem}
Proofs are deferred to Appendix~\ref{app:proof}.

\begin{remark}
    One should notice that the exponential convergence discussed above does not require
    the strong convexity of the functional $\Ec[\rho]$, which is often
    necessary for Wasserstein gradient flows (\emph{e.g.} via the log-Sobolev
    inequality~\citep{villani2021topics}). The exclusion of $\G[\rho]$ in Theorem~\ref{thm:exponential} is due to technicalities behind the nonlinearity of $\Ec[\rho]$ and the absence of a closed-form expression of the minimizer $\rho^*$, which we leave for future work.
\end{remark}

\subsection{Interacting Particle Method}

Since probability densities $\rho_t$ as infinite-dimensional objects are
generally difficult to keep track of or optimize, we propose to approximate
$\rho_t$ by a set of $N$
\emph{interacting particles} $\{\x_k\}_{k=1}^n$ as 
\begin{equation*}
    \rho_t \approx \dfrac{1}{N} \sum_{k=1}^n \delta(\x - \x_k),
\end{equation*}
in which $\delta(\cdot)$ denotes the Dirac delta function and each particle
$\x_k$ evolves with the time $t$. This method has been widely used in computational fluid dynamics~\citep{koshizuka2018moving}. We also
employ a straightforward time discretization scheme with a small time step
$\tau$ and use the notation $\x_k^{(\ell)} = \x_k(\ell\tau)$, for $k\in[m]$ and $\ell\geq0$.

As studied by~\citet{gallouet2017jko}, WFR gradient flow can be approximated by the splitting scheme, \emph{i.e.},
alternatively updating with (a) the gradient flow of the Wasserstein metric, and (b) that of the Fisher-Rao metric. Next, we will show that these two updates can be implemented with the overdamped Langevin and birth-death dynamics, respectively.

\paragraph{Overdamped Langevin Dynamics.} 

Plugging~\eqref{eq:energy} into the Wasserstein gradient
flow~\eqref{eq:wgf}, we obtain the following governing equation of $\rho_t$:
\begin{equation*}
    \p_t \rho_t = \nabla \cdot \left(\rho_t \nabla \left(\delta_\rho\F +\delta_\rho \G[\rho_t] \right)\right) + \gamma \Delta \rho_t,
\end{equation*}
which is exactly the Fokker-Planck equation of the following stochastic differential equation (SDE):
\begin{equation*}
        \d \x_t = - \nabla \left(\delta_\rho\F + \delta_\rho\G[\rho_t] \right) \d t + \sqrt{2\gamma} \d \w_t,
\end{equation*}
where $\w_t$ is the standard Brownian motion. Therefore, the update in this step can be easily approximated by the
\emph{overdamped Langevin dynamics} of the particles for each $k\in[N]$:
\begin{equation}
    \x_k^{(\ell+1/2)} = \x_k^{(\ell)} - \dfrac{\tau}{2} \nabla\left(\delta_\rho\F + \delta_\rho\G[\rho_t]\right) + \sqrt{\gamma \tau} \varepsilon_k^{(\ell)},
\label{eq:overdamped}
\end{equation}
where $\varepsilon_k^{(\ell)}$ are sampled independently from a standard
Gaussian. As mentioned before, the stochasticity
is introduced by the entropy term $\H[\rho]$.

\paragraph{Birth-death Dynamics.}

To simulate the Fisher-Rao gradient flow~\eqref{eq:frgf},
we draw inspiration from the birth-death process in the queueing theory and the Gillespie
algorithm for biochemical system simulation~\citep{gillespie2007stochastic}.
Specifically, we reformulate~\eqref{eq:frgf} as 
\begin{equation}
    \p_t \log \rho_t = - \widetilde{\delta_\rho \Ec[\rho_t]} := - \Lambda_t,
\label{eq:logfrgf}
\end{equation}
where $\Lambda_t$ is the \emph{instantaneous birth-death rate} function. Using the
forward Euler scheme,~\eqref{eq:logfrgf} can be approximated by
$\rho_{t+\tau/2}\approx \rho_{t} \exp(-\Lambda_t\tau/2)$, \emph{i.e.} $\rho_t$
should increase by $\exp(-\Lambda_t\tau/2) - 1$ if $\Lambda_t<0$, and if
$\Lambda_t>0$, it should decrease by $1-\exp(-\Lambda_t\tau/2)$ . 

In the context of our interacting particle method, we compute the instantaneous birth-death rate
function $\Lambda_{(\ell+1/2)\tau}$ for each particle $\x_k^{(\ell+1/2)}$. This
is done by approximating the expectation as
\begin{equation}
    \begin{aligned}
        &\Lambda_{(\ell+1/2)\tau}(\x_k^{(\ell+1/2)}) = \delta_\rho \Ec[\rho_t](\x_k^{(\ell+1/2)}) - \E{\delta_\rho\Ec[\rho_t]}{\rho_t}\\
        &\approx \delta_\rho \Ec[\rho_t](\x_k^{(\ell+1/2)}) - \dfrac{1}{N} \sum_{k'=1}^N \delta_\rho \Ec[\rho_t](\x_{k'}^{(\ell+1/2)}).
    \end{aligned}
\label{eq:lambda}
\end{equation}
Then for each particle $\x_k^{(\ell+1/2)}$, depending on the sign of $\Lambda_{(\ell+1/2)\tau}$, we either duplicate it with probability
$\exp(-\Lambda_{(\ell+1/2)\tau}(\x_k^{(\ell+1/2)})\tau/2) - 1$ or remove it with
probability $1 - \exp(-\Lambda_{(\ell+1/2)\tau}(\x_k^{(\ell+1/2)})\tau/2)$. To compensate for the change in the total mass, we also randomly remove
(or duplicate) a particle from the entire population whenever a duplication
(or removal) occurs. As demonstrated in Proposition 5.1 of~\citet{lu2019accelerating}, the above birth-death
implementation converges to the original Fisher-Rao gradient
flow~\eqref{eq:frgf} when $\tau\to0$ and the number of particles $N\to\infty$.

In conclusion, we summarize our proposed interacting particle method in
Algorithm~\ref{alg:moo} and refer to \emph{Particle-WFR} in below. Due to space limit, we refer to
Appendix~\ref{app:implementation} for additional implementation details.

\IncMargin{1.5em}
\begin{algorithm}[t]
\caption{Particle-WFR Method for MOO}
\label{alg:moo}
\Indm  
\KwData{Objective functions $f_i$, $i\in[m]$, feasible region $\D$, time step $\tau$, number of particles $N$, number of iterations $T$}
\KwResult{Evolved population $\{\x_k^{(T)}\}_{k=1}^N$}
\Indp
Randomly initialize $\{\x_k^{(0)}\}_{k=1}^N\subset\D$\;
\For{$\ell=1$ \KwTo $T$}{
    \tcp{Overdamped Langevin dynamics}
    \For{$k=1$ \KwTo $N$}{
        Update $\x_k^{(\ell+1/2)}$ from $\x_k^{(\ell)}$ by~\eqref{eq:overdamped}\;
    }
    \tcp{Birth-death dynamics}
    \For{$k=1$ \KwTo $N$}{
        $\x_{k}^{(\ell+1)}\leftarrow\x_{k}^{(\ell+1/2)}$\;
        Compute $\lambda = \Lambda_{(\ell+1/2)\tau}(\x_k^{(\ell+1/2)})$ by~\eqref{eq:lambda}\;
        $k'\sim\mathrm{Unif}([N])$, $\eta\sim\mathrm{Unif}([0,1])$\;
        \If{$\eta < |1 - \exp(-\lambda\tau/2)|$}{
            \eIf{$\lambda < 0$}{
            $\x_{k'}^{(\ell+1)}\leftarrow\x_{k}^{(\ell+1/2)}$\;
            }{
            $\x_{k}^{(\ell+1)}\leftarrow\x_{k'}^{(\ell+1/2)}$\;
            }
        }
    }
}
\end{algorithm}
\DecMargin{1.5em}

\begin{remark}
    While most existing methods adopt the MGDA gradient $\g^\dagger(\x)$ for updates, $\|\g^\dagger(\x)\|=0$ only indicates local Pareto optimality, not a global guarantee. In our method, local Pareto optimal points are eliminated from the population by the birth-death dynamics acting on the dominance potential $\F_2[\rho]$ and replaced by other particles with lower dominance potential, thus ensuring global Pareto optimality.
\end{remark}

\section{EXPERIMENT RESULTS}
\label{sec:exp}

\begin{figure*}[!htb]
    \centering
    \begin{subfigure}[b]{0.3\linewidth}
        \includegraphics[width=\textwidth]{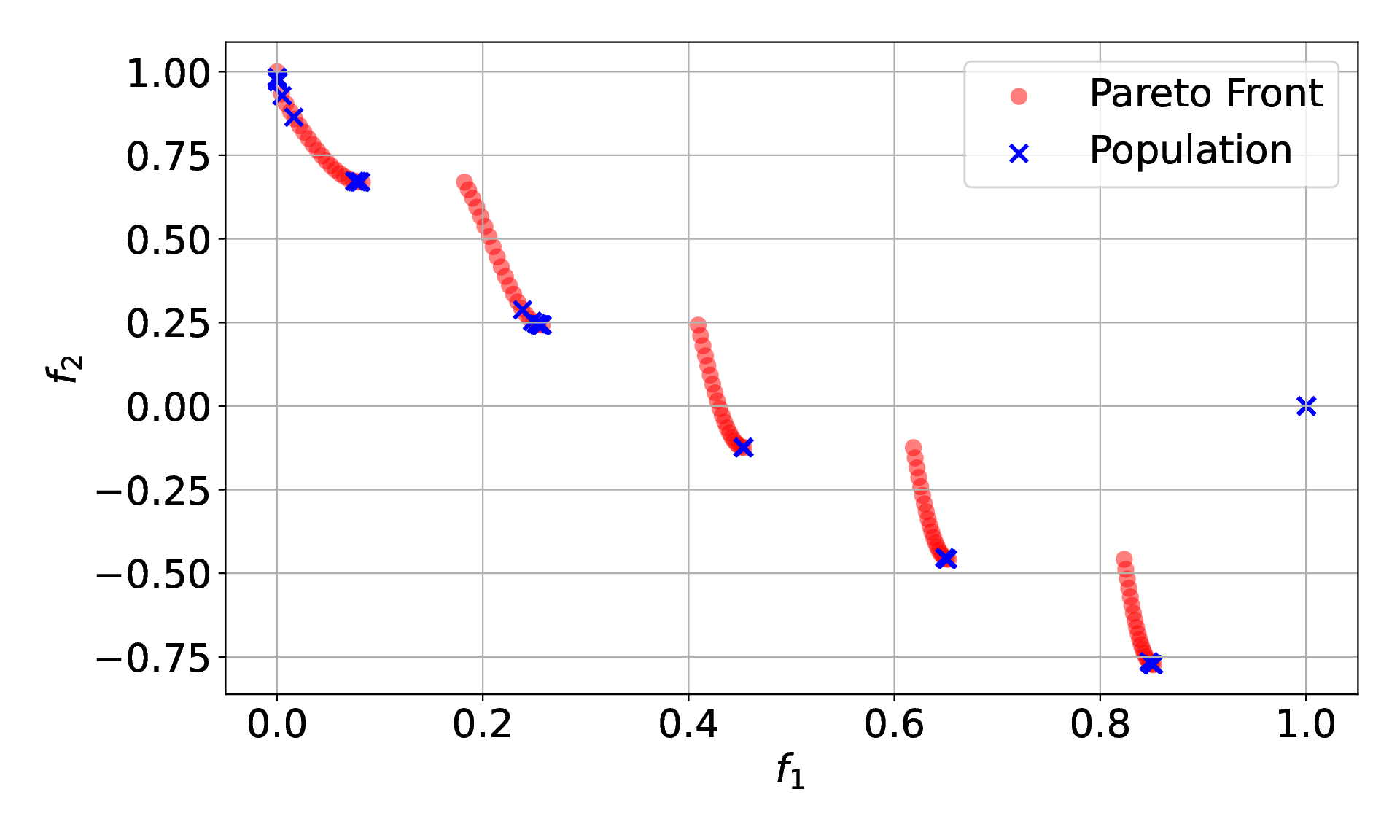}
        \caption{Weighted Sum}
        \label{fig:zdt3_linear}
    \end{subfigure}
    \begin{subfigure}[b]{0.3\linewidth}
        \includegraphics[width=\textwidth]{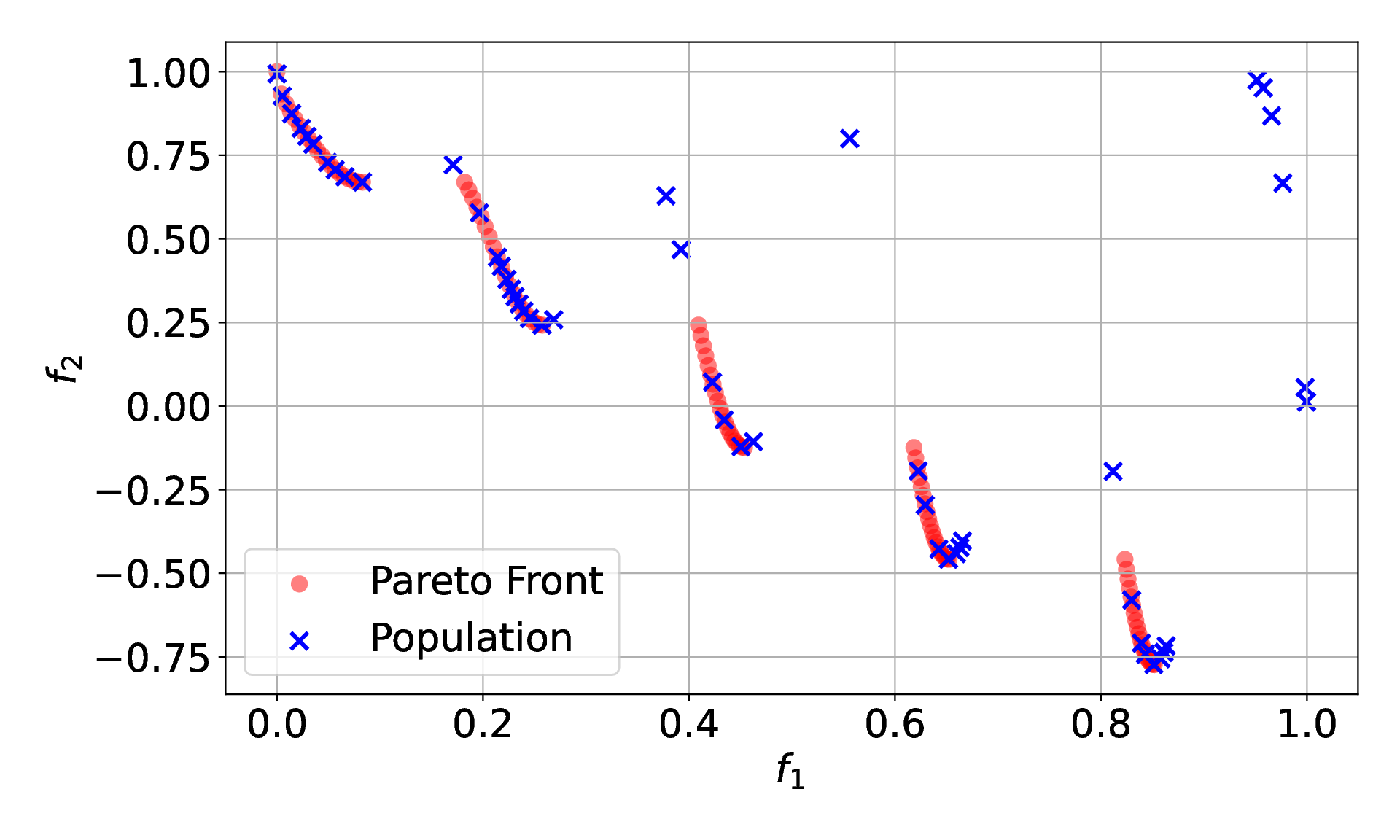}
        \caption{MOO-SVGD}
        \label{fig:zdt3_moosvgd}
    \end{subfigure}
    \begin{subfigure}[b]{0.3\linewidth}
        \includegraphics[width=\textwidth]{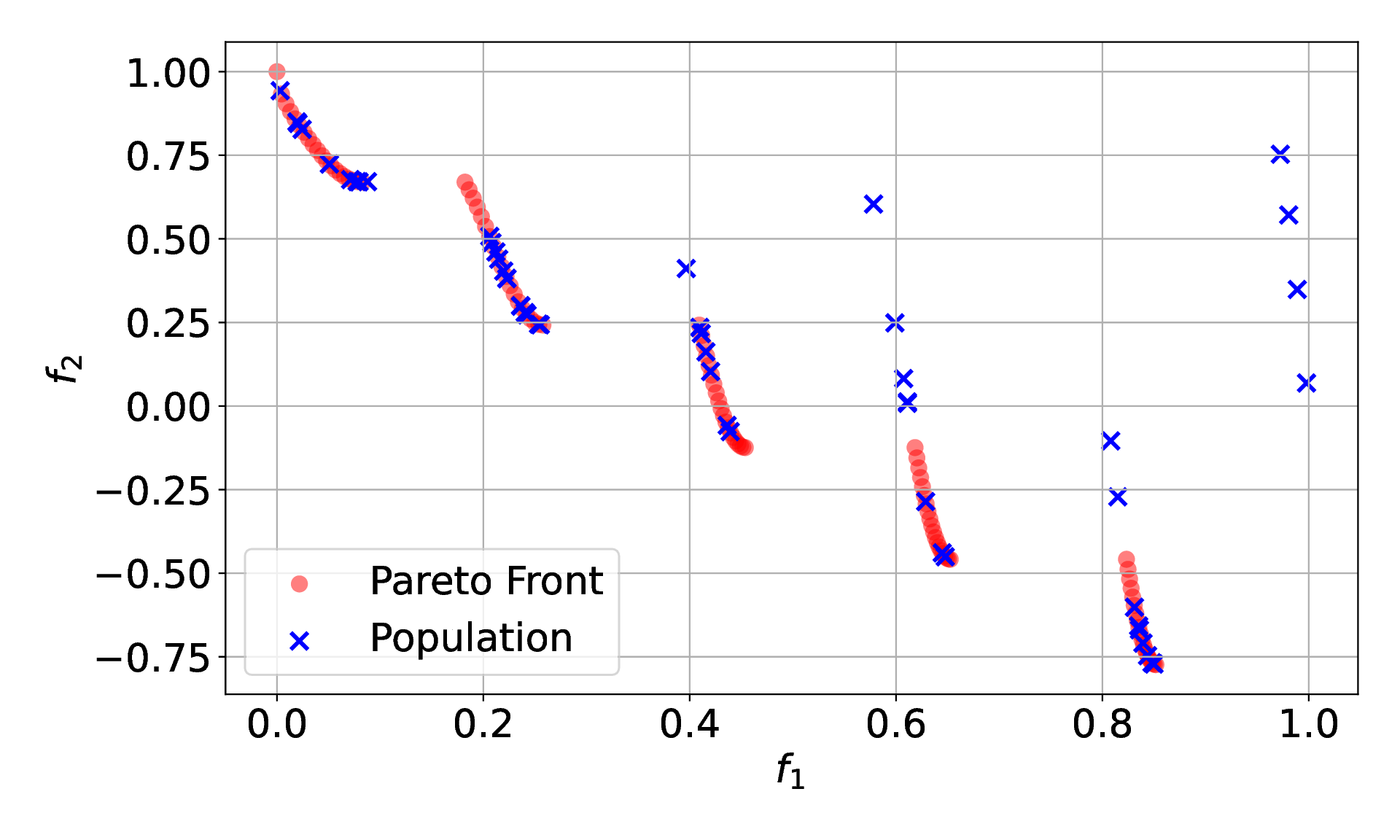}
        \caption{MOO-LD}
        \label{fig:zdt3_moold}
    \end{subfigure}

    \begin{subfigure}[b]{0.3\linewidth}
        \includegraphics[width=\textwidth]{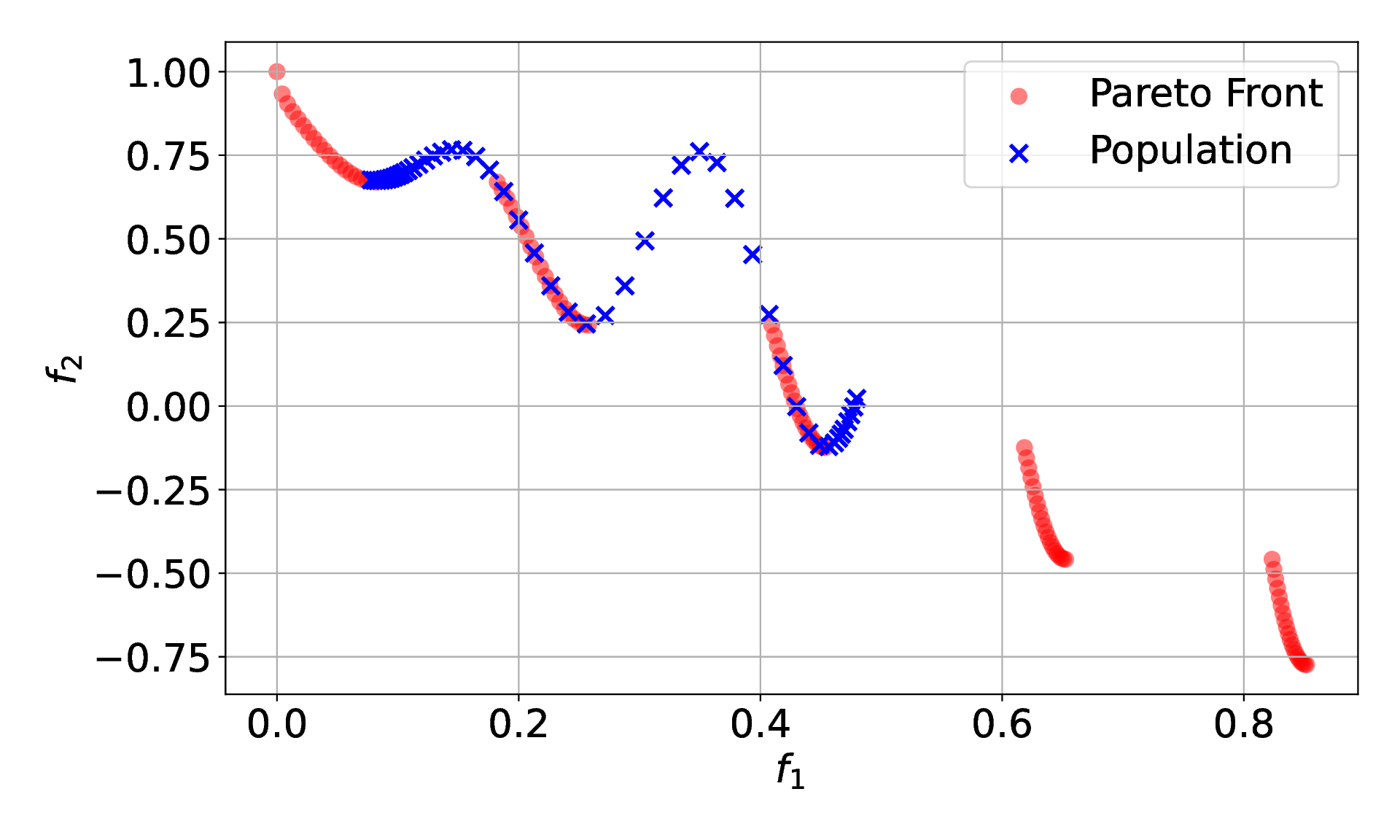}
        \caption{COSMOS}
        \label{fig:zdt3_cosmos}
    \end{subfigure}
    \begin{subfigure}[b]{0.3\linewidth}
        \includegraphics[width=\textwidth]{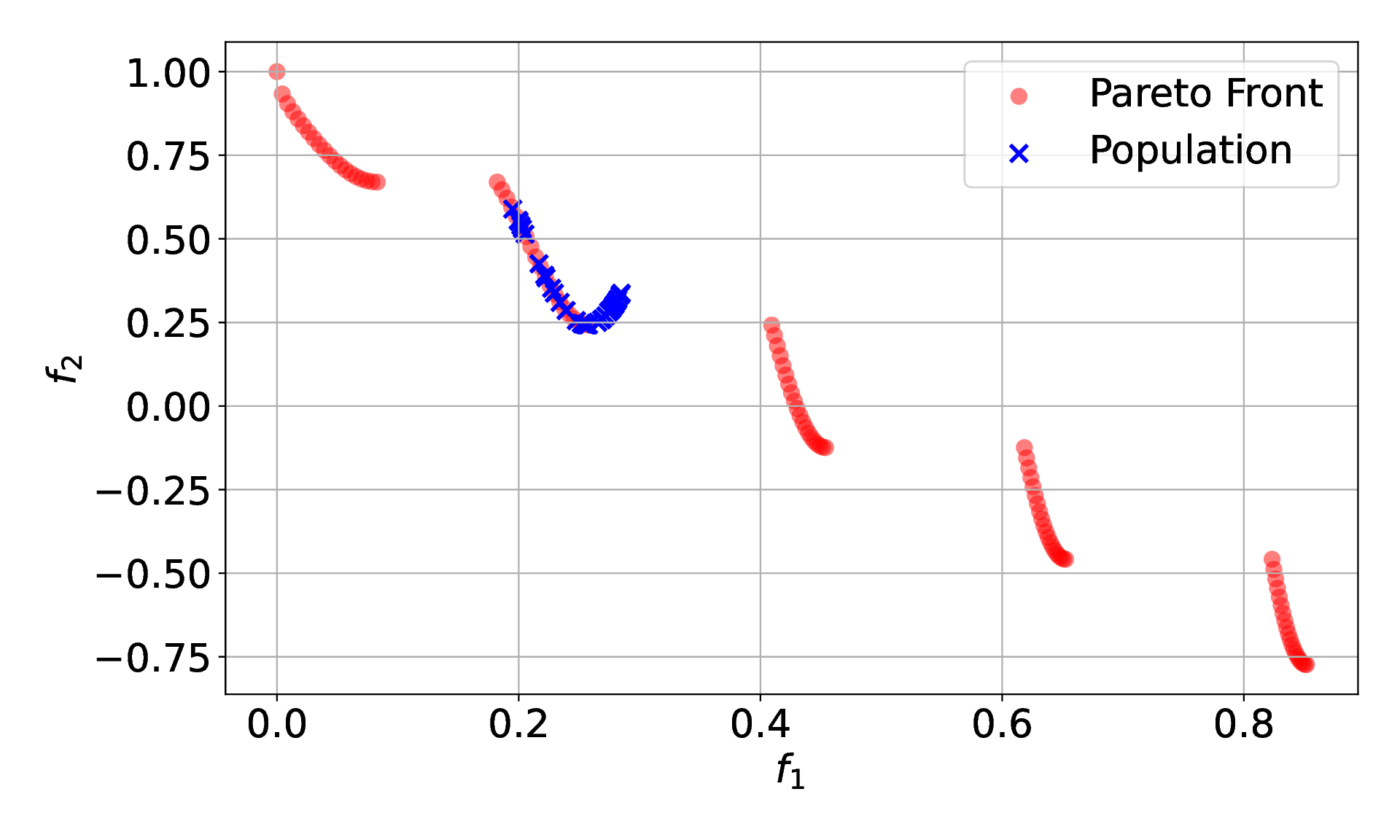}
        \caption{GMOOAR-HV}
        \label{fig:zdt3_argmo}
    \end{subfigure}
    \begin{subfigure}[b]{0.3\linewidth}
        \includegraphics[width=\textwidth]{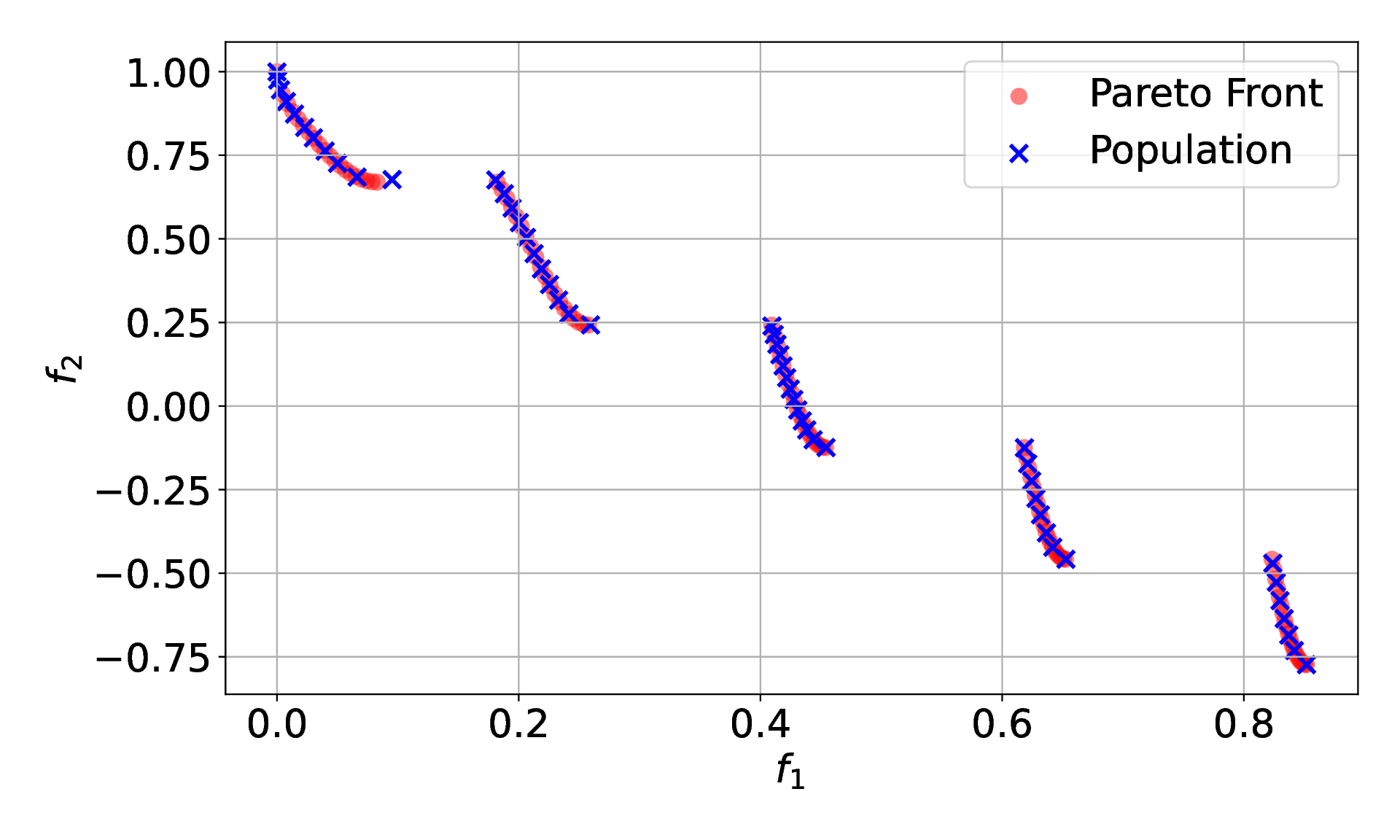}
        \caption{\bf Particle-WFR}
        \label{fig:zdt3_wfr}
    \end{subfigure}
    \caption{Performance comparison of different methods on the ZDT3 problem. The Pareto front is shown in red, and the solutions found by different methods are shown in blue. Our method (Particle-WFT) perfectly captures the complicated geometry of the Pareto front.}
    \label{fig:zdt3}
\end{figure*}

\begin{figure*}[!htb]
    \centering
    \begin{subfigure}[b]{0.16\linewidth}
        \includegraphics[width=\textwidth]{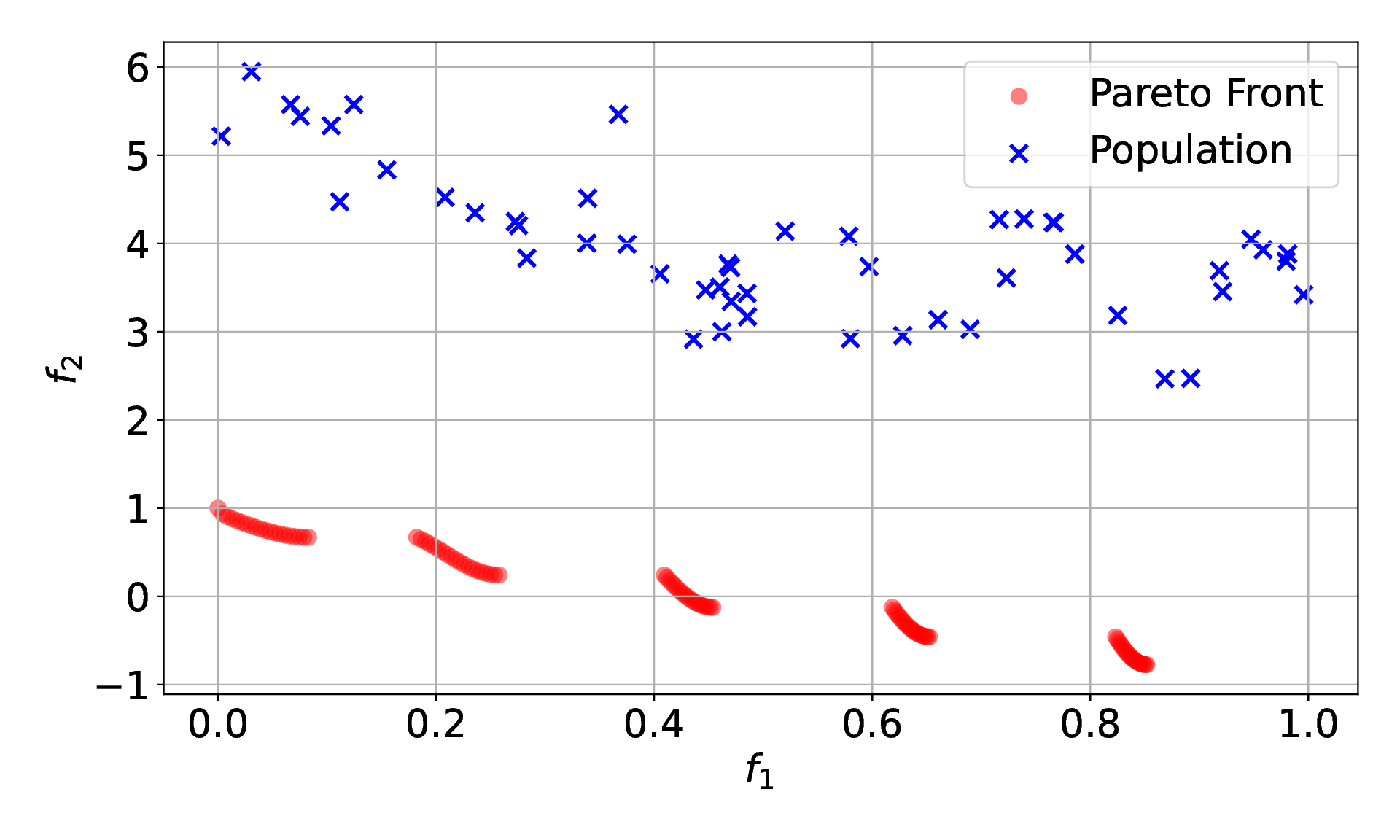}
        \caption{Epoch 0}
        \label{fig:zdt3_particle_0}
    \end{subfigure}
    \begin{subfigure}[b]{0.16\linewidth}
        \includegraphics[width=\textwidth]{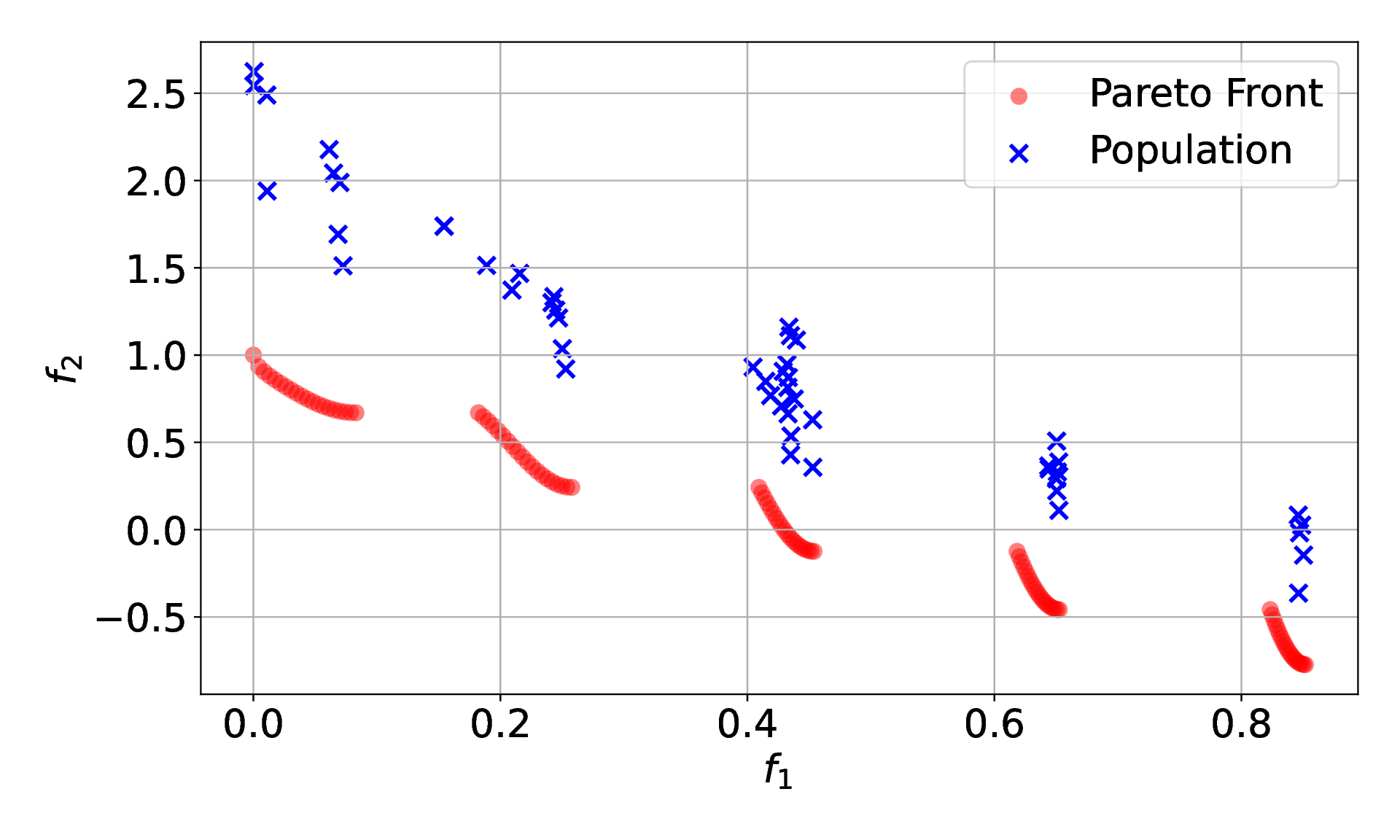}
        \caption{Epoch 1000}
        \label{fig:zdt3_particle_1000}
    \end{subfigure}
    \begin{subfigure}[b]{0.16\linewidth}
        \includegraphics[width=\textwidth]{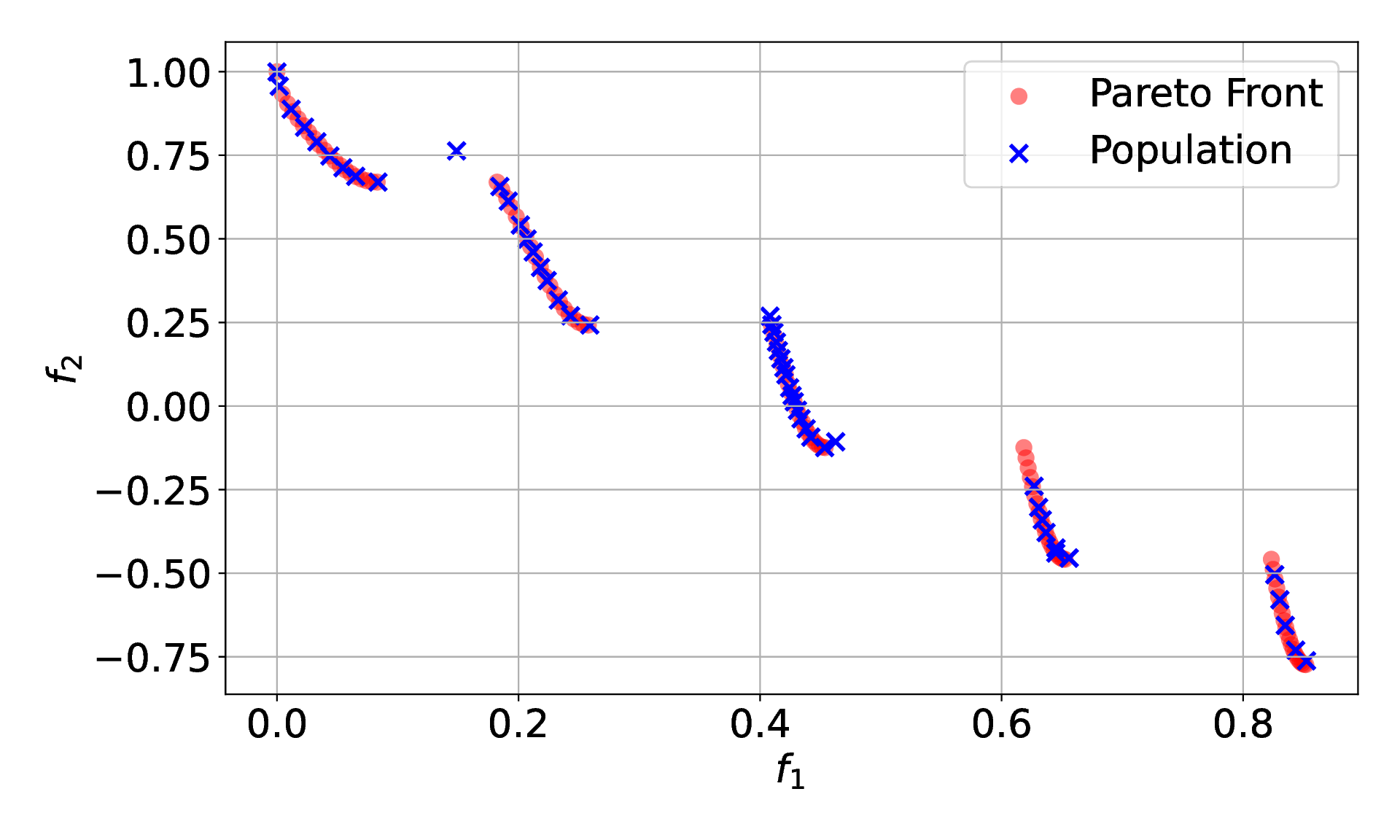}
        \caption{Epoch 2000}
        \label{fig:zdt3_particle_2000}
    \end{subfigure}
    \begin{subfigure}[b]{0.16\linewidth}
        \includegraphics[width=\textwidth]{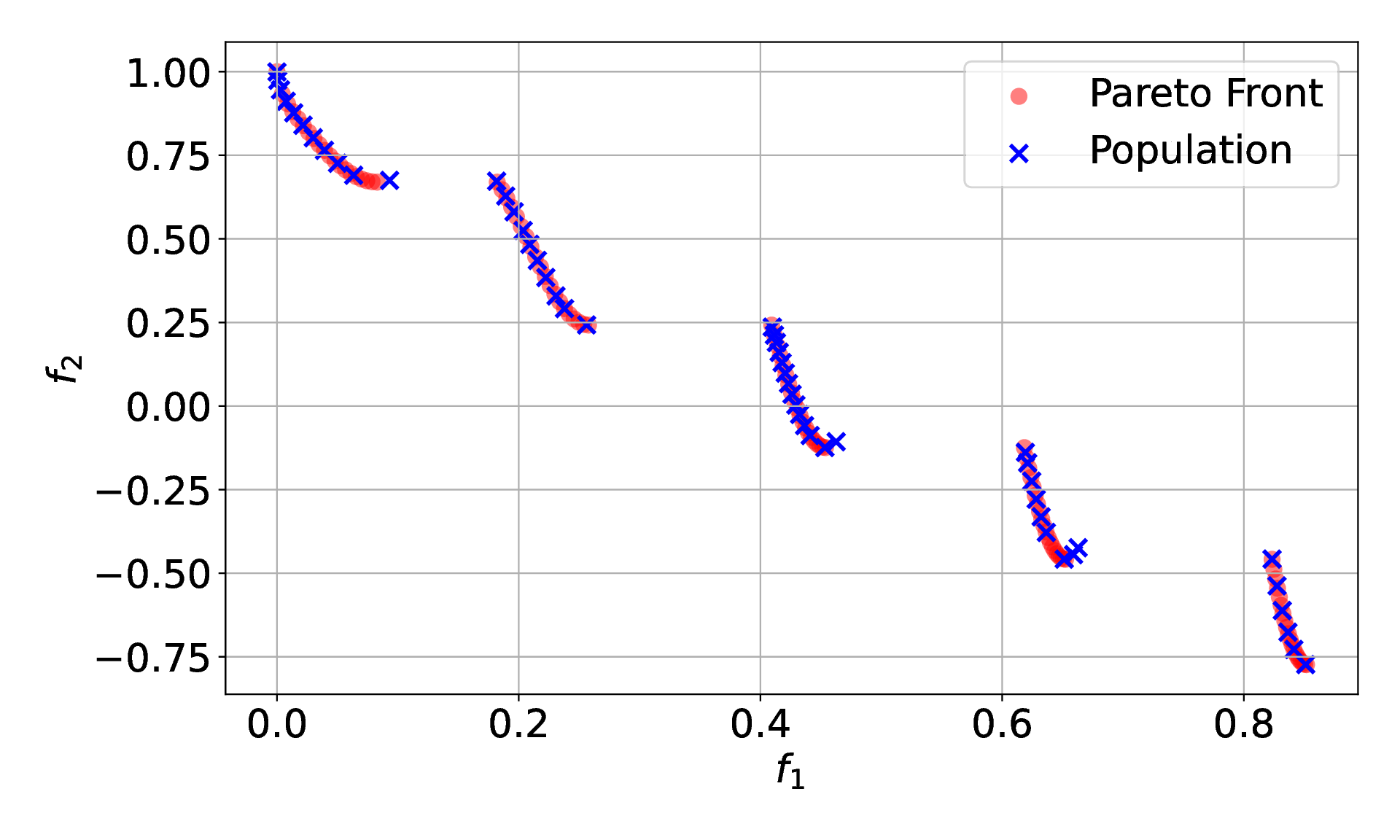}
        \caption{Epoch 3000}
        \label{fig:zdt3_particle_3000}
    \end{subfigure}
    \begin{subfigure}[b]{0.16\linewidth}
        \includegraphics[width=\textwidth]{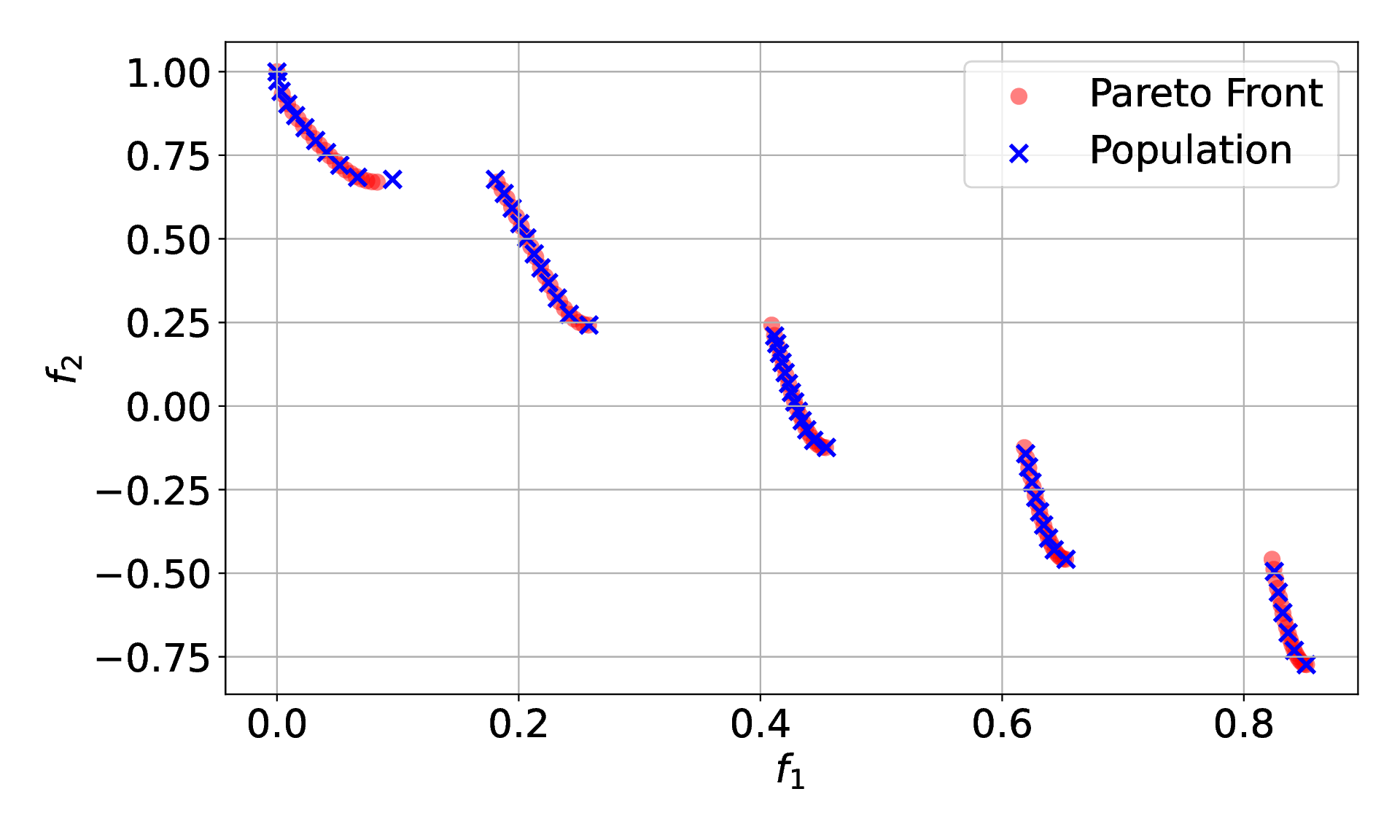}
        \caption{Epoch 4000}
        \label{fig:zdt3_particle_4000}
    \end{subfigure}
    \begin{subfigure}[b]{0.16\linewidth}
        \includegraphics[width=\textwidth]{images/zdt3_particle_5000.png}
        \caption{Epoch 5000}
        \label{fig:zdt3_particle_5000}
    \end{subfigure}
    \caption{Evolution of the particle population by Particle-WFR on the ZDT3 problem. The Pareto front is shown in red, and the current population is shown in blue.}
    \label{fig:zdt3_particle_evo}
\end{figure*}

In this section, we conduct experiments on both synthetic and real-world datasets to evaluate the performance of our method. We compare our method with several recent state-of-the-art gradient-based methods, including PHN-LS and PHN-EPO~\citep{navon2020learning}, COSMOS~\citep{ruchte2021scalable}, MOO-SVGD and MOO-LD\footnote{
Our implementation of MOO-SVGD and MOO-LD made several necessary modifications to its open-source version. Additionally, we utilized a different number of epochs, leading to results that are slightly distinct from those presented in~\citet{liu2021profiling}.}~\citep{liu2021profiling}, and GMOOAR-HV and GMOOAR-U~\citep{chen2022multi}.

\subsection{ZDT3 Problem}

We first consider the ZDT3 problem~\citep{zitzler2000comparison}, which has also been studied by~\citet{custodio2011direct, liu2021profiling}. The ZDT3 problem is a 30-dimensional two-objective optimization problem ($d=30$, $m=2$), with the closed-form formula in Appendix~\ref{app:zdt3}. Unlike ZDT1 and ZDT2 problems with continuous smooth Pareto fronts that can be easily handled Using the naive weighted sum method~\citep{boyd2004convex} or preference vectors-based penalty methods~\citep{lin2019pareto}, the ZDT3 problem presents a non-convex consisting of five disconnected segments. This complicated geometry makes it particularly challenging for gradient-based MOO methods. Due to the page limit, the experiments and discussions on the ZDT1 and ZDT2 problems are provided in Appendix~\ref{app:zdt1} and~\ref{app:zdt2}, respectively.

\begin{figure*}[!htb]
    \centering
    \begin{subfigure}[b]{0.32\linewidth}
        \includegraphics[width=\textwidth]{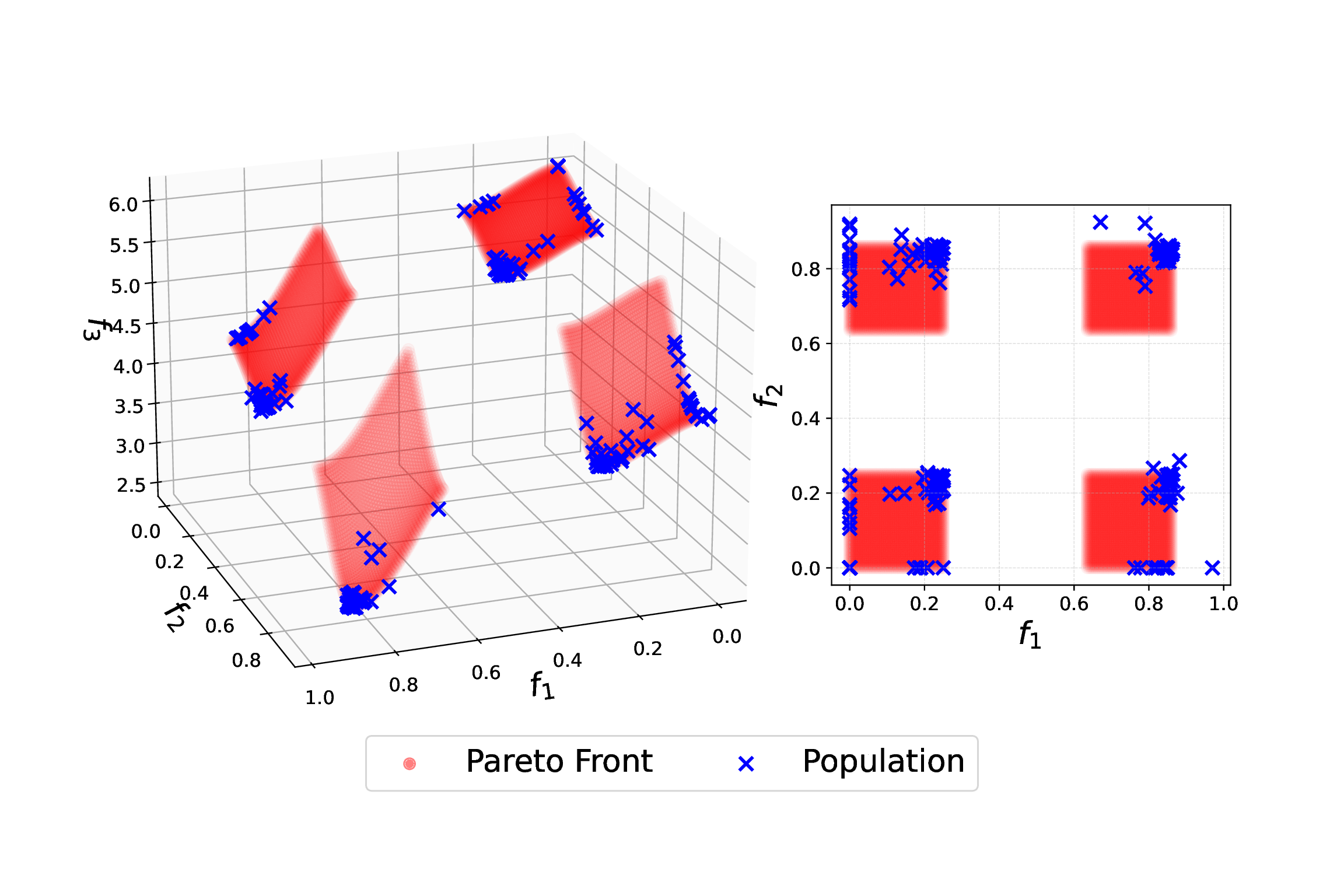}
        \caption{Weighted Sum}
        \label{fig:dtlz7_linear}
    \end{subfigure}
    \begin{subfigure}[b]{0.32\linewidth}
        \includegraphics[width=\textwidth]{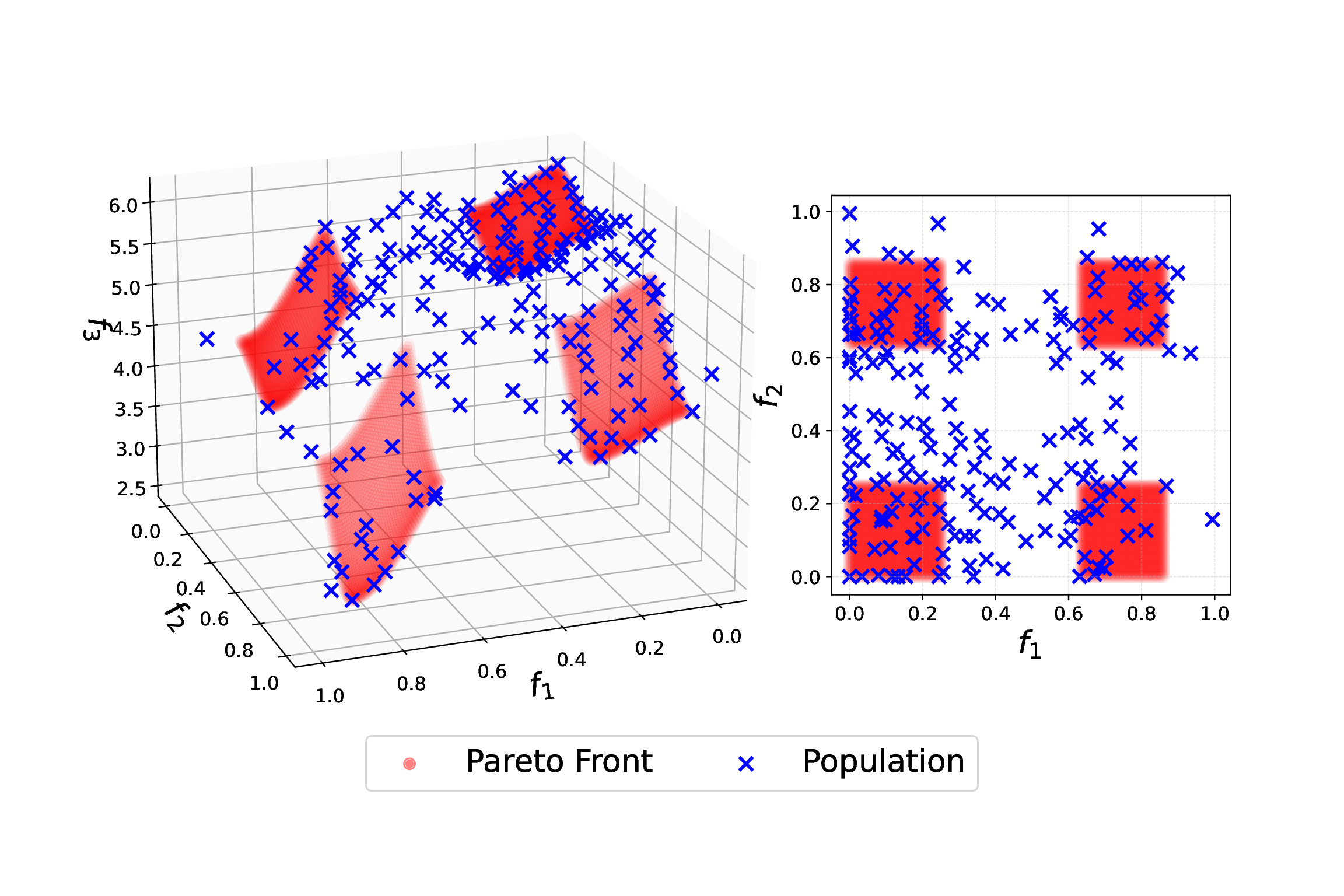}
        \caption{MOO-SVGD}
        \label{fig:dtlz7_moosvgd}
    \end{subfigure}
    \begin{subfigure}[b]{0.32\linewidth}
        \includegraphics[width=\textwidth]{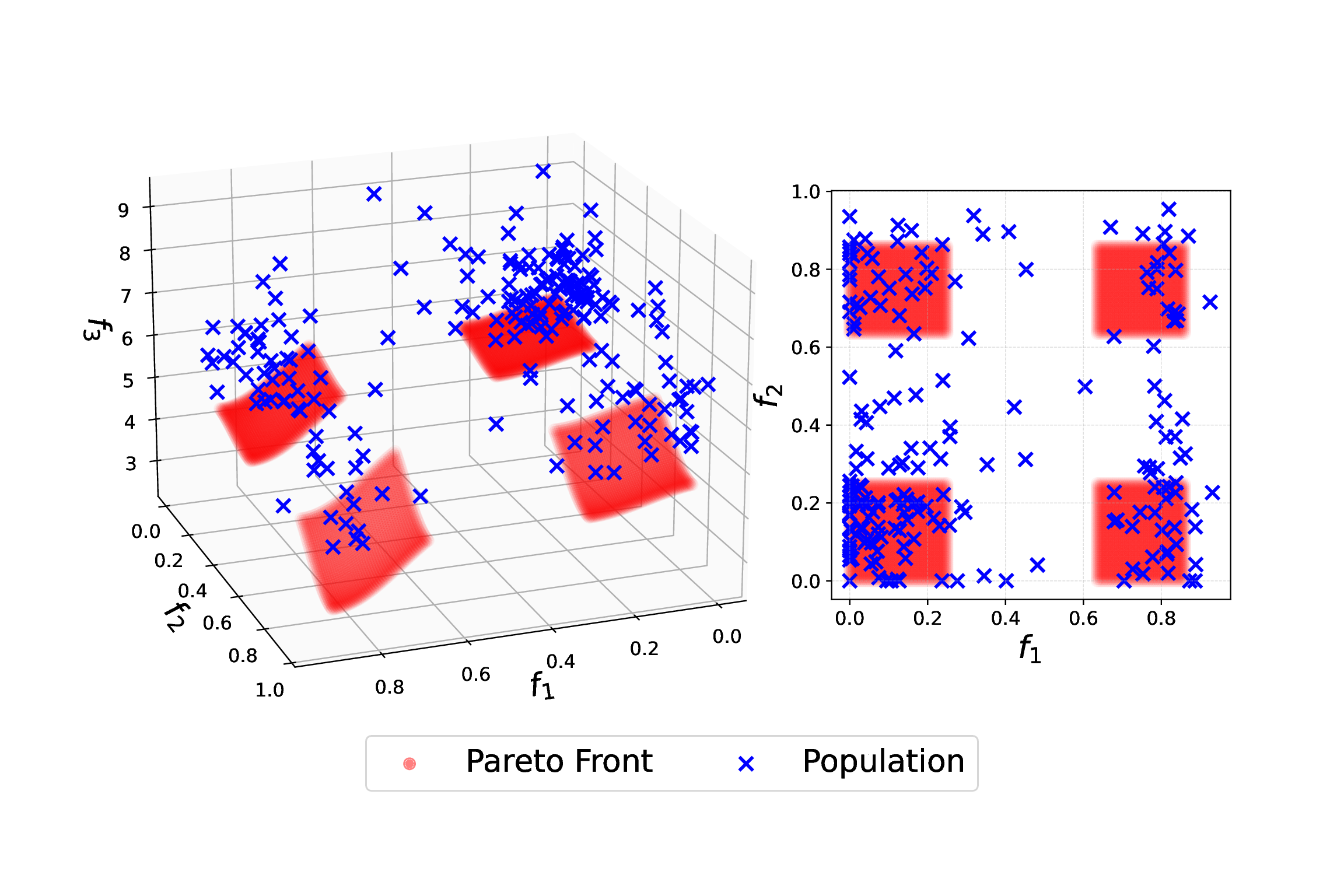}
        \caption{MOO-LD}
        \label{fig:dtlz7_moold}
    \end{subfigure}

    \begin{subfigure}[b]{0.32\linewidth}
        \includegraphics[width=\textwidth]{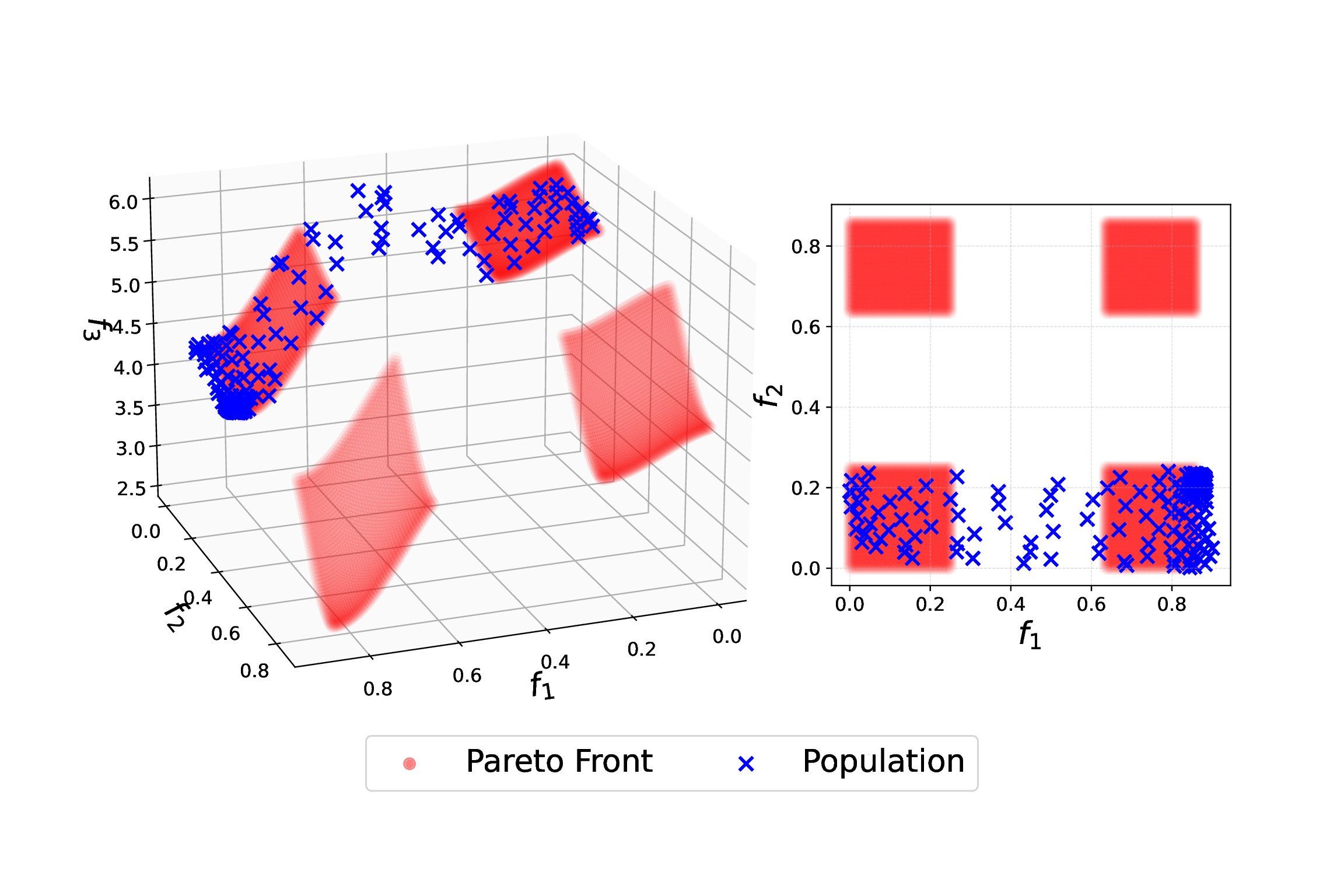}
        \caption{COSMOS}
        \label{fig:dtlz7_cosmos}
    \end{subfigure}
    \begin{subfigure}[b]{0.32\linewidth}
        \includegraphics[width=\textwidth]{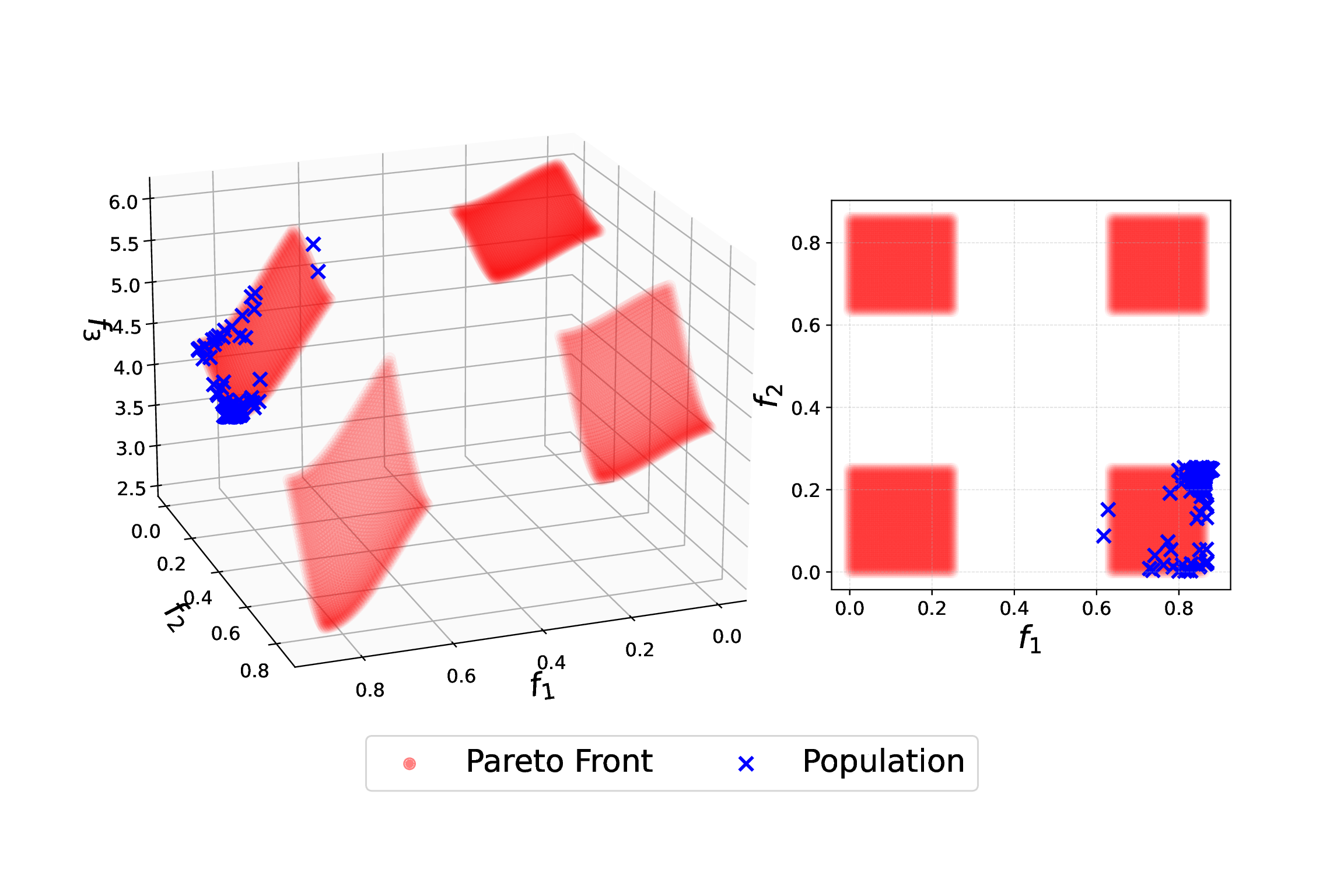}
        \caption{GMOOAR-HV}
        \label{fig:dtlz7_argmo}
    \end{subfigure}
    \begin{subfigure}[b]{0.32\linewidth}
        \includegraphics[width=\textwidth]{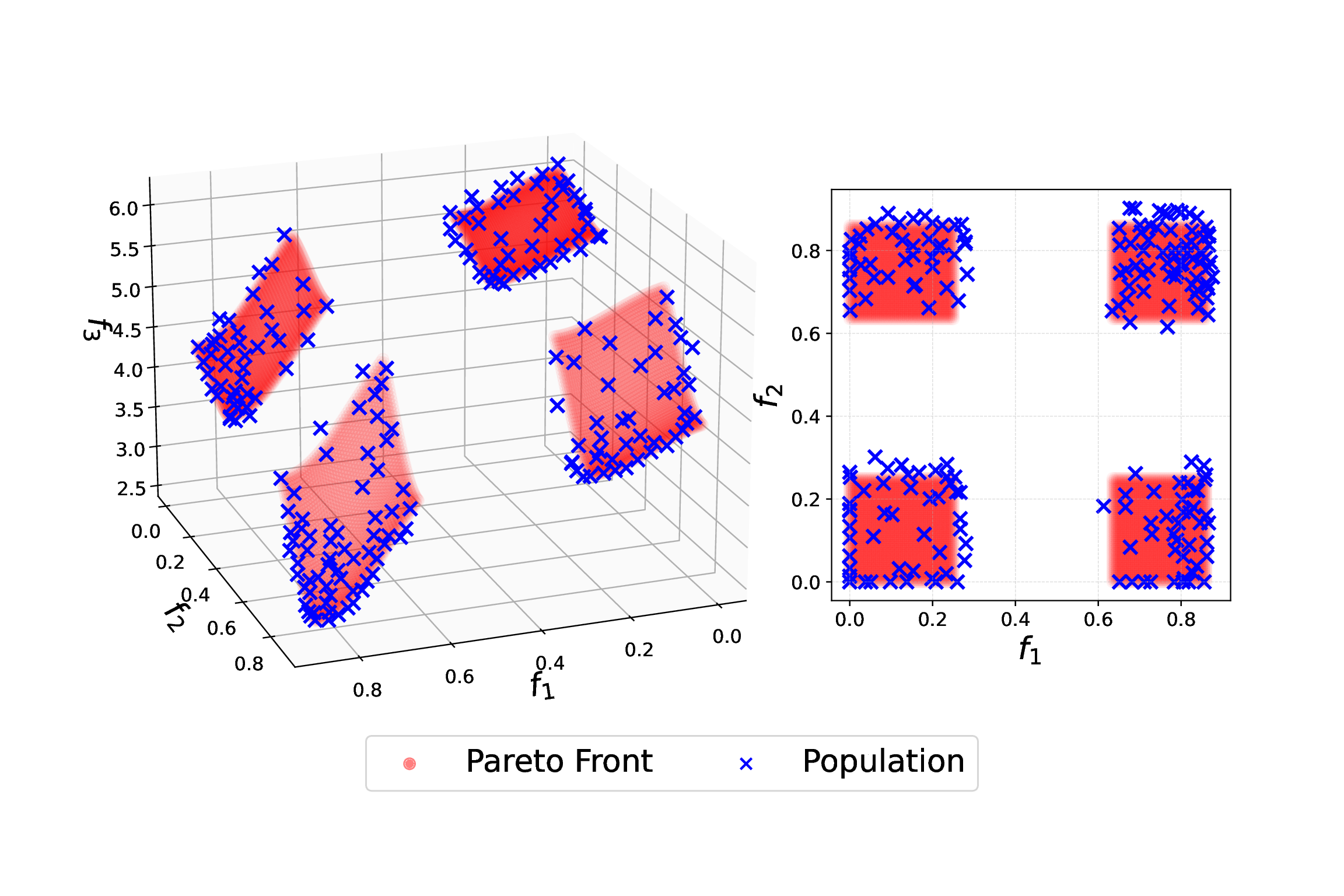}
        \caption{\bf Particle-WFR}
        \label{fig:dtlz7_wfr}
    \end{subfigure}
    \caption{Performance comparison of different methods on the DTLZ7 problem. The Pareto front is shown in red, and the solutions found by different methods are shown in blue. Alongside the 3D visualization, a bird's-eye perspective is also provided for each problem. Our method (Particle-WFR) achieves the best coverage of the Pareto front across all methods.}
    \label{fig:dtlz7}
\end{figure*}

Figure~\ref{fig:zdt3} compares the performance of our method with five methods: the naive weighted sum method, COSMOS, MOO-SVGD, and GMOOAR-HV\footnote{We are only comparing with GMOOAR-HV in the ZDT3 and DTLZ7 problem because the result of GMOOAR-U is similar for these two examples.}. For a fair comparison, we use the same number of particles (or equivalently, uniformly distributed preference vectors or test rays) $N=50$ and run all methods over 5000 iterations (or equivalently, epochs). As observed, the weighted sum method only identifies the convex hull of the Pareto front, aligning with the theoretical analysis of~\citet{boyd2004convex}. Due to the inherent continuity of neural networks, the solutions obtained by COSMOS delineate only a small portion of the image manifold of $\f(\x)$ and consequently interpolate different Pareto front segments. The optimization of preference vectors in GMOOAR-HV also fails in this case, resulting in limited coverage of the Pareto front. Both MOO-SVGD and MOO-LD achieve good coverage of the Pareto front but fall short of discerning and eliminating local Pareto optimal solutions. In contrast, our method outperforms all others, achieving a comprehensive and accurate coverage of the Pareto front.

We also showcase the evolution of the particle population by our method Particle-WFR in Figure~\ref{fig:zdt3_particle_evo}. The particles are initially dispersed within the feasible region $\D$. Driven by the objective function potential term $\F_1[\rho]$, the particles are gradually attracted to the Pareto front, during which the birth-death dynamics systematically purge any particles that fall into local Pareto optimal regions. Roughly after epoch 2000, most of the particles are concentrated on the Pareto front, and the subsequent epochs focus on fine-tuning until a balance between the repulsive potential term $\G[\rho]$ and the dominance potential term $\F_2[\rho]$ is reached, ensuring optimal coverage of the Pareto front.

\subsection{DTLZ7 Problem}

Moving to three-objective optimization, we consider the DTLZ7 problem~\citep{deb2002scalable}, which is another 30-dimensional optimization problem with the closed-form description in Appendix~\ref{app:dtlz7}.
As highlighted by~\citet{li2013shift}, the DTLZ7 problem features a mixed, disconnected, and multimodal Pareto front, making it one of the most challenging problems within the DTLZ test suite when compared with the three-objective MaF1 and DTLZ2 problems used in~\citet{liu2021profiling,chen2022multi}. 

The solutions obtained by the naive weighted sum method, MOO-SVGD, MOO-LD, COSMOS, GMOOAR-HV, and Particle-WFR, are shown in Figure~\ref{fig:dtlz7}. We fix the number of particles $N=200$ and the number of iterations as 3000. Similar phenomena as in the ZDT3 problem are observed: the solutions obtained by the weighted sum method cluster around the corners of the Pareto front. In contrast, the solutions obtained by COSMOS and GMOOAR-HV only span a limited portion of the Pareto front. Although MOO-SVGD can explore the entire Pareto front, a majority of the solutions end up in the local Pareto optimal regions. Relying on the noise for exploration, MOO-LD fails to balance the diversity and the convergence of the particle population, leading to a sub-optimal coverage of the Pareto front. In contrast, our method, Particle-WFR, achieves the best coverage of the Pareto front across all methods. 

\subsection{MSLR-WEB10K Dataset}
\label{sec:ltr}

\begin{figure*}[!htb]
    \centering
    \begin{subfigure}[b]{0.31\linewidth}
        \includegraphics[width=\textwidth]{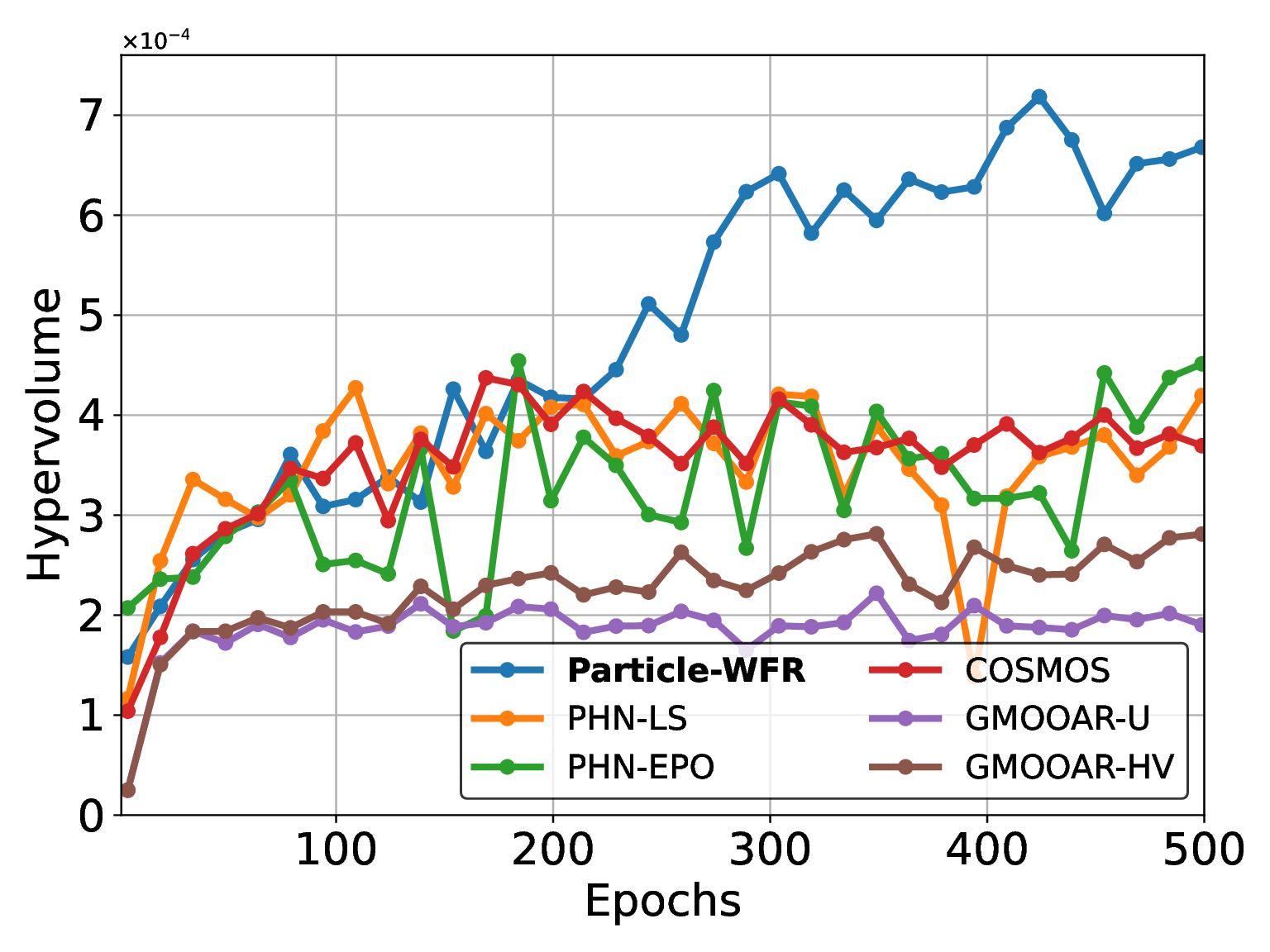}
        \caption{$N=8$}
        \label{fig:mslr_8}
    \end{subfigure}
    \begin{subfigure}[b]{0.31\linewidth}
        \includegraphics[width=\textwidth]{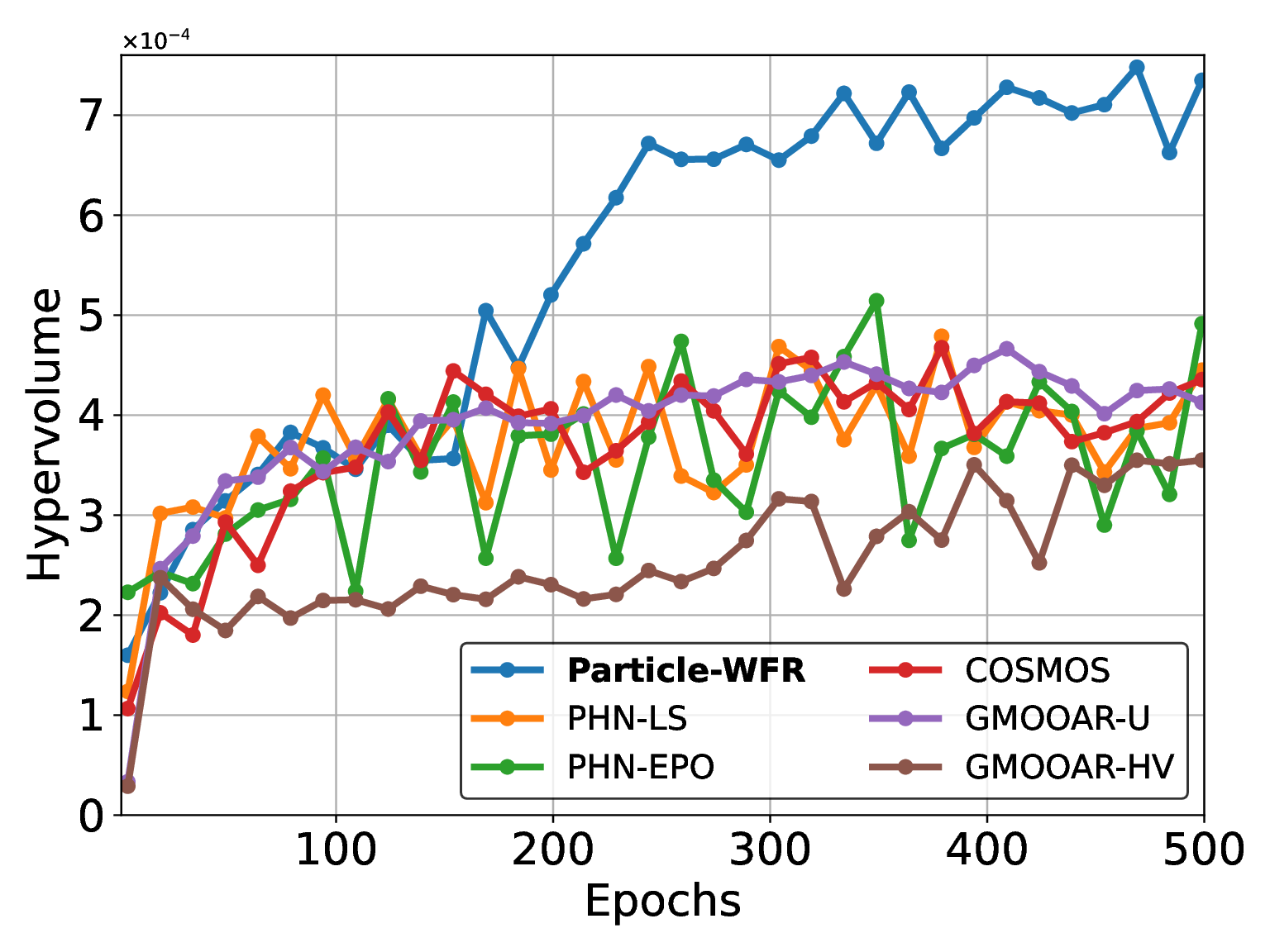}
        \caption{$N=12$}
        \label{fig:mslr_12}
    \end{subfigure}
    \begin{subfigure}[b]{0.31\linewidth}
        \includegraphics[width=\textwidth]{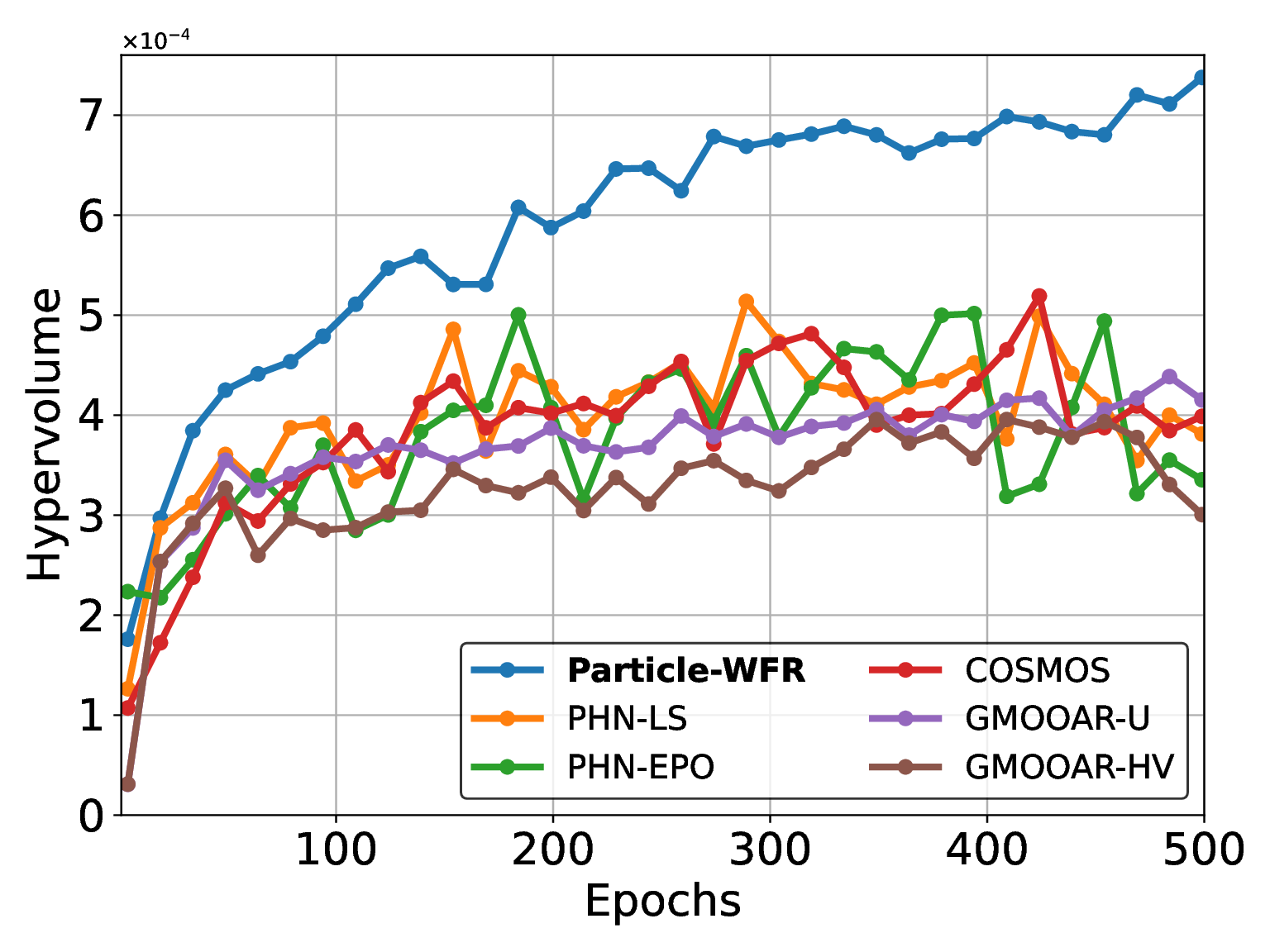}
        \caption{$N=16$}
        \label{fig:mslr_16}
    \end{subfigure}
    \caption{Performance comparison of different methods on the MSLR-WEB30K dataset. Our method achieves the best HV value on test NDCG@10 and performance improves as particle count $N$ increases from 8 to 16.}
    \label{fig:mslr}
\end{figure*}

In this example, We test MOO methods on the learning-to-rank (LTR) task~\citep{dai2011multi,hu2018collaborative,carmel2020multi,mahapatra2023multi,mahapatra2023query}. In LTR tasks~\citep{liu2009learning}, we deal with a collection of \emph{query groups} $\Psi =  \{\Psi^{(p)}\}_{p=1}^{|\Psi|}$, where each query group $\Psi^{(p)}$ consists of $n^{(p)}$ items. These items are characterized by a feature vector $\x^{(p)}_j \in \R^{d_f}$, generated from upstream tasks, and an associated relevance label $y^{(p)}_j$. The goal is to derive an ordering $\pi^{(p)}$ for the items in each query group $\Psi^{(p)}$ given the feature vectors $\x^{(p)}_{j}$, optimizing the utility $u(\pi^{(p)}; \{y^{(p)}_j\}_{j=1}^{n^{(p)}})$ of the ordered list. The Normalized Discounted Cumulative Gain (NDCG)~\citep{wang2013theoretical} is a widely adopted ranking metric. 

Following the current LTR techniques, we employ a neural network $f_\theta$, with $\theta$ denoting the parameters, that accepts the feature vector as input and produces a score, based on which we sort the items in each query group and obtain the ordering $\pi^{(p)}$. The neural network is trained using the empirical loss of the following form:
\begin{equation}
    \L(\theta; \Psi) = \dfrac{1}{|\Psi|} \sum_{p=1}^{|\Psi|}      \ell\left(\{f_\theta(\x_j^{(p)})\}_{j=1}^{n^{(p)}}; \{y_j^{(p)}\}_{j=1}^{n^{(p)}}\right),
\label{eq:loss}
\end{equation}
where $\ell(\cdot, \cdot)$ is the query group-wise loss function. As the differentiable surrogate for the non-differentiable NDCG metric, we adopt the \emph{Cross-Entropy loss}~\citep{cao2007learning} for $\ell$, which is one of the most robust choices as supported by~\citet{qin2021neural}. Further details on the settings, metrics, and losses can be found in Appendix~\ref{app:ltr}.

In real-world scenarios, an item might have multiple labels of interest, denoted as $y_{j}^{(p), i}$. Each of these labels signifies the relevance of the corresponding item concerning the $i$-th ranking objective for $i \in [m]$. This gives rise to an MOO problem w.r.t. the neural network parameters $\theta$ and the $m$ loss functions, $\L_i(\theta;\Psi)$ as the objective functions, obtained by substituting $\{y_j^{(p)}\}_{j=1}^{n^{(p)}}$ in~\eqref{eq:loss} with the respective label $\{y_j^{(p),i}\}_{j=1}^{n^{(p)}}$.

\begin{remark}
Unlike various studies from~\citet{sener2018multi} that integrate multi-task learning into their experimental designs, we use the multi-objective LTR task for benchmarking, as suggested by~\citet{ruchte2021multi}.
\end{remark}

We conduct experiments on the Microsoft Learning-to-Rank Web Search (MSLR-WEB10K) dataset~\citep{qin2013introducing}.
The MSLR-WEB10K dataset consists of 10,000 query groups ($|\Psi| = 10^4$) 
and each item is associated with 136 features and a relevance label. Following the practice of~\citet{mahapatra2023multi}, we treat the first 131 features as the input ($d_f = 131$) and combine the last 5 features, \emph{viz.} Query-URL Click Count, URL Dwell Time, Quality Score 1, Quality Score 2, with the relevance label, as six different ranking objectives ($m=6$). Our Particle-WFR method is implemented with Distributed Data Parallel (DDP) in PyTorch~\citep{paszke2019pytorch} with extensive scalability, and further details are provided in Appendix~\ref{app:mslr_imp}.

In Figure~\ref{fig:mslr}, we present the learning curves of the hypervolume (HV)\footnote{Hypervolume is a quality indicator for assessing the solutions in MOO (\emph{cf.} Appendix~\ref{app:hv}).} values of the testing NDCG@10 across several benchmark methods: PHN-LS, PHN-EPO, COSMOS, GMOOAR-HV, and GMOOAR-U on the MSLR-WEB10K dataset over 500 epochs of training in the neural network setting\footnote{We do not include a comparison with MOO-SVGD and MOO-LD, as their performance in neural network contexts has been found limited~\citep{chen2022multi} and our method consistently outperforms these other methods on synthetic datasets.}. As listed in Table~\ref{tab:mslr}, hypernetwork-based methods struggle to cover the Pareto front. This might be partially attributed to their dependence on the structure of the loss space and their intrinsic sensitivity, resulting in less desirable generalization. Our Particle-WFR method outperforms the other methods, achieving the highest HV value in three instances where the number of particles $N= 8, 12, 16$. Notably, the HV value of our method has surpassed all other methods after only 100 epochs. This demonstrates the effectiveness and efficacy of our method in solving the MOO problem within the neural network-driven LTR context.

\section{CONCLUSION}

This paper proposes a novel interacting particle method based on the Wasserstein-Fisher-Rao gradient flow for solving the MOO problem. Our method enjoys interpretable and intuitive physical meanings with provable convergence guarantees. We implement the Wasserstein-Fisher-Rao gradient flow by the splitting scheme, where the Wasserstein gradient flow is approximated by the overdamped Langevin dynamics and the Fisher-Rao gradient flow by the birth-death dynamics. We compare our proposed method with several recent state-of-the-art methods on challenging datasets. The results show that our method is favorable when dealing with complicated Pareto fronts. 

\subsubsection*{Acknowledgements}

This work was conducted during Yinuo Ren's internship as an Applied Scientist at Amazon Search. We thank the Amazon Search team for their support and feedback. 
We also thank the anonymous reviewers for their insightful comments and suggestions. 


\bibliography{main}


\appendix
\onecolumn

\section{RELATED WORKS}
\label{app:related_works}

\paragraph{Gradient-Free MOO Methods.}

In the previous decades, research in multi-objective optimization has been focused on the evolutionary algorithms and particle swarm methods~\citep{tamaki1996multi,deb2002fast,parsopoulos2002particle,konak2006multi,reyes2006multi,zhang2007moea}. These methods are based on the idea of maintaining a population of solutions, which are iteratively improved by applying genetic operators such as mutation and crossover. Distributed computing techniques have also been explored in this context~\citep{zhou2011multiobjective}. However, because these methods assume that gradient information is unavailable, they are often computationally expensive and do not scale well to high-dimensional problems. Another line of research in MOO is the Bayesian
optimization~\citep{laumanns2002bayesian,belakaria2020uncertainty,konakovic2020diversity,tu2022joint}. These methods are effective for small-scale black-box optimization problems but also suffer from the curse of dimensionality in machine learning tasks.

\paragraph{Gradient-Based Methods.} 
In recent years, gradient-based methods have been proposed to solve the MOO problem. These methods are primarily designed and believed to effectively sacle up for high0dimensional machine learning applications.
To profile the Pareto front, one of the most straightforward gradient-based approaches is to parametrize the Pareto front by preference vectors, \emph{i.e.} the vector formed by the values of the objective functions on the Pareto front. The prototype of all preference vector-based methods is the \emph{weighted sum method}, aka \emph{Linear Scalarization (LS)} that uses a linear combination of weights and
objective functions to find optimal solutions. However, the weighted sum method cannot handle the concavity of the Pareto front~\citep{boyd2004convex}. To address this issue, several methods are proposed to find local Pareto optimal points catering to predetermined preference vectors, including PF-SMG~\citep{liu2021stochastic}, PMTL~\citep{lin2019pareto}, and EPO~\citep{mahapatra2020multi}. These methods often rely on selecting preference vectors, which may be difficult or even impossible for complicated Pareto fronts, resulting in sub-optimal solutions.

Recently, hypernetwork-type approaches~\citep{lin2020controllable,ruchte2021scalable,chen2022multi,hoang2023improving} have been proposed to learn the Pareto front directly from data originated from PHN-LS and PHN-EPO~\citep{navon2020learning}, which generalizes the weighted sum method and EPO directly. Hypernetwork methods aim to design neural networks that accept preference vectors as inputs and directly generate solutions on the Pareto front. However, as demonstrated in our experiments, these methods are generally not very robust when dealing with complicated Pareto fronts and challenging tasks.

\paragraph{Gradient Flows and Interacting Particle Method.} Gradient flows have been studied widely as one of the most important techniques in the literature of optimal transport and sampling~\citep{villani2021topics}. Originated from~\citep{jordan1998variational}, Wasserstein gradient flow is one of the most renowned gradient flows (\emph{cf.}~\citet{santambrogio2017euclidean}). Stein Variational Gradient flow (SVGD)~\citep{liu2016stein,liu2017stein} can be viewed as the gradient flow with respect to a kernelized Wasserstein metric. MOO-SVGD~\citep{liu2021profiling} is a recent method that uses SVGD to solve the MOO problem. 
Wasserstein-Fisher-Rao metric and the corresponding gradient flow~\citep{chizat2018interpolating,liero2018optimal,liero2016optimal} are recently proposed to study unbalanced optimal transport problems and have been applied to sampling by interacting particle methods~\citep{lu2019accelerating,yan2023learning}.

\paragraph{Multi-Objective Learning-to-Rank.} MOO finds extensive applications in Learning-to-Rank (LTR) because it naturally involves multiple, potentially conflicting ranking metrics, such as precision and recall, or multiple relevance labels, such as product quality and purchase likelihood in e-commerce product search. Unlike supervised learning problems, where each sample has a clearly defined target as a single categorical label or numerical value, LTR tasks aim to identify an optimal permutation within a large and discrete search space for each query-group. This optimization is usually conducted to maximize a linear additive ranking metric, such as Normalized Discount Cumulative Gain (NDCG) \citep{wang2013theoretical}. As a parametrized ranking model always operates as a scoring function, generating numerical scores to rank documents within a query, all ranking metrics are non-differentiated with respect to the predicted scores. To reframe LTR as a supervised learning problem, various differentiable loss functions are introduced as alternatives to optimize ranking metrics; see \citet{qin2021neural} and references therein.  

In the context of Multi-Objective LTR, existing work can be categorized into two main approaches: label aggregation \citep{dai2011multi, carmel2020multi} and loss aggregation \citep{hu2018collaborative, mahapatra2023multi, mahapatra2023query}. In the former, labels assigned to the same document are combined into a single label, which is then used as input for a ranking loss function. In the latter, the aim is to optimize aggregated ranking loss functions, each corresponding to a different objective. In our experiments, we adopt later approach.

\section{MISSING PROOFS}
\label{app:proof}

In this section, we provide the missing proofs in the theoretical analysis part (Section~\ref{sec:theory}) of the main text.

\begin{proof}[Proof of Theorem~\ref{thm:convergence}]
    The decay of the functional $\Ec[\rho]$~\eqref{eq:decay} is due to the following 
    calculation by plugging in~\eqref{eq:wfrgf}:
    \begin{equation*}
        \begin{aligned}
            \p_t \Ec[\rho_t] &= \int_{\D} \delta_\rho \Ec[\rho_t] \p_t \rho_t \d \x\\
            &= \int_{\D} \delta_\rho \Ec[\rho_t] \nabla \cdot (\rho_t \nabla \delta_\rho\Ec[\rho_t]) - \rho_t  \delta_\rho \Ec[\rho_t]  \widetilde {\delta_\rho \Ec[\rho_t]} \d \x\\
            &= - \int_{\D} \rho_t \left\|\nabla \delta_\rho \Ec[\rho_t]\right\|^2 + \rho_t \widetilde{\delta_\rho \Ec[\rho_t]}^2 \d \x,
        \end{aligned}
    \end{equation*}
    where the last equality is due to the integration-by-parts and the
    fact that 
    $$
    \int_\D \rho_t \widetilde{\delta_\rho
    \Ec[\rho_t]}\d \x =\int_\D \rho_t\left(\delta_\rho
    \Ec[\rho_t] - \E{\delta_\rho \Ec[\rho_t]}{\rho_t}\right)\d \x = 0.
    $$

    Since $\rho^*$ is the minimizer of $\Ec[\rho]$, we have the following
    optimality condition:
    \begin{equation*}
        \nabla \delta_\rho \Ec[\rho^*] = 0, \textit{ a.e.}, \text{   and   } \widetilde{\delta_\rho \Ec[\rho^*]} = 0, \textit{ a.e.},
    \end{equation*}
    which implies
    \begin{equation}
        \delta_\rho\Ec[\rho^*] = \alpha_1 \delta_\rho \F_1 +  \alpha_2 \delta_\rho \F_2 + \beta \delta_\rho \G[\rho^*] - \gamma \delta_\rho \H[\rho^*] = C^*, \textit{ a.e.},
        \label{eq:star}
    \end{equation}
    where $C^* = \E{\delta_\rho \Ec[\rho^*]}{\rho^*}$.

    Now suppose $\rho'$ is another minimizer of $\Ec[\rho]$, a
    similar argument yield
    \begin{equation}
        \delta_\rho\Ec[\rho'] = \alpha_1 \delta_\rho \F_1 +  \alpha_2 \delta_\rho \F_2 + \beta \delta_\rho \G[\rho'] - \gamma \delta_\rho \H[\rho'] \equiv C',
        \label{eq:prime}
    \end{equation}
    where $C' = \E{\delta_\rho \Ec[\rho']}{\rho'}$.

    Subtracting the above two equations~\eqref{eq:star} and~\eqref{eq:prime}, we obtain
    \begin{equation*}
        \begin{aligned}
            C' - C^*=&\beta \delta_\rho \G[\rho^*] - \beta \delta_\rho \G[\rho'] - \gamma \delta_\rho \H[\rho^*] + \gamma \delta_\rho \H[\rho']\\
            =&\beta \int_\D R(\f(\cdot), \f(\y)) (\rho^* - \rho')(\d \y) + \gamma\log \rho^* - \gamma \log \rho'.
        \end{aligned}
    \end{equation*}

    Multiply both sides with $\rho^* - \rho'$ and integrate over $\D$, we have
    \begin{equation*}
        \begin{aligned}
            0 =& \int_D (C' - C^*) (\rho^* - \rho') (\d \x)\\
            =& \beta \int_\D (\rho^* - \rho')(\d \x)  R(\f(\x), \f(\y)) (\rho^* - \rho')(\d \y) + \gamma \int_D \log \dfrac{\rho^*}{\rho'}(\y) (\rho^* - \rho')(\d \y), 
        \end{aligned}
    \end{equation*}
    implying that $\rho^* = \rho'$ and therefore the minimizer $\rho^*$ is unique.
\end{proof}

To prove Theorem~\ref{thm:exponential}, we need the following lemma:
\begin{lemma}[{\citet[Theorem 2.4 and Remark 2.6]{lu2023birth}}]
    Let $\rho_0$ and $\rho^*$ be two probability measures absolutely continuous with respect to the Lebesgue measure and have the density functions $\rho_0(\x)$ and $\rho^*(\x)$, respectively. Suppose that the initial condition $\rho_0$ satisfies 
    \begin{equation*}
        \inf_{\x\in\D}\dfrac{\rho_0(\x)}{\rho^*(\x)}\geq e^{-M}
    \end{equation*}
    for some constant $M$, we have for all $t>0$, the Wasserstion-Fisher-Rao gradient flow $\rho_t$~\eqref{eq:wfrgf} with respect to the KL divergence $\KL(\rho_t|\rho^*)$ between $\rho_t$ and $\rho^*$ satisfies
    \begin{equation*}
        \KL(\rho_t|\rho^*)\leq M e^{- t} + e^{- t+Me^{- t}}\KL(\rho_0|\rho^*).
    \end{equation*}
\label{lemma}
\end{lemma}

Then, we are ready to prove Theorem~\ref{thm:exponential}.
\begin{proof}[Proof of Theorem~\ref{thm:exponential}]
    The constant $C$ may change from line to line in this proof.

    As we are considering the scenario where the repulsive potential term $\G[\rho]$ is turned off, \emph{i.e.} $\beta=0$, the functional $\Ec[\rho]$ can be simplified as
    \begin{equation*}
        \Ec[\rho] = \alpha_1 \F_1[\rho] + \alpha_2 \F_2[\rho] - \gamma \H[\rho],
    \end{equation*}
    and the corresponding minimizer of the energy potential $\Ec[\rho]$ satisfies
    \begin{equation}
        \delta_\rho \Ec[\rho^*] = \alpha_1 \delta_\rho \F_1 + \alpha_2 \delta_\rho \F_2 - \gamma \delta_\rho \H[\rho^*] = C,
        \label{eq:star2}        
    \end{equation}
    that is
    \begin{equation*}
        \log \rho^* = -\dfrac{\alpha_1 \delta_\rho \F_1 + \alpha_2 \delta_\rho \F_2}{\gamma} + C,
    \end{equation*}
    and thus the minimizer $\rho^*$ satisfies the Gibbs-type distribution as in~\eqref{eq:gibbs}:
    \begin{equation*}
        \rho^* \propto \exp\left(-\dfrac{\alpha_1 \delta_\rho \F_1 + \alpha_2 \delta_\rho \F_2}{\gamma}\right)=\exp\left(-\dfrac{\alpha_1\|\g^\dagger\|^2 + \alpha_2 \int_{\P} D(\f(\cdot), \f(\y)) \mu_{\P}(\d \y)}{\gamma}\right),
    \end{equation*}
    which is also the unique minimizer of $\Ec[\rho]$, as the result of Theorem~\ref{thm:convergence}.

    Notice that the Fr\'echet derivative of the energy functional $\Ec[\rho]$ can be rewritten as:
    \begin{equation}
        \begin{aligned}
            \delta_\rho \Ec[\rho] &=  \gamma \left(\dfrac{\alpha_1\delta_\rho\F_1 + \alpha_2 \delta_\rho\F_2}{\gamma}- \delta_\rho\H[\rho] \right)\\
            &= \gamma \left( -\log \rho^* + \log \rho \right) + C\\
            & = \gamma \log \dfrac{\rho}{\rho^*} + C = \gamma \delta_\rho \KL(\rho | \rho^*) + C,
        \end{aligned}
    \end{equation}
    we reparametrize the time with $\tau = t/\gamma$, and thus rewrite the Wasserstein-Fisher-Rao gradient flow~\eqref{eq:wfrgf} as
    \begin{equation*}
        \begin{aligned}
            \p_t \rho_\tau & = \dfrac{1}{\gamma}\p_\tau \rho_\tau = \dfrac{1}{\gamma}\left[\nabla\cdot\left(\rho_\tau \nabla \delta_\rho\Ec[\rho_\tau]\right) - \rho_\tau \left(\delta_\rho\Ec[\rho_\tau] - \int_{\D} \rho_\tau \delta_\rho\Ec[\rho_\tau]\d \x  \right)\right]\\
            &= \nabla\cdot\left(\rho_\tau \nabla \log \dfrac{\rho_\tau}{\rho^*}\right) - \rho_\tau \left(\log\dfrac{\rho_\tau}{\rho^*} - \int_{\D} \rho_\tau \log \dfrac{\rho_\tau}{\rho^*} \d \x  \right),
        \end{aligned}
    \end{equation*}
    that is the Wasserstein-Fisher-Rao gradient flow of the KL divergence between $\rho_\tau$ and $\rho^*$.

    Together with the following assumption on the initial distribution in the theorem statement:
    $$
    \inf_{\x\in\D}\dfrac{\rho_0(\x)}{\rho^*(\x)}\geq e^{-M},
    $$
    we have by Lemma~\ref{lemma} that
    \begin{equation*}
        \KL(\rho_\tau|\rho^*)\leq M e^{- t} + e^{- t+Me^{- t}}\KL(\rho_0|\rho^*),
    \end{equation*}
    and thus~\eqref{eq:exponential} holds.
\end{proof}

\section{ALGORITHM DETAILS}
\label{app:implementation}

In this section, we provide additional implementation details of our method (Algorithm~\ref{alg:moo}), including several techniques and heuristics on the
hyperparameter selection and optimization strategy.

\paragraph{Multiple-stage optimization.} 

In general, the hyperparameters $\alpha_i$, $i=1,2$, $\beta$
and $\gamma$ in the expression~\eqref{eq:energy} should be chosen in a way that
the corresponding terms are balanced, and thus the minimizer $\rho^*$ satisfies
the desired properties, \emph{i.e.} diversity and global Pareto optimality. 

Since the main aim of multi-objective optimization is to profile the Pareto front instead of aggregating all potentials together as in~\eqref{eq:energy}, an alternative understanding of the
problem is the following constraint optimization problem:
\begin{equation}
        \min_{\mathrm{supp} \rho \subset \D} \quad \G[\rho]\quad\text{s.t.} \quad \F[\rho] \leq C,
\end{equation}
where the tolerance $C$ controls how close the population $\rho$ is to the Pareto front. This type of constraint optimization problem is often solved by the interior point method or the primal-dual method~\citep{wright1997primal}. However, unfortunately, the Euclidean geometry exploited by these methods is generally unavailable in the space of probability measures.

Another popular method of solving problems of this kind is the \emph{penalty
method}. The idea is to relax the original problem by adding a penalty term that
penalizes the violation of the constraint. The convergence of this method is
only asymptotic, \emph{i.e.} the penalty parameter should be gradually increased
to infinity.

Therefore, by viewing our method as a penalty relaxation to the above constraint optimization problem, one of the most important heuristics for the hyperparameter selection and tuning in our method is to split the
optimization into multiple stages: 
\begin{itemize}
    \item In the first stage, we only impose a relatively small penalty on the constraint
    (with small dominance potential coefficient $\alpha_2$, but large repulsive potential
    coefficient $\beta$ and diffusion coefficient $\gamma$
    in~\eqref{eq:energy}), encouraging the population to diversify and explore;
    \item As we gradually increase the penalty, the objective functions
    (constraints) will be balanced with the structural potentials being
    optimized, and thus, the population will be pushed towards
    the Pareto front while preserving diversity.
    \item In the final stage, we impose a larger penalty (with a large 
    dominance potential coefficient $\alpha_2$, but small repulsive potential
    coefficient $\beta$ and diffusion coefficient $\gamma$
    in~\eqref{eq:energy}), eliminating dominated particles and thus obtaining a
    population that is close to the Pareto front.
\end{itemize}

\paragraph{Numerical approximation of $\nabla \|\g^\dagger\|^2$.} 

Denote $\bm{\alpha}^\dagger(\x) = (\alpha_1^\dagger(\x), \cdots, \alpha_m^\dagger(\x))$ as the
optimal solution of~\eqref{eq:mgda} at point $\x$. Then we have $\g^\dagger(\x)
= \sum_{i=1}^m \alpha_i^\dagger(\x) \nabla f_i(\x) := (\nabla \f) \bm{\alpha}^\dagger$.  
Moreover, the optimality of $\bm{\alpha}^\dagger$ yields the
following relation:
\begin{equation}
    \nabla \bm{\alpha}^\dagger (\nabla \f)^\top \g^\dagger = \bm{0}.
\end{equation}

Then $\nabla\delta_\rho\F_1[\rho]$ can be computed as
\begin{equation}
    \begin{aligned}
        \nabla\delta_\rho\F_1[\rho]&=\nabla\|\g^\dagger\|^2 = 2 (\nabla \g^\dagger) \g^\dagger\\
        &= 2(\nabla^2 \f : \bm{\alpha}^\dagger + \nabla \bm{\alpha}^\dagger (\nabla\f)^\top) \g^\dagger\\
        &= 2(\nabla^2 \f : \bm{\alpha}^\dagger) \g^\dagger,
    \end{aligned}
\end{equation}
where $\nabla^2 \f : \bm{\alpha}^\dagger = \sum_{i=1}^m \alpha_i^\dagger
\nabla^2 f_i$.

However, the computation of $\nabla^2 \f$ is very expensive, as it involves the second-order derivatives of the objective functions. Therefore, we treat $\nabla^2 \f:\bm{\alpha}^\dagger$ as a preconditioner and approximate it by the identity matrix. 

\paragraph{Numerical approximation of $\delta_\rho \F_2$.} 

In~\eqref{eq:dominanceg}, we use a predetermined measure $\mu_\P$ on the Pareto front, which is not known in advance. However, as mentioned earlier, we only turn on the dominance potential term $\F_2[\rho]$ in the final stage of our method, and thus, we can use the empirical measure of the population $\rho_t$ as a proxy of $\mu_\P$.

Furthermore, one can apply the following relaxation to the dominance kernel $D(\cdot,\cdot)$ presented in~\eqref{eq:dominance_kernel}:
\begin{equation*}
    D(\f(\x, \f(\y))) = \prod_{i=1}^m \left(\max\{0, f_i(\x) - f_i(\y)\} + c {\bm 1}\{f_i(\x) - f_i(\y)\geq0\} \right),
\end{equation*}
where $c>0$ is a small constant and $\bm 1\{\cdot\}$ is the indicator function. This relaxation deals with the rare scenario that there exists a point $\x$ that is dominated by $\y$ in some objectives but has the exactly same values in some other objectives.

\paragraph{Numerical approximation of $\delta_\rho \H[\rho_t]$.} 

The computation of the instantaneous birth-death rate $\Lambda_t$~\eqref{eq:lambda} in the implementation of the birth-death dynamics involves the computation of the Fr\'echet derivative of the entropy term $\H[\rho_t]$. However, as we are approximating $\rho_t$ by a set of particles $\{\x_k\}_{k=1}^N$, the Fr\'echet derivative $\delta_\rho \H[\rho_t]$ cannot be directly computed. Instead, we approximate it by the following \emph{kernel density estimation} technique, as also adopted by~\citet{lu2019accelerating}:
\begin{equation}
    \rho_t(\x) \approx \dfrac{1}{N} \sum_{k=1}^N K\left(\dfrac{\x - \x_k}{h}\right),
\end{equation}
where $K(\cdot)$ is a smooth kernel function, such as the Gaussian kernel $K(\x)
= \exp(-\|\x\|^2)$, where $h$ is the bandwidth parameter. Then the Fr\'echet derivative of $\delta_\rho \H[\rho_t]$ can be approximated by 
\begin{equation}
    \delta_\rho \H[\rho_t](\x) \approx -  \log \dfrac{1}{N} \sum_{k=1}^N K(\x - \x_k).
\end{equation}

Another possible technique is to use the Gaussian mixture model (GMM) to approximate the probability density $\rho_t$~\citep{yan2023learning}, which would result in a more straightforward computation of the Fr\'echet derivative $\delta_\rho \H[\rho_t]$. However, finding the particle weights deviates from our method's goal, and we choose to use the kernel density estimation for simplicity. 

In practice, as we are imposing the repulsive potential term explicitly, the contribution of the entropy term is relatively small in the stochastic birth-death dynamics.

\section{EXPERIMENT DETAILS}
\label{app:exp}

In this section, we provide additional details of the experiments conducted in Section~\ref{sec:exp}. We present the closed-form formulas of the ZDT3 problem in Appendix~\ref{app:zdt3}, and additional experiments on the ZDT1 and ZDT2 problems are shown in~\ref{app:zdt1} and~\ref{app:zdt2}, respectively. 
The closed-form formula of the DTLZ7 problem is provided in Appendix~\ref{app:dtlz7}.
We also provide the experiment details of the learning-to-rank task and the MSLR-WEB10K dataset in Appendix~\ref{app:mslr}.

\subsection{ZDT3 Problem}
\label{app:zdt3}

The closed-form formula of the ZDT3 problem is as follows:
\begin{equation}
    \begin{aligned}
        f_1(\x) & = x_1,\\
        f_2(\x) & = g(\x) h(f_1(\x), g(\x)),
    \end{aligned}
\label{eq:zdt}
\end{equation}
where $\x = (x_1,\cdots, x_{30})$, \begin{equation}
    g(\x) = 1 + \frac{9}{29}\sum_{i=2}^{30} x_i.
    \label{eq:zdt_g}
\end{equation}
and 
\begin{equation}
    h(f_1, g) = 1 - \sqrt{\dfrac{f_1(\x)}{g(\x)}} - \dfrac{f_1(\x)}{g(\x)}\sin(10\pi f_1(\x)).
\end{equation}
The feasible region is $\D = [0,1]^{30}$. 

\subsubsection{Additional Experiments on the ZDT1 Problem}
\label{app:zdt1}

The ZDT1 problem is another 30-dimensional two-objective optimization problem of the same form as in~\eqref{eq:zdt}
but with $$
h(f_1, g) = 1 - \sqrt{\dfrac{f_1(\x)}{g(\x)}}.
$$
The feasible region is $\D = [0,1]^{30}$. The Pareto front of this problem is convex, continuous, and smooth, making it a relatively easy problem for MOO methods. This problem is also considered in~\citet{liu2021profiling}.

\begin{figure}[!htb]
    \centering
    \begin{subfigure}[b]{0.3\linewidth}
        \includegraphics[width=\textwidth]{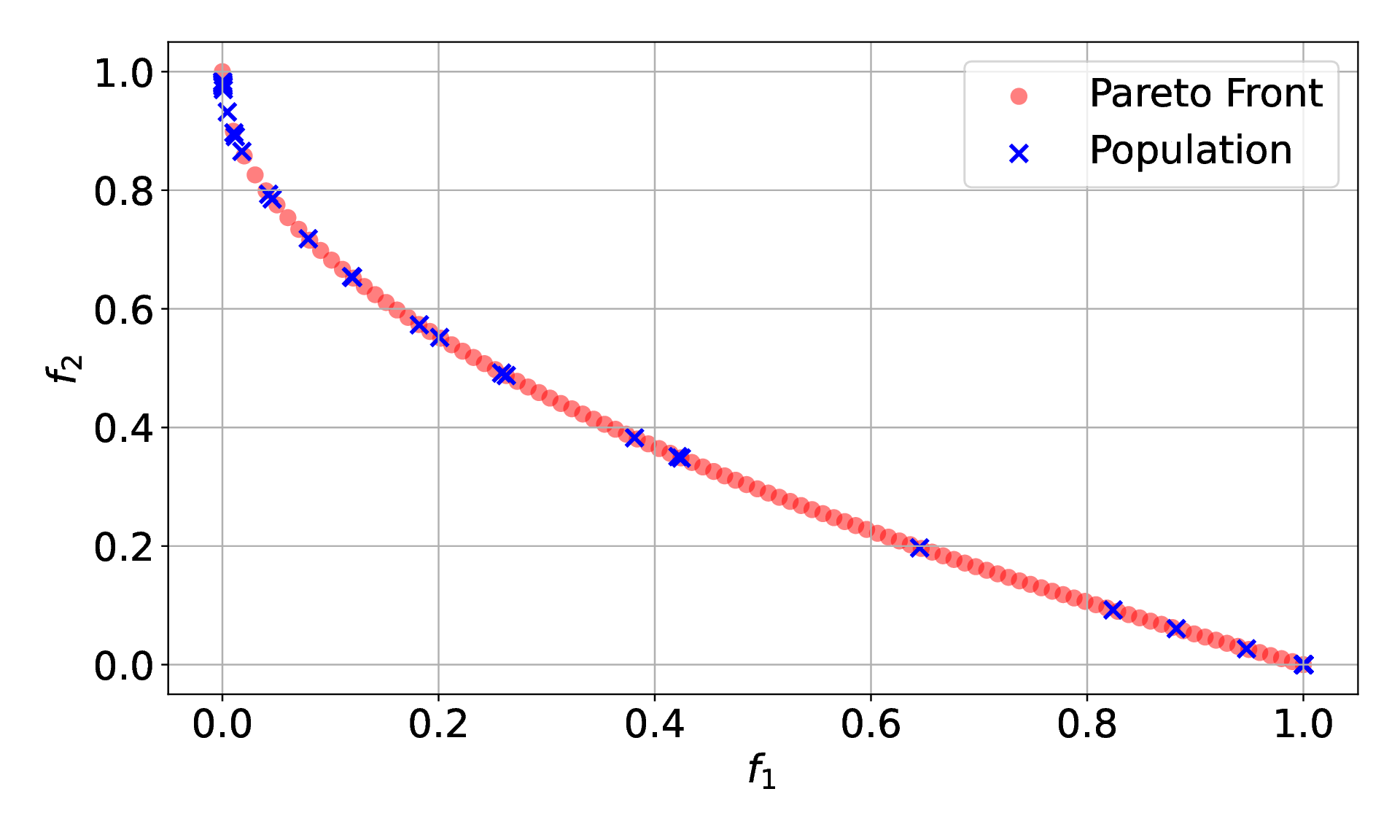}
        \caption{Weighted Sum}
        \label{fig:zdt1_linear}
    \end{subfigure}
    \begin{subfigure}[b]{0.3\linewidth}
        \includegraphics[width=\textwidth]{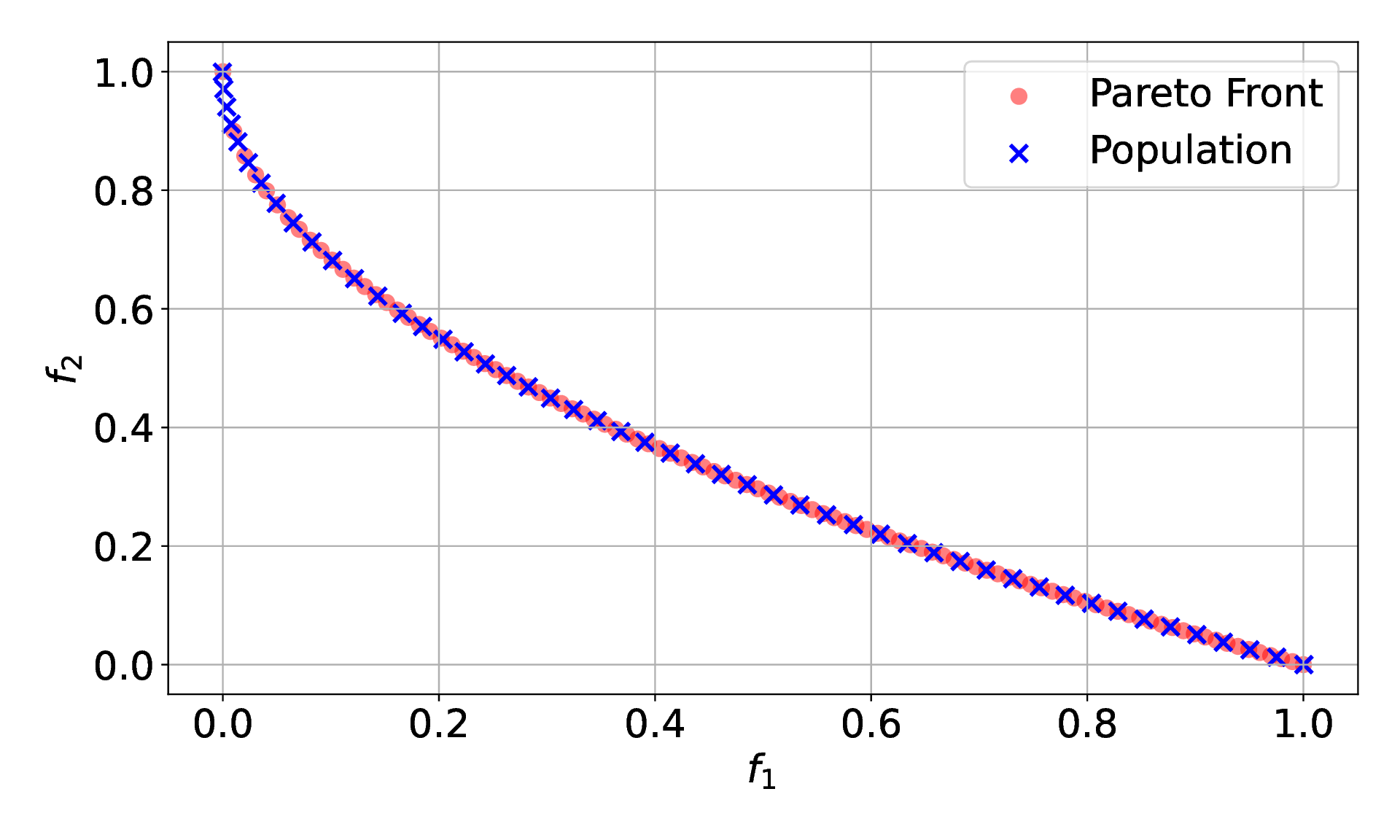}
        \caption{MOO-SVGD}
        \label{fig:zdt1_moosvgd}
    \end{subfigure}
    \begin{subfigure}[b]{0.3\linewidth}
        \includegraphics[width=\textwidth]{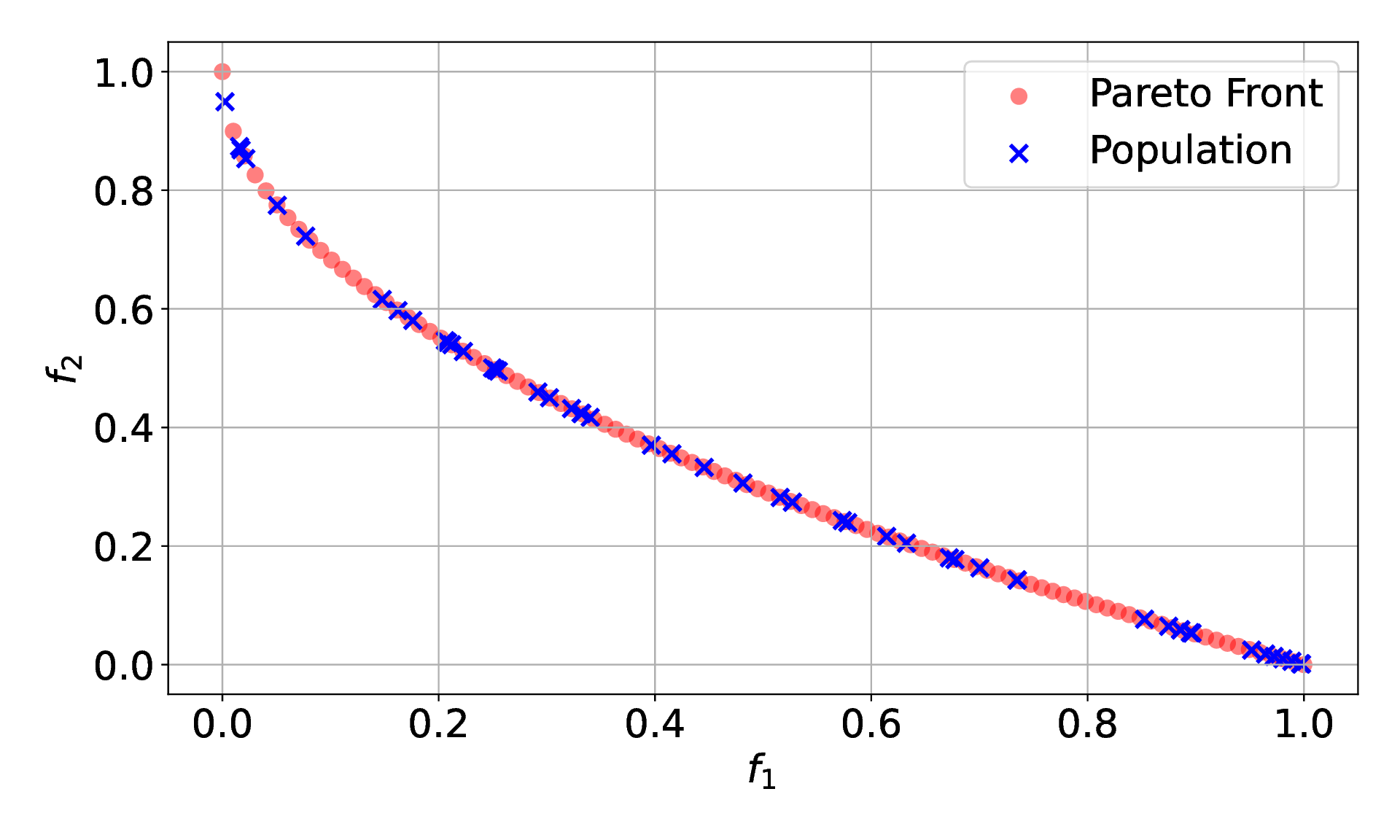}
        \caption{MOO-LD}
        \label{fig:zdt1_moold}
    \end{subfigure}

    \begin{subfigure}[b]{0.3\linewidth}
        \includegraphics[width=\textwidth]{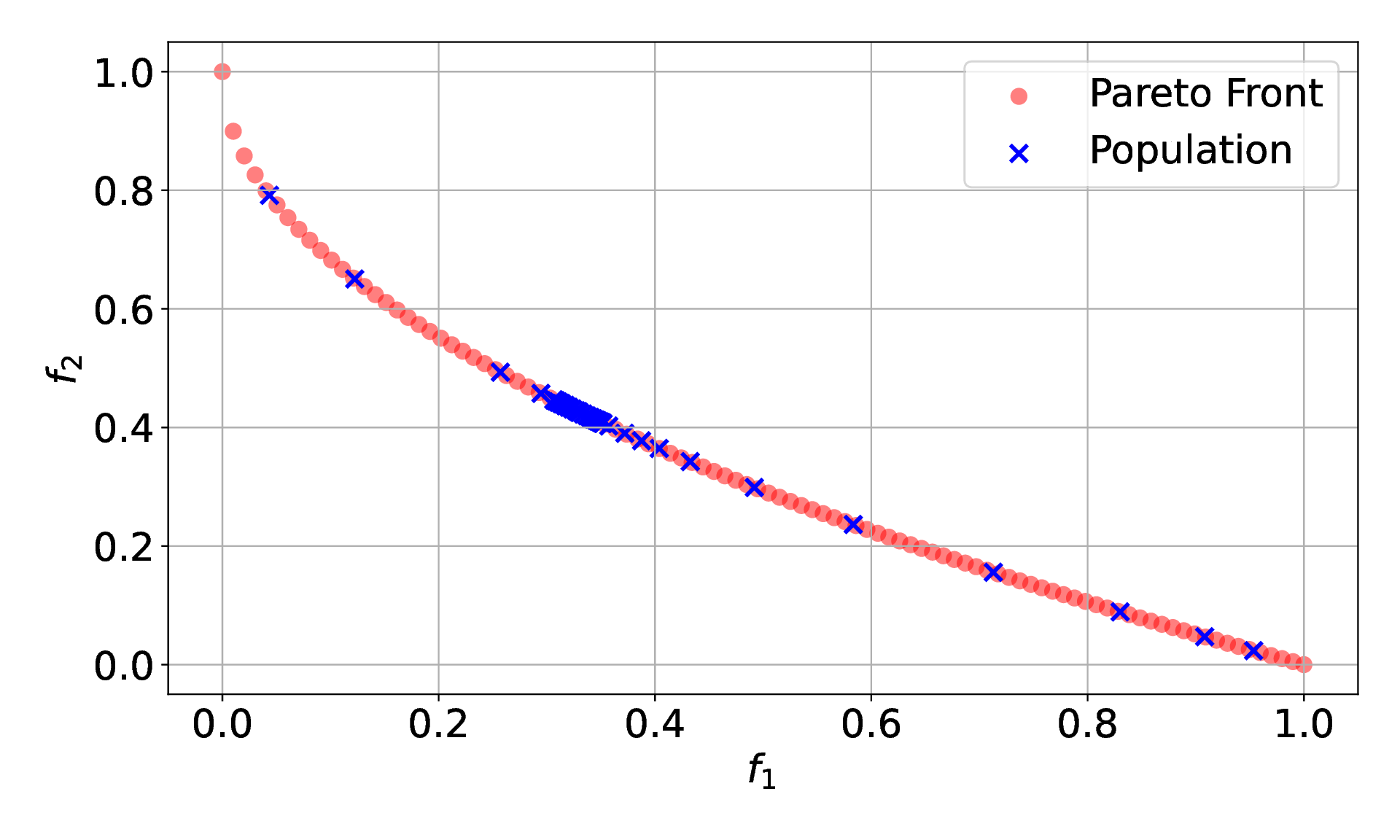}
        \caption{COSMOS}
        \label{fig:zdt1_cosmos}
    \end{subfigure}
    \begin{subfigure}[b]{0.3\linewidth}
        \includegraphics[width=\textwidth]{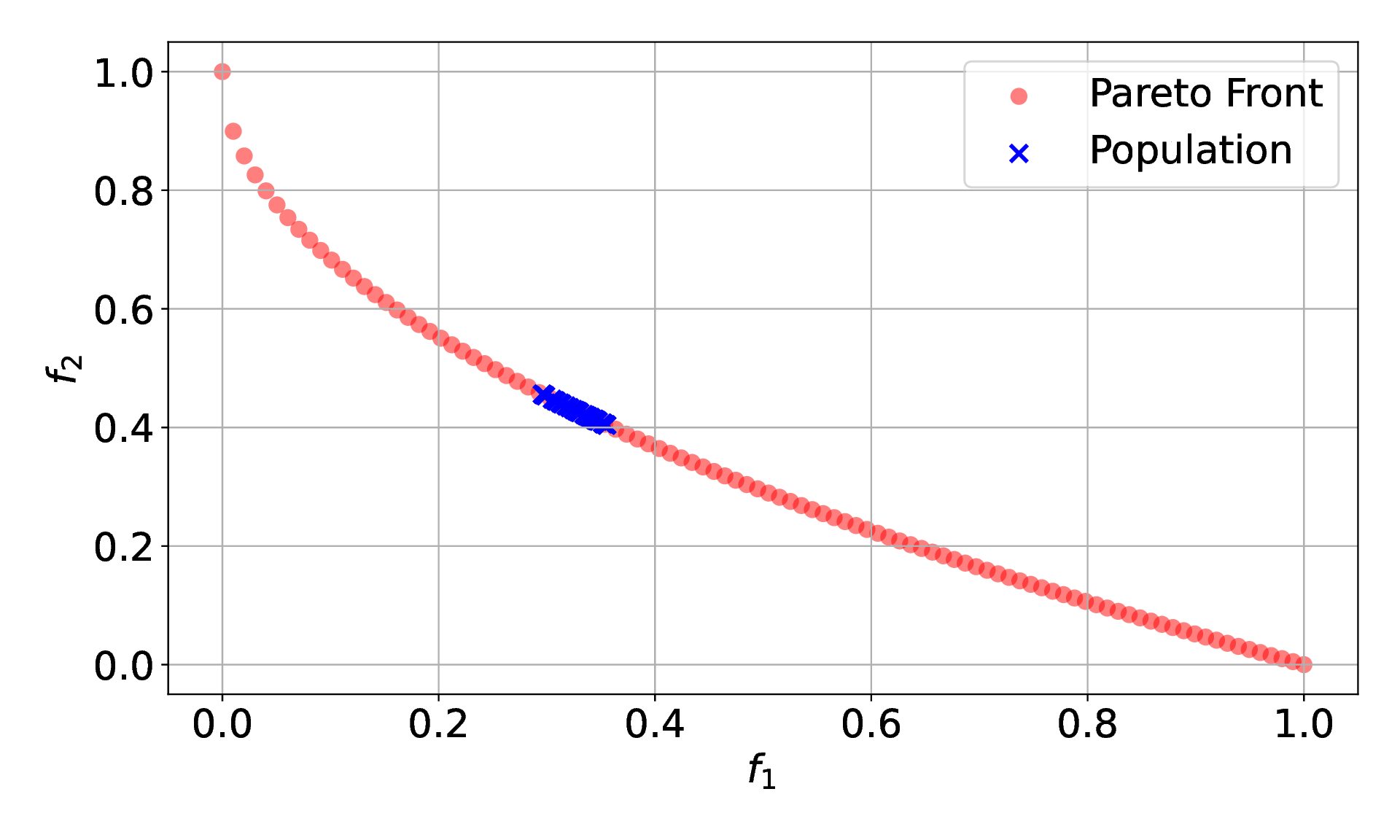}
        \caption{GMOOAR-HV}
        \label{fig:zdt1_argmo}
    \end{subfigure}
    \begin{subfigure}[b]{0.3\linewidth}
        \includegraphics[width=\textwidth]{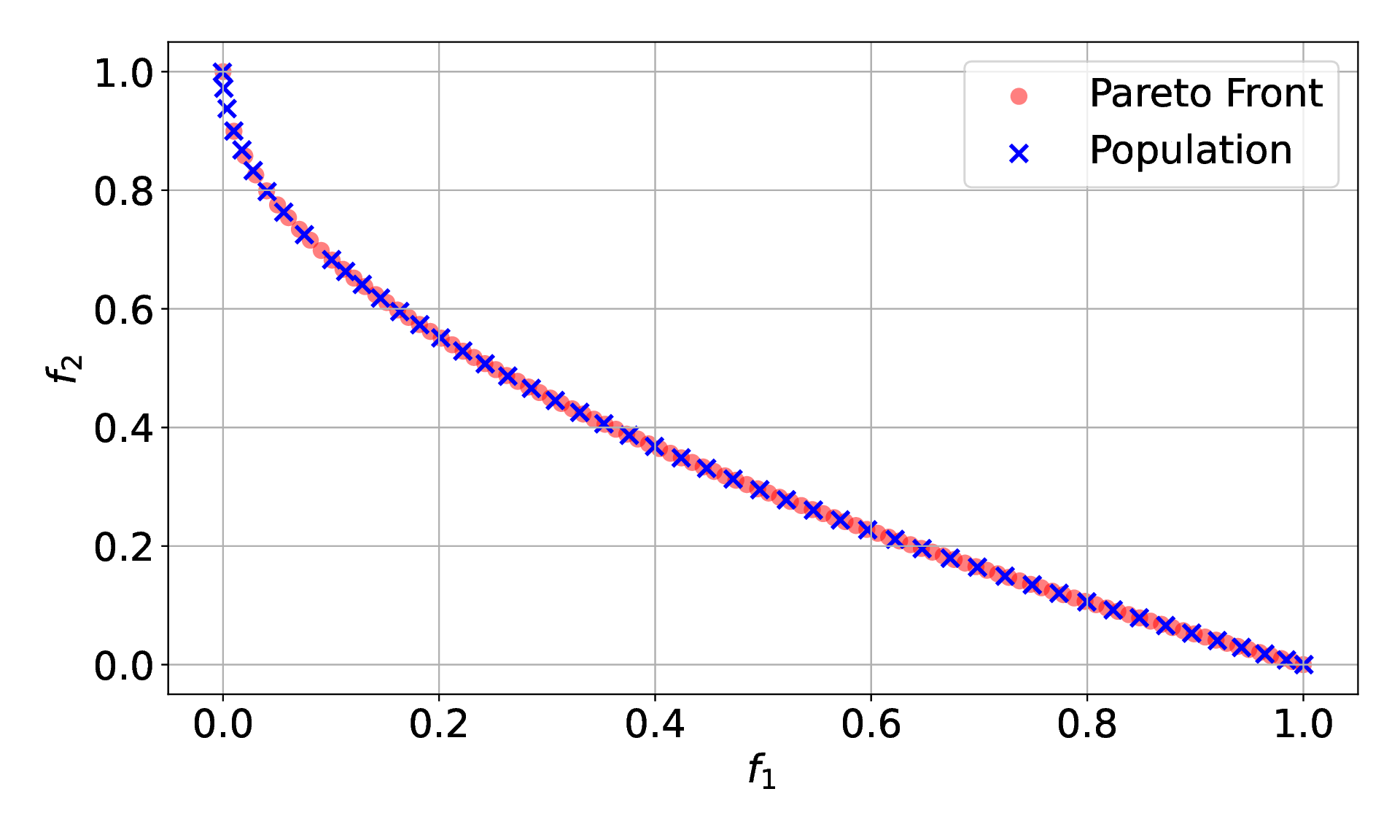}
        \caption{\bf Particle-WFR}
        \label{fig:zdt1_wfr}
    \end{subfigure}
    \caption{Performance comparison of different methods on the ZDT1 problem. The Pareto front is shown in red, and the solutions found by different methods are shown in blue.}
    \label{fig:zdt1}
\end{figure}

\begin{figure}[!htb]
    \centering
    \begin{subfigure}[b]{0.16\linewidth}
        \includegraphics[width=\textwidth]{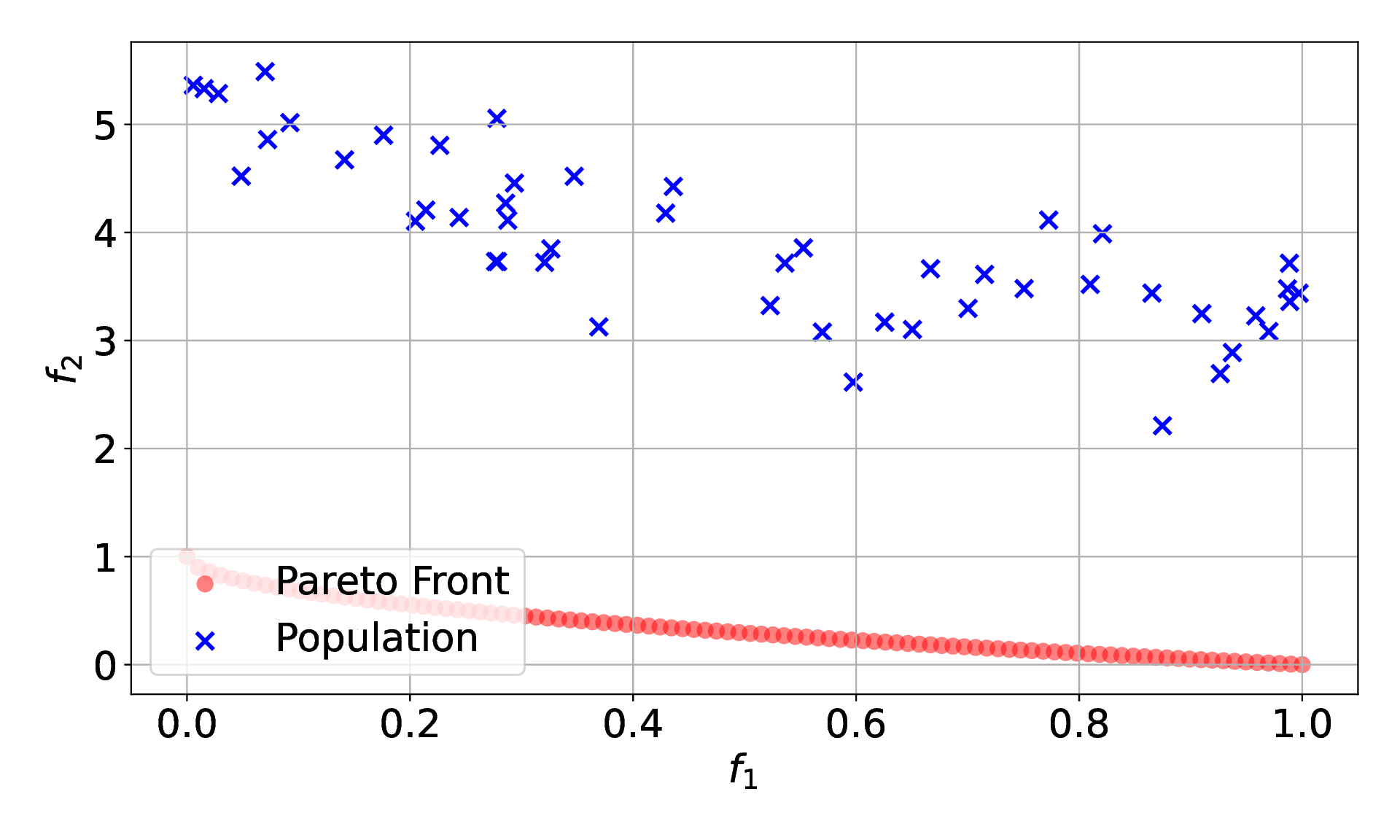}
        \caption{Epoch 0}
        \label{fig:zdt1_particle_0}
    \end{subfigure}
    \begin{subfigure}[b]{0.16\linewidth}
        \includegraphics[width=\textwidth]{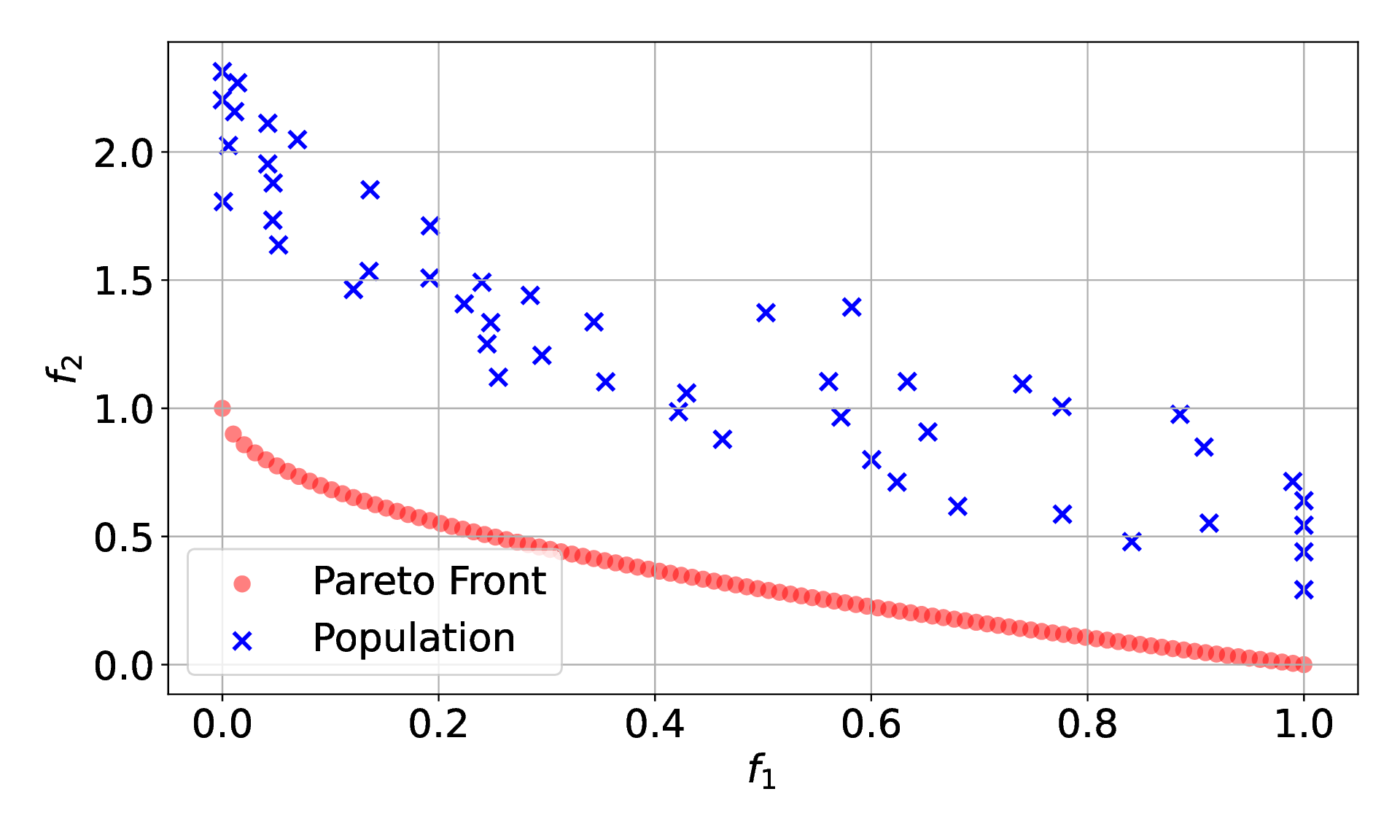}
        \caption{Epoch 1000}
        \label{fig:zdt1_particle_1000}
    \end{subfigure}
    \begin{subfigure}[b]{0.16\linewidth}
        \includegraphics[width=\textwidth]{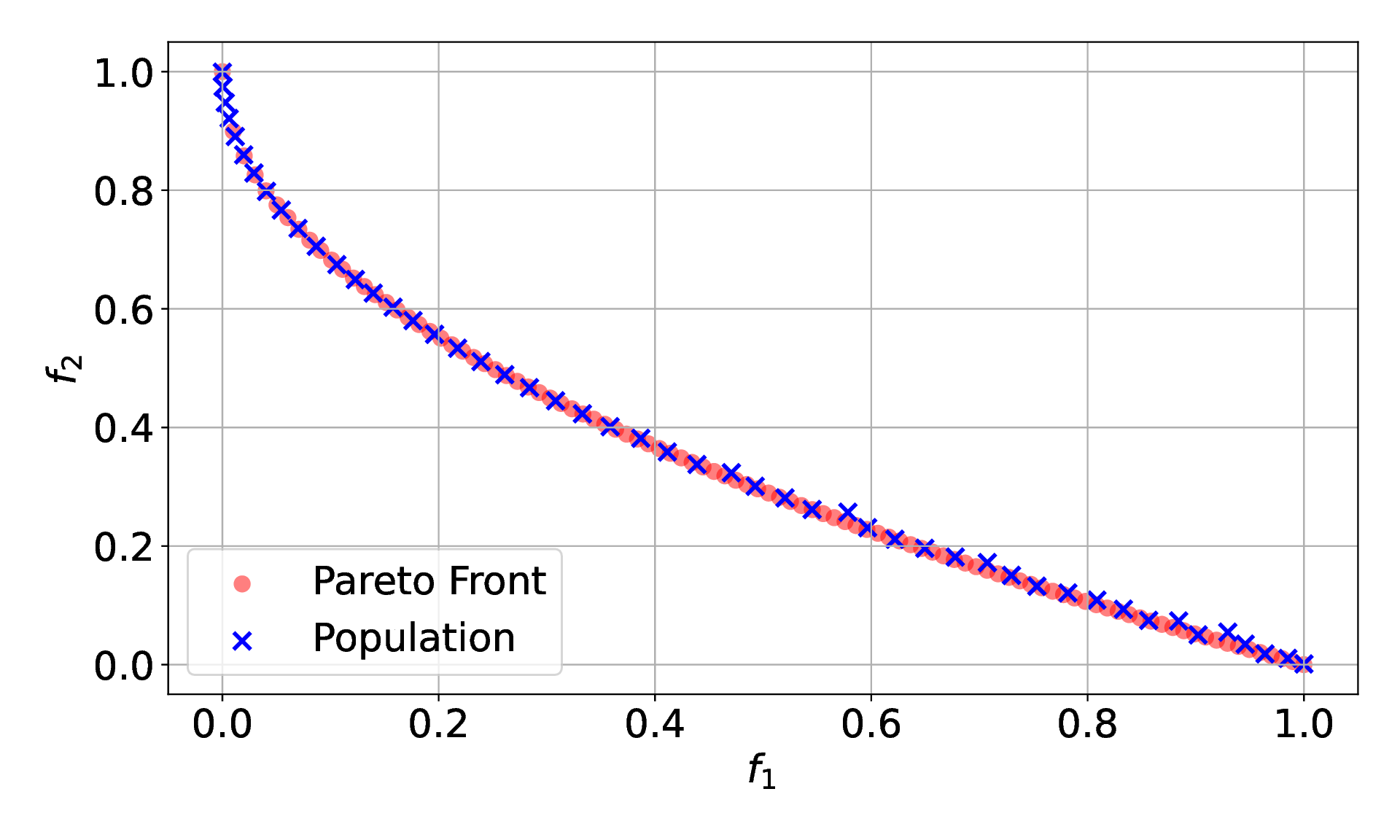}
        \caption{Epoch 2000}
        \label{fig:zdt1_particle_2000}
    \end{subfigure}
    \begin{subfigure}[b]{0.16\linewidth}
        \includegraphics[width=\textwidth]{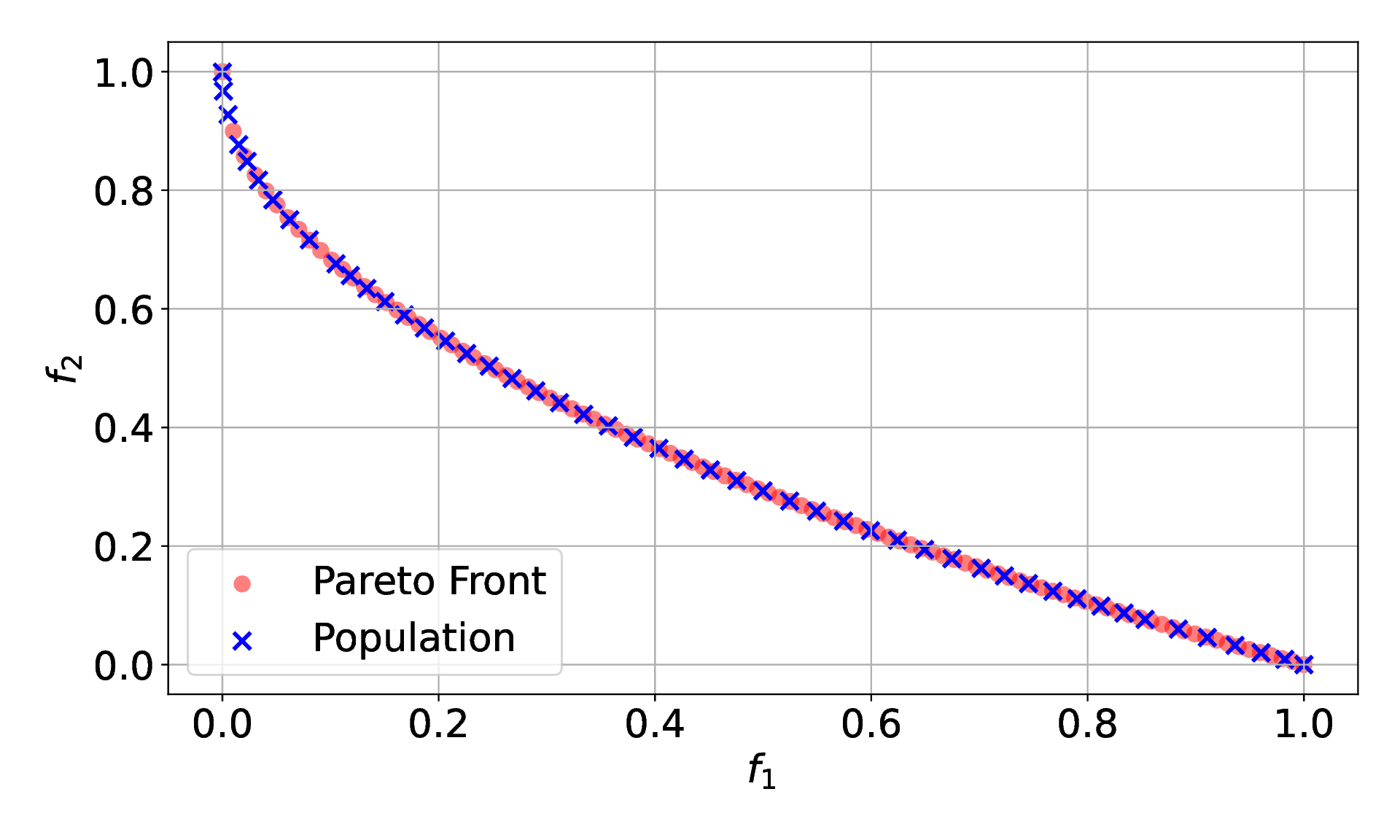}
        \caption{Epoch 3000}
        \label{fig:zdt1_particle_3000}
    \end{subfigure}
    \begin{subfigure}[b]{0.16\linewidth}
        \includegraphics[width=\textwidth]{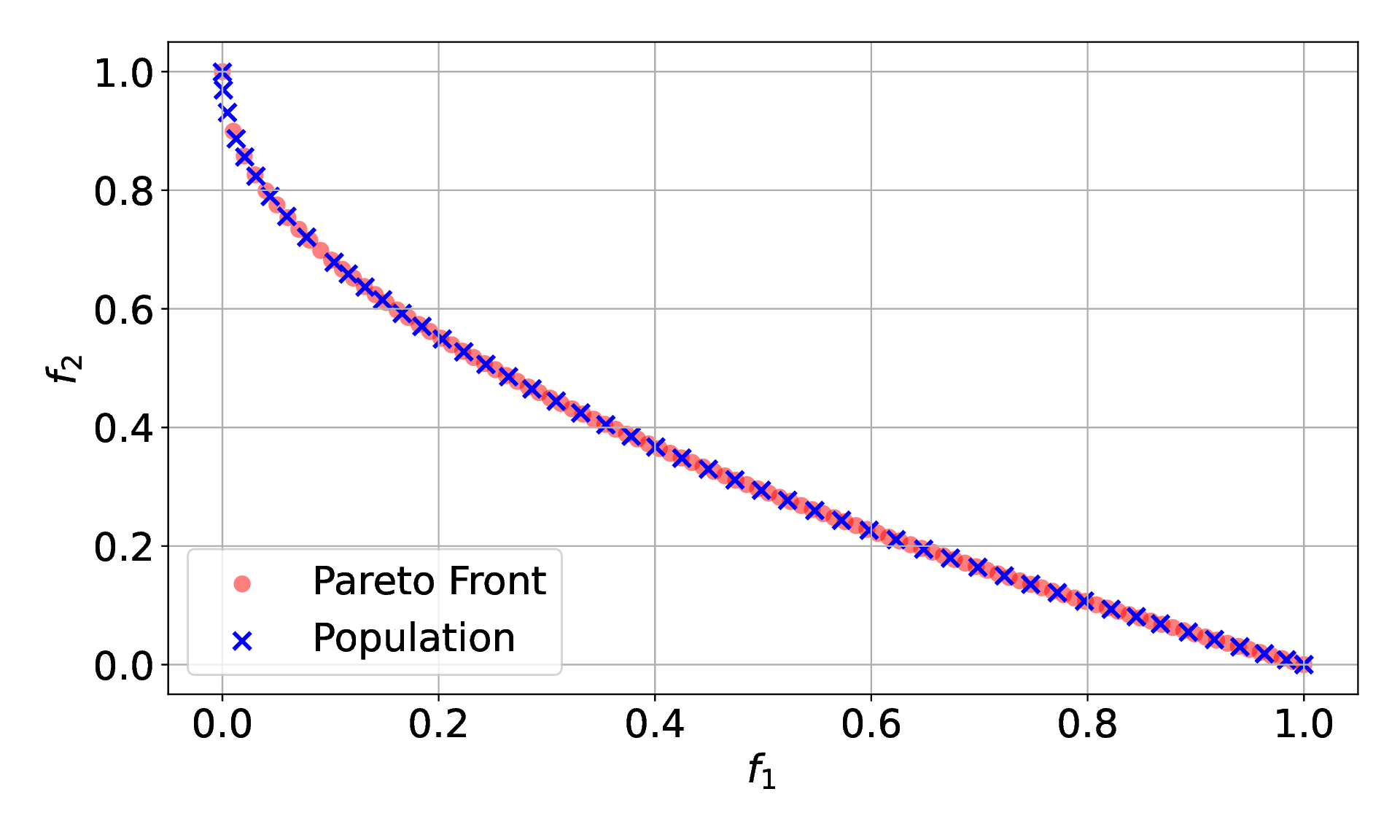}
        \caption{Epoch 4000}
        \label{fig:zdt1_particle_4000}
    \end{subfigure}
    \begin{subfigure}[b]{0.16\linewidth}
        \includegraphics[width=\textwidth]{images/zdt1_particle_5000.png}
        \caption{Epoch 5000}
        \label{fig:zdt1_particle_5000}
    \end{subfigure}
    \caption{Evolution of the particle population by Particle-WFR on the ZDT1 problem. The Pareto front is shown in red, and the current population is shown in blue.}
    \label{fig:zdt1_particle_evo}
\end{figure}

As shown in Figure~\ref{fig:zdt1}, most of the methods can cover the whole Pareto front, but the weighted sum method, MOO-LD, and COSMOS are not able to cover the Pareto front uniformly. Our Particle-WFR method and MOO-SVGD give the best and most uniform coverage of the Pareto front. The evolution of the particle population by Particle-WFR is shown in Figure~\ref{fig:zdt1_particle_evo}. 

\subsubsection{Additional Experiments on the ZDT2 Problem}
\label{app:zdt2}

The ZDT2 problem~\citep{zitzler2000comparison} is also a 30-dimensional two-objective optimization problem of the same form as in~\eqref{eq:zdt}
but with $$
h(f_1, g) = 1 - \left(\dfrac{f_1(\x)}{g(\x)}\right)^2.
$$
where the feasible region is $\D = [0,1]^{30}$. Unlike the ZDT1 problem, the Pareto front of the ZDT2 problem is concave. Similar examples of concave Pareto fronts have been used in the literature, including the Fonseca problem~\citep{fonseca1995overview,sener2018multi,lin2019pareto,mahapatra2020multi} and the DTLZ2 problem~\citep{chen2022multi}.

\begin{figure}[!htb]
    \centering
    \begin{subfigure}[b]{0.3\linewidth}
        \includegraphics[width=\textwidth]{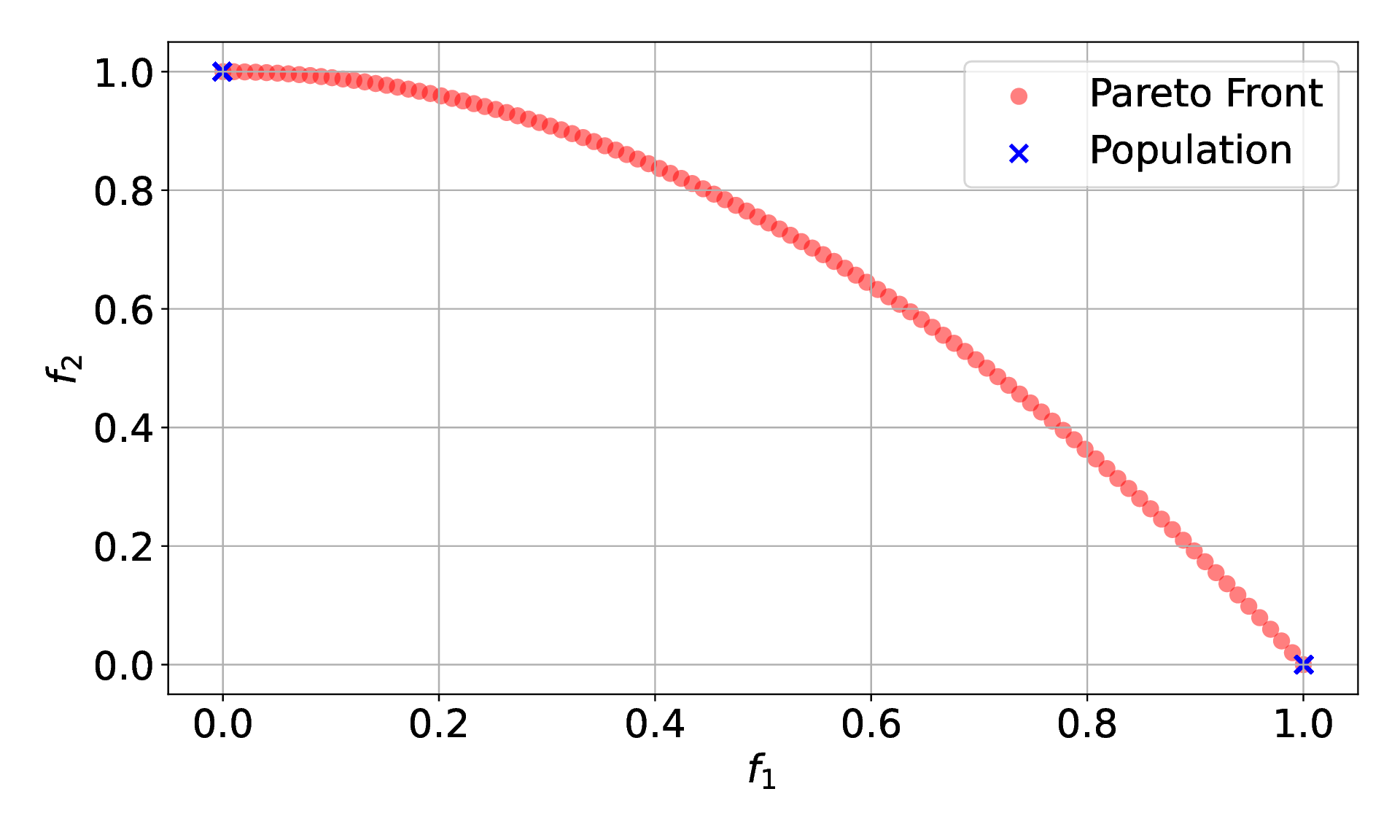}
        \caption{Weighted Sum}
        \label{fig:zdt2_linear}
    \end{subfigure}
    \begin{subfigure}[b]{0.3\linewidth}
        \includegraphics[width=\textwidth]{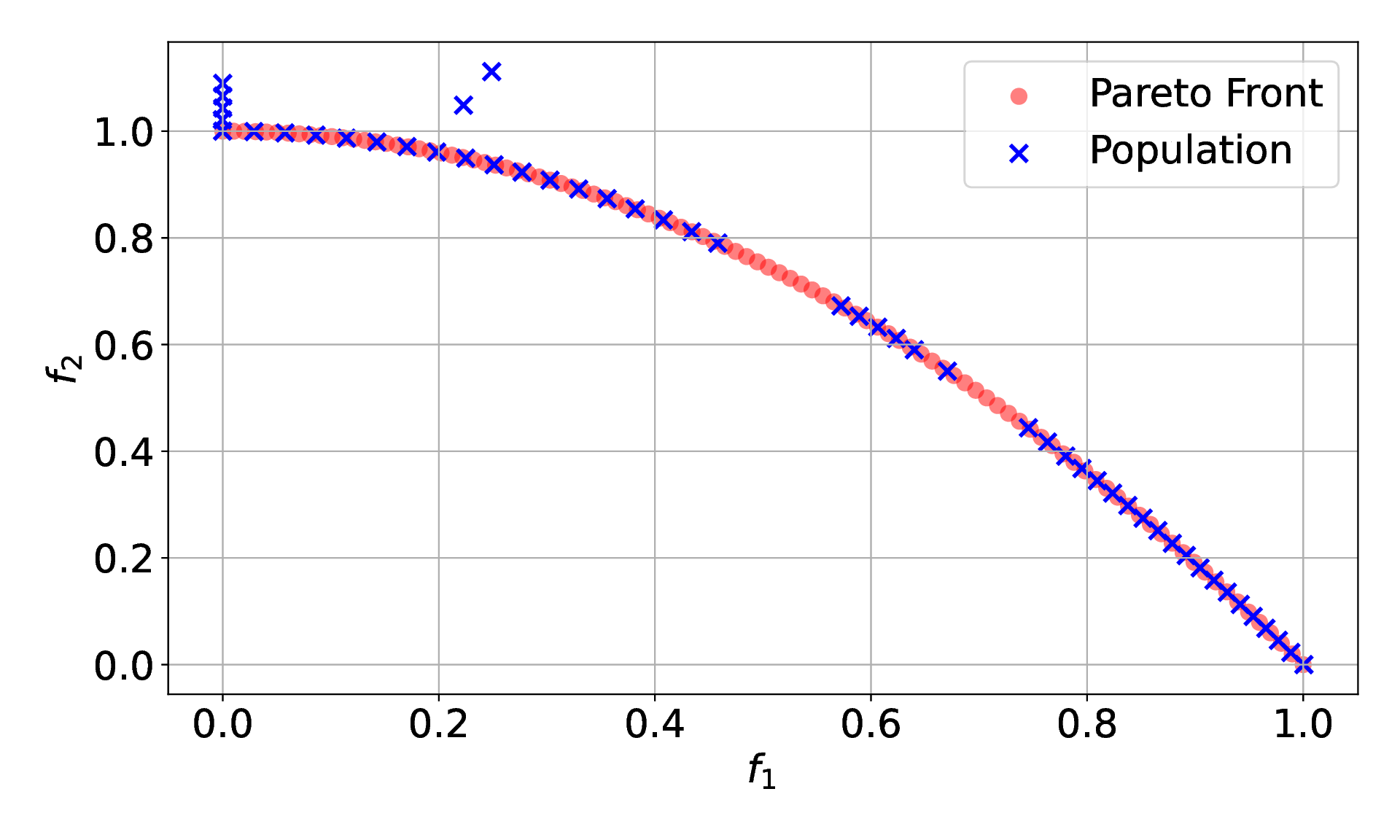}
        \caption{MOO-SVGD}
        \label{fig:zdt2_moosvgd}
    \end{subfigure}
    \begin{subfigure}[b]{0.3\linewidth}
        \includegraphics[width=\textwidth]{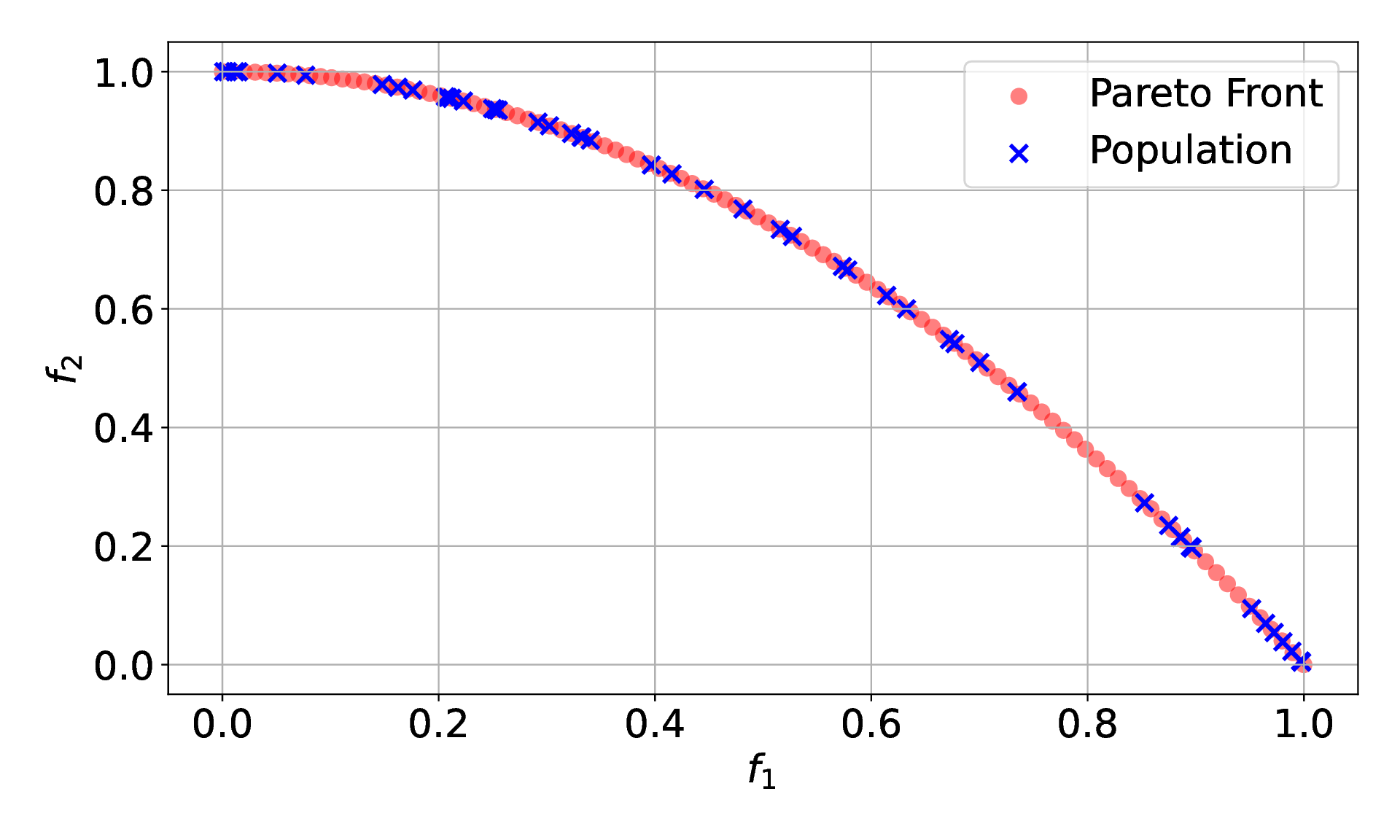}
        \caption{MOO-LD}
        \label{fig:zdt2_moold}
    \end{subfigure}

    \begin{subfigure}[b]{0.3\linewidth}
        \includegraphics[width=\textwidth]{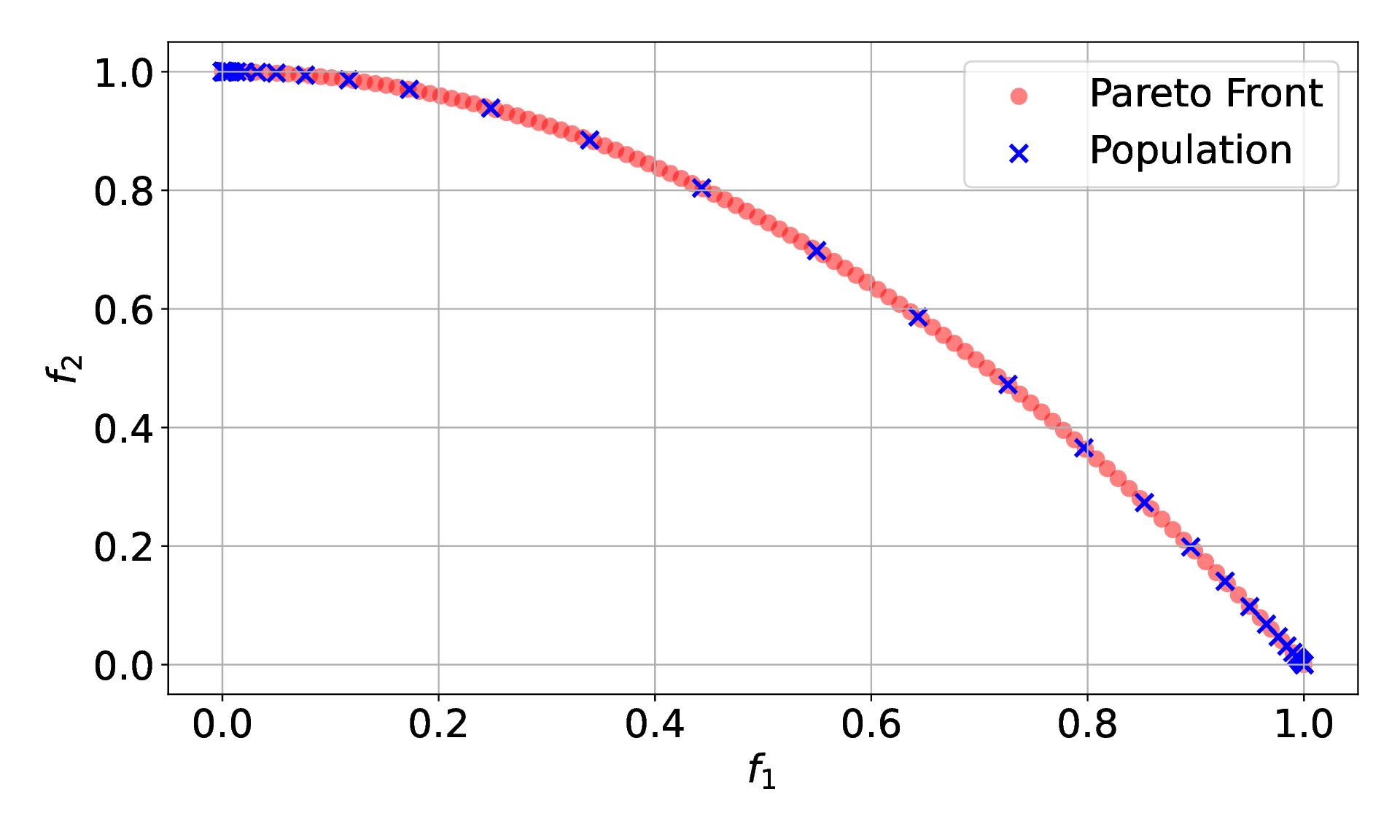}
        \caption{COSMOS}
        \label{fig:zdt2_cosmos}
    \end{subfigure}
    \begin{subfigure}[b]{0.3\linewidth}
        \includegraphics[width=\textwidth]{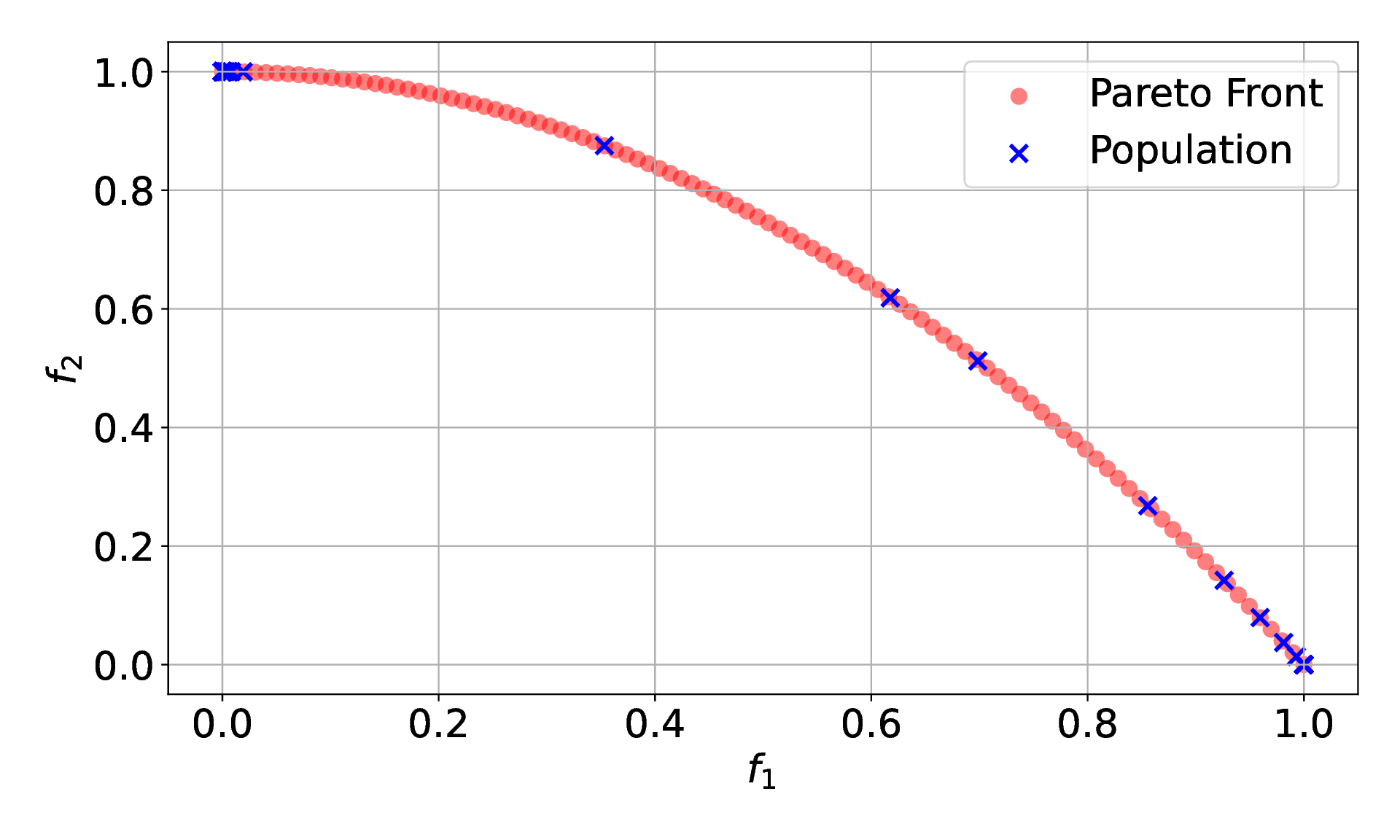}
        \caption{GMOOAR-HV}
        \label{fig:zdt2_argmo}
    \end{subfigure}
    \begin{subfigure}[b]{0.3\linewidth}
        \includegraphics[width=\textwidth]{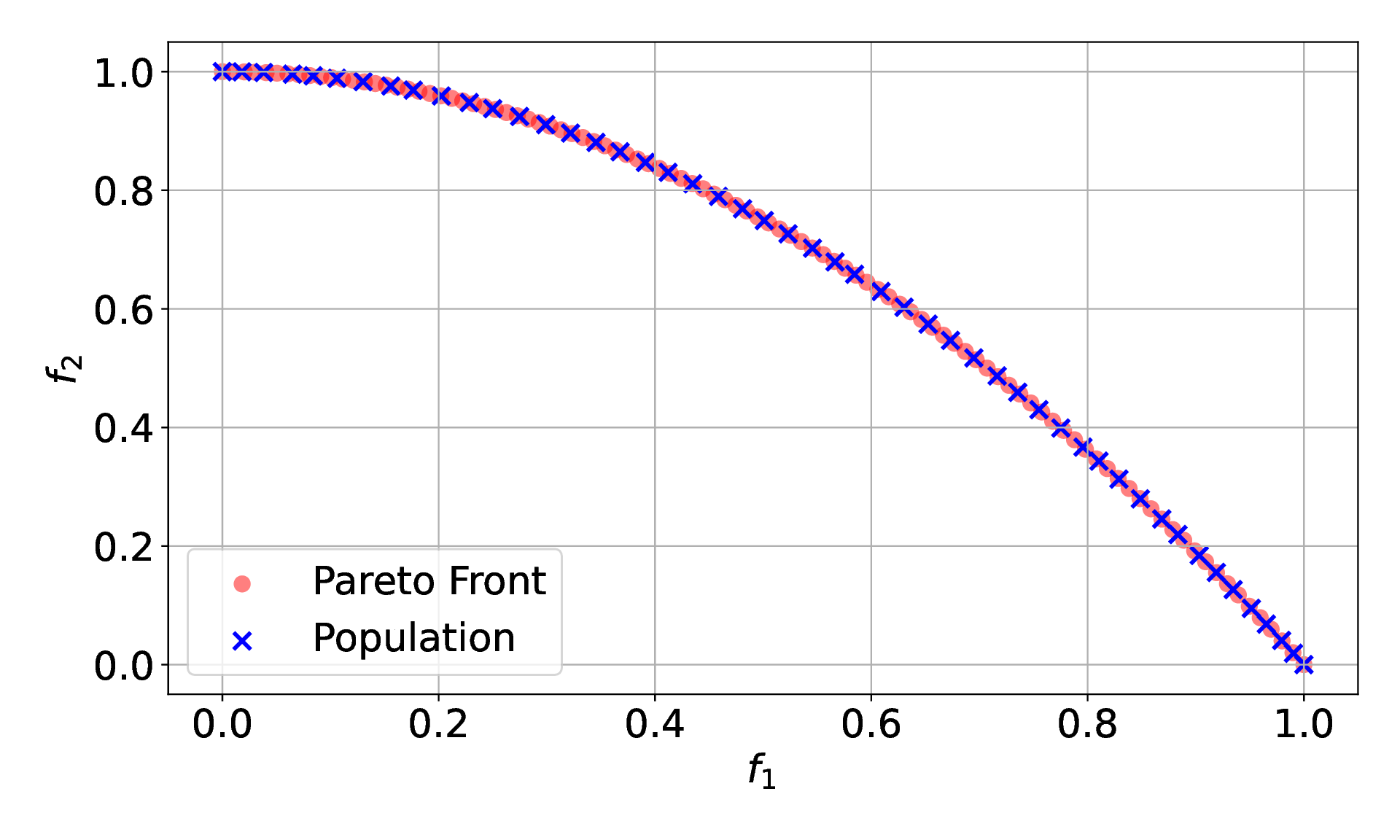}
        \caption{\bf Particle-WFR}
        \label{fig:zdt2_wfr}
    \end{subfigure}
    \caption{Performance comparison of different methods on the ZDT2 problem. The Pareto front is shown in red, and the solutions found by different methods are shown in blue.}
    \label{fig:zdt2}
\end{figure}

\begin{figure}[!htb]
    \centering
    \begin{subfigure}[b]{0.16\linewidth}
        \includegraphics[width=\textwidth]{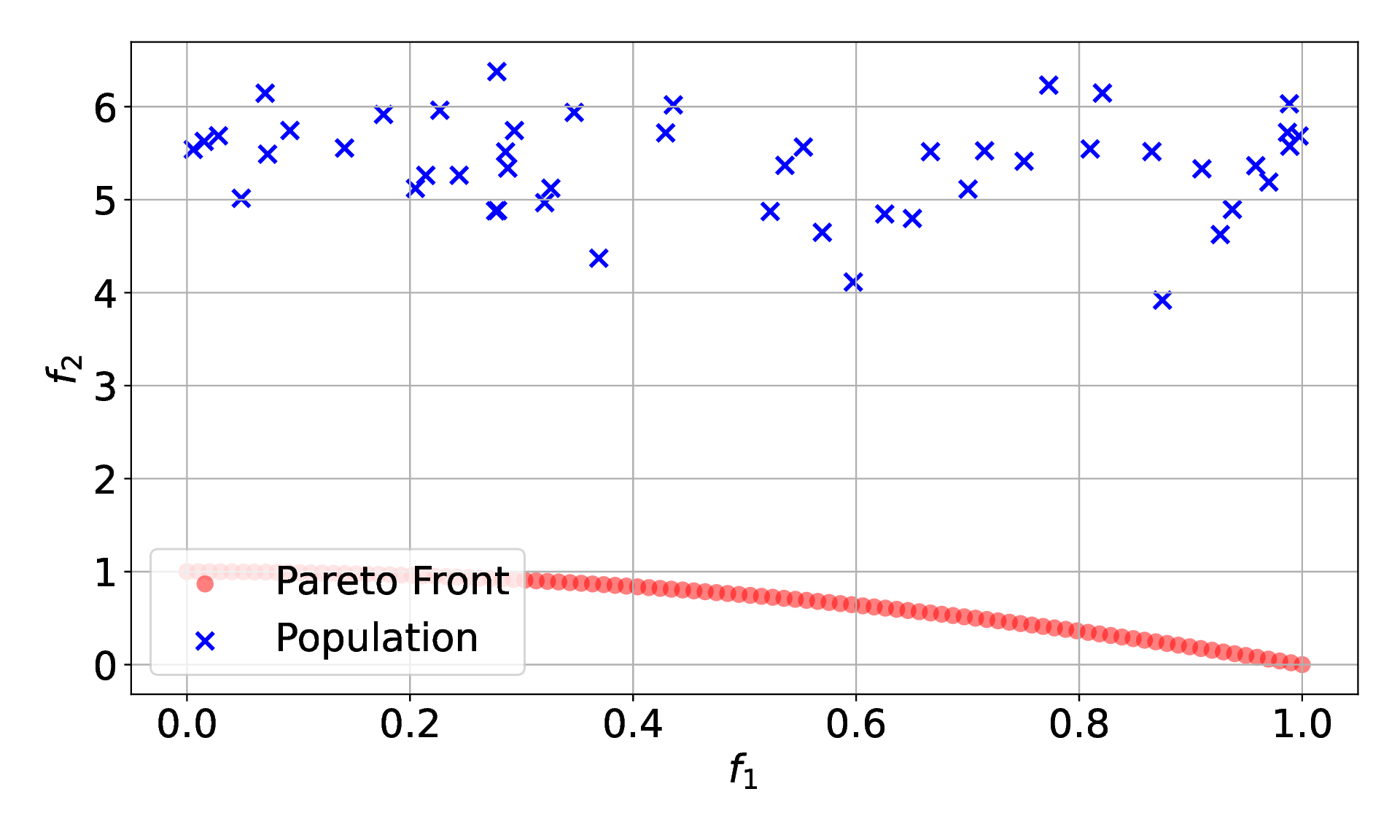}
        \caption{Epoch 0}
        \label{fig:zdt2_particle_0}
    \end{subfigure}
    \begin{subfigure}[b]{0.16\linewidth}
        \includegraphics[width=\textwidth]{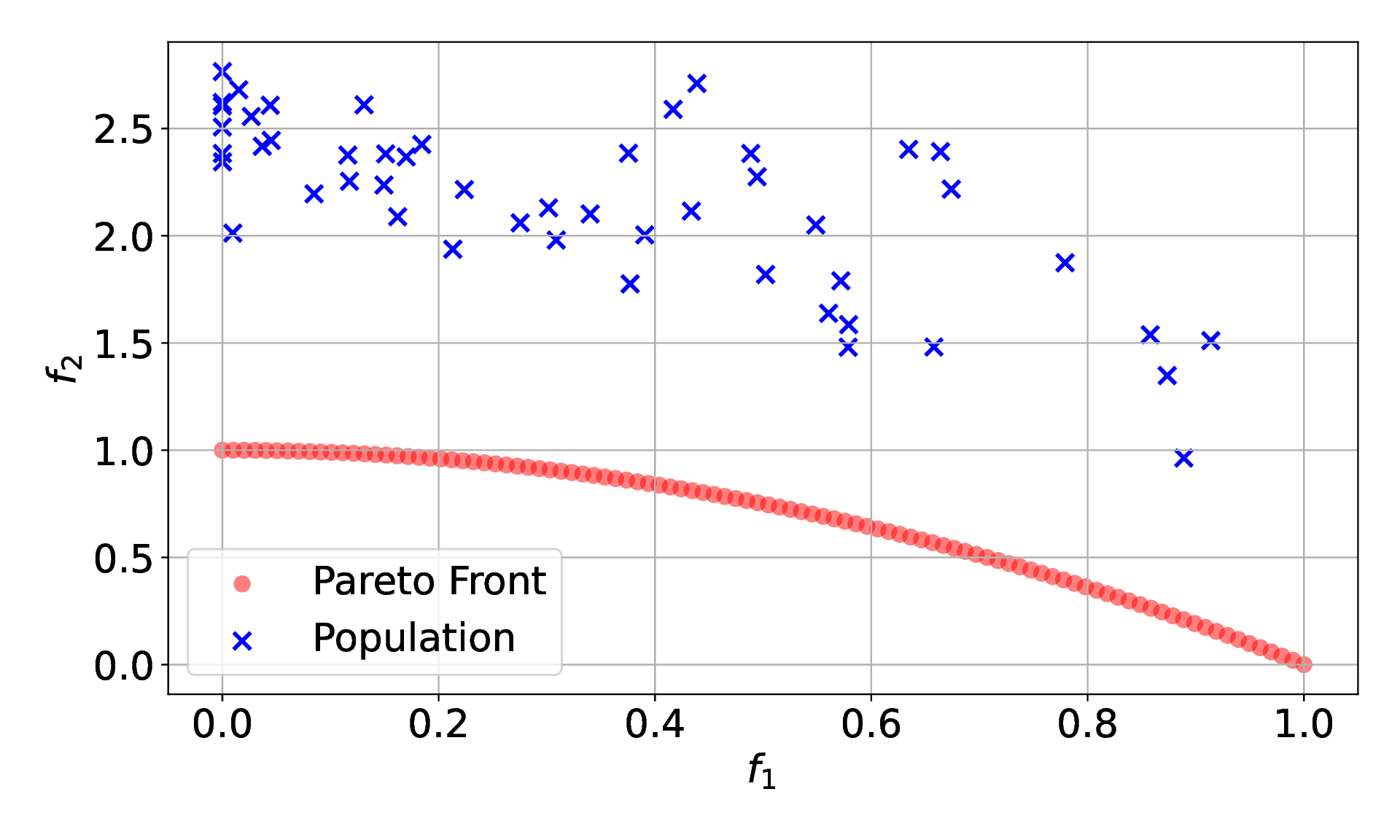}
        \caption{Epoch 1000}
        \label{fig:zdt2_particle_1000}
    \end{subfigure}
    \begin{subfigure}[b]{0.16\linewidth}
        \includegraphics[width=\textwidth]{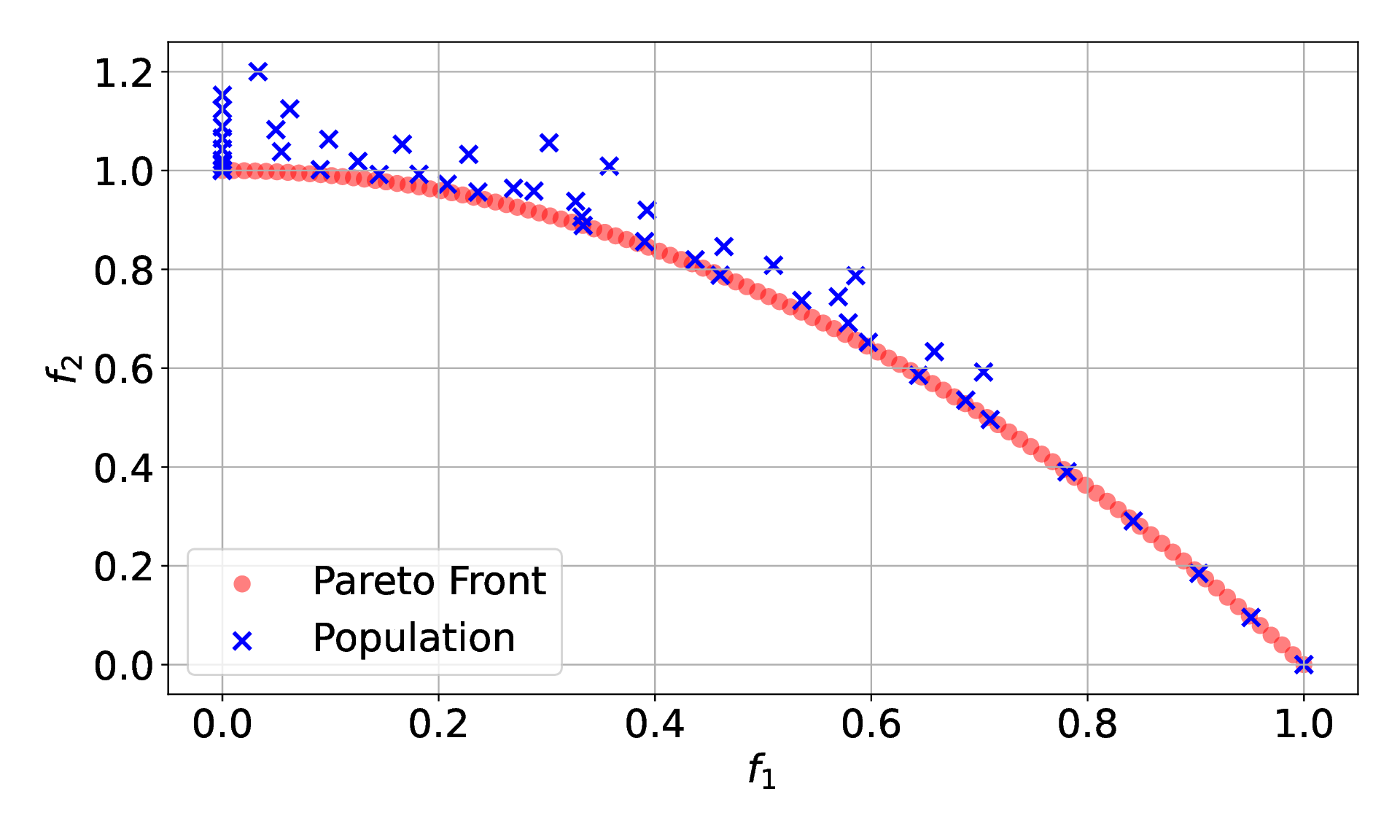}
        \caption{Epoch 2000}
        \label{fig:zdt2_particle_2000}
    \end{subfigure}
    \begin{subfigure}[b]{0.16\linewidth}
        \includegraphics[width=\textwidth]{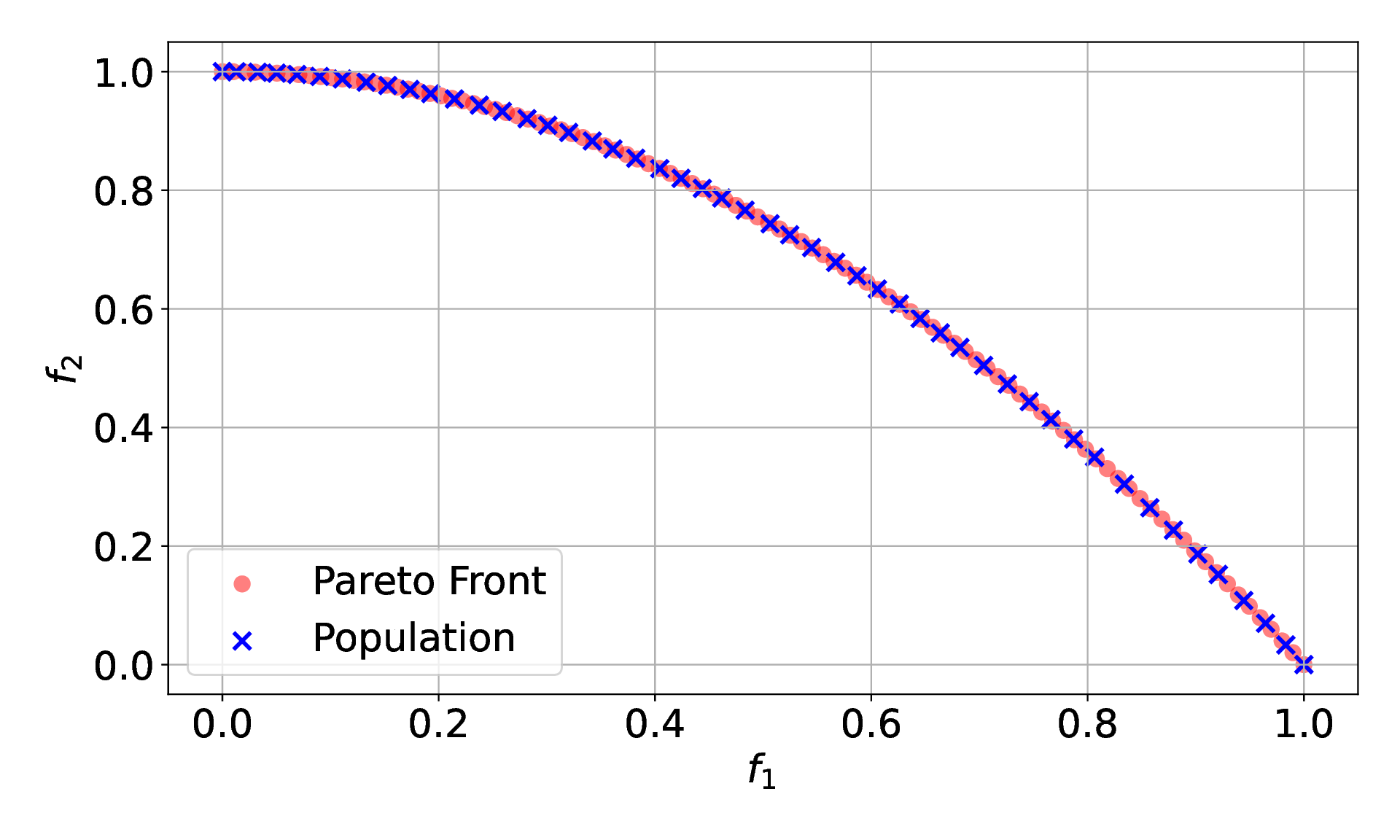}
        \caption{Epoch 3000}
        \label{fig:zdt2_particle_3000}
    \end{subfigure}
    \begin{subfigure}[b]{0.16\linewidth}
        \includegraphics[width=\textwidth]{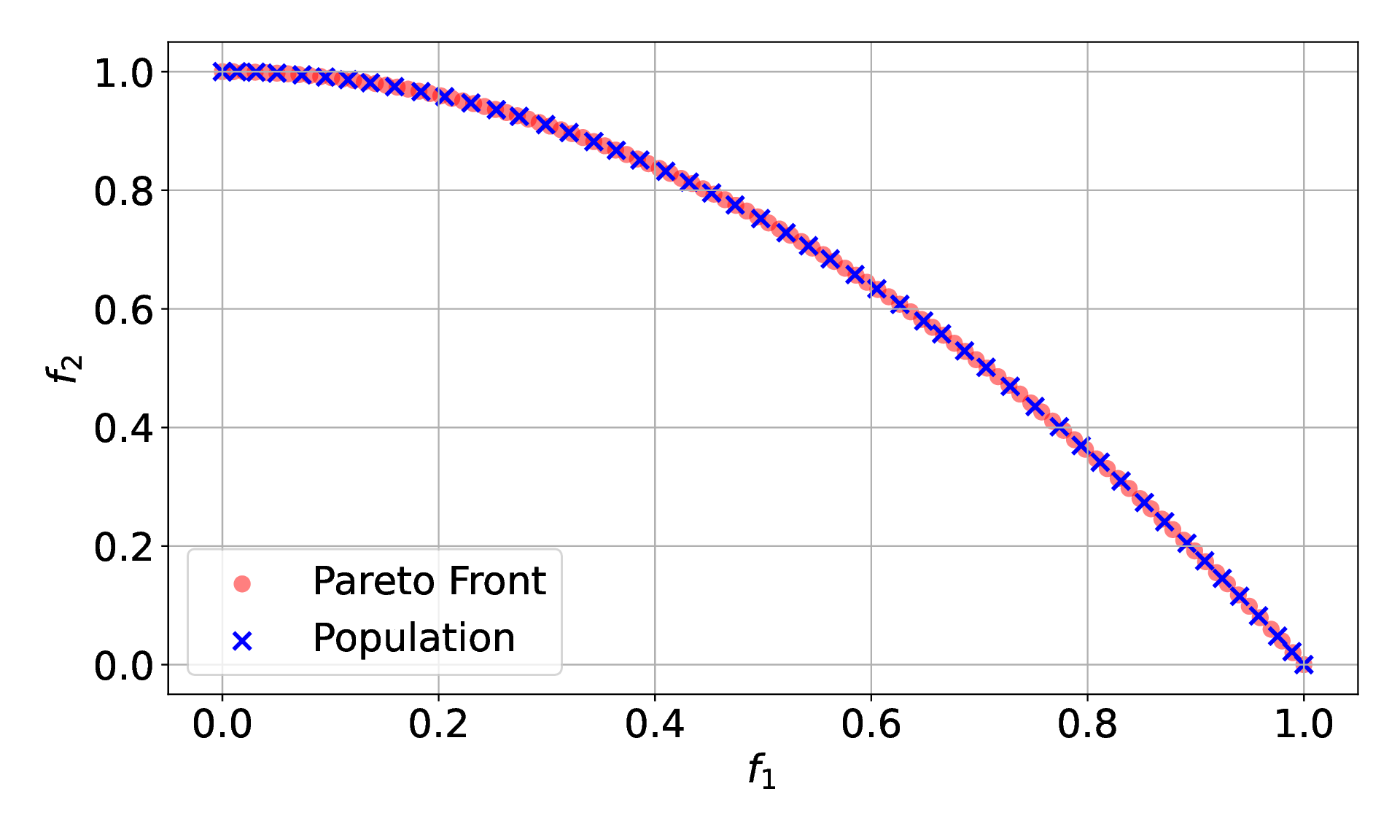}
        \caption{Epoch 4000}
        \label{fig:zdt2_particle_4000}
    \end{subfigure}
    \begin{subfigure}[b]{0.16\linewidth}
        \includegraphics[width=\textwidth]{images/zdt2_particle_5000.png}
        \caption{Epoch 5000}
        \label{fig:zdt2_particle_5000}
    \end{subfigure}
    \caption{Evolution of the particle population by Particle-WFR on the ZDT2 problem. The Pareto front is shown in red, and the current population is shown in blue.}
    \label{fig:zdt2_particle_evo}
\end{figure}

As shown in Figure~\ref{fig:zdt2}, our Particle-WFR method can still cover the whole Pareto front uniformly. The weighted sum method fails in this case, and all solutions are concentrated on the two ends of the Pareto front. MOO-LD, COSMOS, and GMOOAR-HV are able to cover the Pareto front but not uniformly. MOO-SVGD performs well at most regions of the Pareto front, but two gaps of solutions are observed on the Pareto front, and several sub-optimal points exist near the upper-left end $(0, 1)$ of the Pareto front. The evolution of the particle population by Particle-WFR is shown in Figure~\ref{fig:zdt2_particle_evo}.

\subsection{DTLZ7 Problem}
\label{app:dtlz7}
 
The DTLZ7 problem has the following form of $\f(\x)$:
\begin{equation}
    \begin{aligned}
        f_1(\x) & = x_1,\\
        f_2(\x) & = x_2,\\
        f_3(\x) & = (1 + g(\x))h(f_1(\x), f_2(\x), g(\x)),
    \end{aligned}
\end{equation}
where $\x = (x_1,\cdots, x_{30})$, $$g(\x) = 1 + \frac{9}{29}\sum_{i=3}^{30} x_i$$ and 
$$h(f_1(\x), f_2(\x), g(\x)) = 3 - \sum_{i=1}^{2} \frac{f_i(\x)}{1 + g(\x)}(1+\sin(3\pi f_i(\x))).$$ The feasible region is also $\D = [0,1]^{30}$.

\subsection{MSLR-WEB10K Dataset}
\label{app:mslr}

In this section, we provide additional details of the learning-to-rank task and the MSLR-WEB10K dataset.

\subsubsection{Learning-to-Rank Task}
\label{app:ltr}

For readers' convenience, we repeat the settings of the learning-to-rank task as provided in Section~\ref{sec:ltr} in the main text as follows.

Suppose we have a collection of \emph{query groups} $\Psi =  \{\Psi^{(p)}\}_{p=1}^{|\Psi|}$, where each query group $\Psi^{(p)}$ consists of $n^{(p)}$ documents. These items are characterized by a feature vector $\x^{(p)}_j \in \R^{d_f}$, generated from upstream tasks, and an associated relevance label $y^{(p)}_j$. In our case, the relevance labels are assumed to be positive. The goal is to derive an ordering $\pi^{(p)}$, \emph{i.e.}
\begin{equation*}
    \begin{aligned}
        \pi^{(p)} : \{1, \cdots, n^{(p)}\} &\rightarrow \{1, \cdots, n^{(p)}\}\\
        j &\mapsto \pi^{(p)}_j,
    \end{aligned}
\end{equation*}
for the items in each query group $\Psi^{(p)}$ given the feature vectors $\{\x^{(p)}_{j}\}_{j=1}^{n^{(p)}}$, that optimizes the utility $u(\pi^{(p)}; \{y^{(p)}_j\}_{j=1}^{n^{(p)}})$ of the ordered list. We will denote the set of all possible orderings of the items in $\Psi^{(p)}$ as $\Pi^{(p)}$.

Utility functions (or ranking metrics) are operators that measure the quality of the ordering $\pi^{(p)}$ with respect to the relevance labels $\{y^{(p)}_j\}_{j=1}^{n^{(p)}}$. Intuitively, the utility function should be large if the items with higher relevance labels are ranked higher in the ordering $\pi^{(p)}$.
One of the most widely adopted measures for the utility function $u(\cdot;\cdot)$ above is the Normalized Discount Cumulative Gain (NDCG)~\citep{wang2013theoretical}, which is defined as
\begin{equation}
    \ndcg(\pi^{(p)};\{y^{(p)}_j\}_{j=1}^{n^{(p)}}) = \dfrac{\dcg(\pi^{(p)};\{y^{(p)}_j\}_{j=1}^{n^{(p)}})}{\dcg(\pi^{(p),*};\{y^{(p)}_j\}_{j=1}^{n^{(p)}})},
\end{equation}
where $\dcg(\pi^{(p)};\{y^{(p)}_j\}_{j=1}^{n^{(p)}})$ is the discounted cumulative gain of the ordering $\pi^{(p)}$, defined as
\begin{equation*}
    \dcg(\pi^{(p)};\{y^{(p)}_j\}_{j=1}^{n^{(p)}}) = \sum_{j=1}^{n^{(p)}} \dfrac{2^{y^{(p)}_{\pi^{(p)}_j}} - 1}{\log_2(1+j)},
\end{equation*} 
and $\pi^{(p),*}$ is the optimal ordering of the items in $\Psi^{(p)}$, \emph{i.e.}
\begin{equation*}
    \pi^{(p),*} = \argmax_{\pi^{(p)}\in \Pi^{(p)}} \dcg(\pi^{(p)};\{y^{(p)}_j\}_{j=1}^{n^{(p)}}).
\end{equation*}

In practical settings, one may also be interested in the truncated versions of the NDCG, denoted as NDCG@$k$, where only the top $k$ items in the ordering $\pi^{(p)}$ are considered, \emph{i.e.}
\begin{equation}
    \ndcg@k(\pi^{(p)};\{y^{(p)}_j\}_{j=1}^{n^{(p)}}) = \dfrac{\dcg@k(\pi^{(p)};\{y^{(p)}_j\}_{j=1}^{n^{(p)}})}{\dcg@k(\pi^{(p),*};\{y^{(p)}_j\}_{j=1}^{n^{(p)}})},
\label{eq:ndcg_k}
\end{equation}
where $\dcg@k(\cdot;\cdot)$ is defined by replacing the summation in $\dcg(\cdot;\cdot)$ with the summation over the top $k$ items in the ordering $\pi^{(p)}$. For datasets with more than one query group, the utility function $u(\cdot;\cdot)$ is defined as the average of the NDCG@$k$ over all the query groups.

Most of the current LTR methods employ a neural network $f_\theta$, with $\theta$ denoting the parameters, to produce a score for each item, based on which the ordering $\pi^{(p)}$ is obtained. In particular, the neural network $f_\theta$ accepts the feature vector $\x_j^{(p)}$ of the $j$-th item in the query group $\Psi^{(p)}$ as input and produces a score $f_\theta(\x_j^{(p)})$ for the item. Then, the ordering $\pi^{(p)}$ is obtained by sorting the items in $\Psi^{(p)}$ according to the scores produced by $f_\theta$.

The neural network is trained using the empirical loss of the following form:
\begin{equation*}
    \L(\theta; \Psi) = \dfrac{1}{|\Psi|} \sum_{p=1}^{|\Psi|}      \ell\left(\{f_\theta(\x_j^{(p)})\}_{j=1}^{n^{(p)}}; \{y_j^{(p)}\}_{j=1}^{n^{(p)}}\right),
\end{equation*}
where $\ell(\cdot, \cdot)$ is the query group-wise loss function. 
One should notice that the loss function $\ell(\cdot;\cdot)$ has a different nature than the utility function $u(\cdot;\cdot)$, as the latter takes in an ordering $\pi^{(p)}$, while the former takes in a set of scores $\{f_\theta(\x_j^{(p)})\}_{j=1}^{n^{(p)}}$. Consequently,
the loss function $\ell(\cdot;\cdot)$ is differentiable, while the utility function $u(\cdot;\cdot)$ is not.

In order to bridge this gap caused by the non-differentiability of the utility functions, such as the NDCG. Many works use differentiable surrogates for the NDCG metric as the loss function for training, including the \emph{Cross-Entropy (CE) loss}~\citep{cao2007learning} for $\ell$, defined as:
\begin{equation}
    \ell_{\mathrm{CE}}\left(\{f_\theta(\x_j^{(p)})\}_{j=1}^{n^{(p)}}; \{y_j^{(p)}\}_{j=1}^{n^{(p)}}\right) = - \sum_{j=1}^{n^{(p)}} y_j^{(p)} \log \dfrac{\exp(f_\theta(\x_j^{(p)}))}{\sum_{j=1}^{n^{(p)}}\exp(f_\theta(\x_j^{(p)}))},
\label{eq:ce}
\end{equation}
where we employ the softmax function to the scores produced by the neural network $f_\theta$ to obtain a probability distribution over the items in $\Psi^{(p)}$.

As analyzed and empirically verified in~\citet{qin2021neural}, this choice of the loss function, referred to as the softmax loss, is one of the simplest and most robust choices for differentiable surrogates in the LTR task. We direct the readers to~\citet{qin2021neural} for more details on other popular surrogates and the discussions therein.

\subsubsection{Implementation Details of MSLR-WEB10K Dataset}
\label{app:mslr_imp}

The Microsoft Learning-to-Rank Web Search (MSLR-WEB10K) dataset~\citep{qin2013introducing} is one of the most widely used benchmark datasets for the LTR task.
The MSLR-WEB10K dataset consists of 10,000 query groups ($|\Psi| = 10^4$), each representing a query issued by a user. Each query group contains a list of items, each of which is a URL retrieved by the search engine in response to the query. Each item is characterized by a feature vector $\x_j^{(p)} \in \R^{136}$, extracted from the webpage, and a relevance label $y_j^{(p)} \in \{0, 1, 2, 3, 4\}$, indicating the relevance of the item to the query. Following the practice of~\citep{mahapatra2023multi}, we treat the first 131 features as the input ($d_f = 131$) and combine the last 5 features, \emph{viz.} Query-URL Click Count, URL Dwell Time, Quality Score 1, Quality Score 2, with the relevance label, as six different ranking objectives ($m=6$). 

We adopt a simple Multi-Layer Perception (MLP) of architecture $[131, 32, 1]$ as the neural network $f_\theta$ for the LTR task. We train the neural network with the Adam optimizer with a learning rate of $10^{-3}$ and a batch size of 512. The training loss is chosen to be the CE loss $\ell_{\mathrm{CE}}$ as defined in~\eqref{eq:ce} and \eqref{eq:loss}. NDCG@$10$~\eqref{eq:ndcg_k} is used as the test metric.
The training process is terminated after 500 epochs.

As the first practice of applying the interacting particle method to a multi-objective LTR task, our Particle-WFR method features the acceleration of Distributed Data Parallel (DDP) in PyTorch \citep{paszke2019pytorch}. The DDP is a distributed training strategy that allows the training process to be distributed across multiple GPUs. In our case, we use 4 GPUs to train the neural network $f_\theta$ in parallel, and we expect further scalability by using more GPUs in real applications. Our code will be made publicly available upon publication.

\subsubsection{Hypervolume Indicator}
\label{app:hv}

The hypervolume (HV) indicator~\citep{zitzler2004indicator} is a widely used metric for evaluating the performance of MOO methods. The hypervolume of a set of points $\hat \P$ that approximate the Pareto front $\P$ is defined as the volume of the dominated region of $\hat\P$ with respect to a reference point $\boldsymbol r$, \emph{i.e.}
\begin{equation}
    {\rm HV}(\hat\P) = \int_{\R^m} \bm 1\{\x \preceq \boldsymbol r \mid  \exists \y\in \hat\P \text{ s.t. } \y \preceq \x \} \mathrm{d}\x,
\end{equation}
where $\preceq$ denotes the Pareto dominance relation as in Definition~\ref{def:pareto}.
Hypervolume not only measures the optimality of the solutions found by the MOO methods but also measures the diversity of the solutions. A larger hypervolume indicates that the solutions are more diverse and closer to the Pareto front.
Furthermore, the Pareto front $\P$ itself achieves the highest possible hypervolume.

\begin{table}[!htb]
    \centering
    \begin{tabular}{c|ccc}
        \toprule[1pt]
        \multirow{2}{*}{Method} & \multicolumn{3}{c}{HV of Test NDCG@10} \\
        & $N=8$ & $N=12$ & $N=16$  \\
        \midrule[1pt]
        PHN-LS~\citep{navon2020learning} & 3.51$\pm$0.63 (4.58) & 4.01$\pm$0.41 (4.79) & 4.17$\pm$0.40 (5.04) \\ 
        PHN-EPO~\citep{navon2020learning} & 3.60$\pm$0.63 (4.51) & 3.65$\pm 0.65$ (4.91) & 4.15$\pm$0.67 (5.23) \\ 
        COSMOS~\citep{ruchte2021scalable} & 3.81$\pm$0.33 (4.37) & 4.00$\pm$0.37 (5.02) & 4.19$\pm$0.32 (5.19) \\ 
        GMOOAR-HV~\citep{chen2022multi} & 2.57$\pm$0.17 (2.86) & 3.25$\pm$0.29 (3.69) & 3.65$\pm$0.32 (4.05) \\ 
        GMOOAR-U~\citep{chen2022multi}  & 1.90$\pm${\bf 0.09} (2.11) & 4.26$\pm$0.25 (4.66) & 4.07$\pm${\bf 0.16} (4.42) \\ 
        {\bf Particle-WFR (Ours)} & {\bf 6.48}$\pm$0.38 ({\bf 7.27}) & \textbf{7.07}$\pm${\bf 0.23} ({\bf 7.48}) & \textbf{6.95}$\pm$0.26 ({\bf 7.60}) \\ 
        \bottomrule[1pt]
    \end{tabular}
    \caption{Performance comparison of different methods on the MSLR-WEB10K dataset. The hypervolume (HV) is presented in the form of mean $\pm$ std (max) over the last 30 epochs, with the unit being $10^{-4}$.} 
    \label{tab:mslr}
\end{table}

Figure~\ref{fig:mslr} and Table~\ref{tab:mslr} presents the performance comparison of different methods on the MSLR-WEB10K dataset. The hypervolume values therein are computed with respect to the NDCG@10 metric evaluated on the test set, apart from the training loss. 
In the case of NDCG@10, we have to modify the definition of the hypervolume indicator to account for the fact that the NDCG@10 metric is being maximized, which can be resolved by adding a minus sign to all the values involved and choosing the reference point $\boldsymbol r$ to be the origin.
The values are of the order of $10^{-4}$ because of the high-dimensionality ($m=6$) of the MOO problem, noticing that the value of NDCG is between 0 and 1.

\end{document}